\documentclass[11pt,twoside]{article} 
\usepackage[table,xcdraw]{xcolor} 
\usepackage{hyperref}             
\usepackage{amsmath,amsfonts,bm}
\usepackage{eqnarray,amsmath}
\usepackage[utf8]{inputenc} 
\usepackage[T1]{fontenc}    
\usepackage{booktabs}       
\usepackage{amsfonts}       
\usepackage{nicefrac}       
\usepackage[nopatch=footnote]{microtype}      
\usepackage{subcaption}
\expandafter\def\csname 
ver@subfig.sty\endcsname{}
\usepackage{eqnarray,amsmath}
\usepackage{epsf}
\usepackage{epsfig}
\usepackage{fancyhdr}
\usepackage{graphics}
\usepackage{graphicx}
\usepackage{psfrag}
\usepackage{fullpage}
\usepackage{pdfpages}

\usepackage{url}

\usepackage{color}

\usepackage{amsthm}
\usepackage{amsmath}
\usepackage{amssymb,bbm}
\usepackage[skip=0pt]{caption}  
\usepackage{algorithmic}
\usepackage{algorithm}
\usepackage{textcomp}
\usepackage{siunitx}
\usepackage{wrapfig}
\usepackage{algorithmic}
\usepackage{algorithm}
\usepackage{multirow}
\usepackage{multicol}
\usepackage{makecell}
\usepackage{colortbl} 
\usepackage{changepage}
\usepackage{academicons}
\usepackage{fontawesome}

\usepackage[utf8]{inputenc} 
\usepackage[T1]{fontenc}    
\usepackage{hyperref}       
\usepackage{url}            
\usepackage{booktabs}       
\usepackage{amsfonts}       
\usepackage{nicefrac}       
\usepackage{microtype}      

\usepackage{graphicx}        
\usepackage{caption}         
\usepackage{subcaption}      
\usepackage{float,wrapfig}           
\usepackage{lipsum}          
\usepackage{amssymb} 
\usepackage{amsmath,amsfonts,bm}
\usepackage{pifont}
\usepackage{amsmath}
\usepackage{amssymb}
\usepackage{mathtools}
\usepackage{amsthm}
\usepackage{amsmath, amssymb, mathrsfs}
\usepackage{float}
\usepackage{placeins}
\usepackage{makecell}
\usepackage[normalem]{ulem} 

\usepackage[capitalize,noabbrev]{cleveref}

\usepackage{colortbl}   
\usepackage{tocloft} 
\usepackage{etoc}
\usepackage{setspace}
\usepackage{natbib}

\theoremstyle{plain}
\newtheorem{theorem}{Theorem}[section]
\newtheorem{proposition}[theorem]{Proposition}
\newtheorem{lemma}[theorem]{Lemma}

\theoremstyle{definition}

\newtheorem{assumption}[theorem]{Assumption}
\theoremstyle{remark}

\definecolor{customyellow}{HTML}{fedf8a}
\usepackage{eqnarray,amsmath}
\usepackage{epsf}
\usepackage{epsfig}
\usepackage{fancyhdr}
\usepackage{graphics}
\usepackage{graphicx}
\usepackage{psfrag}
\usepackage{fullpage}
\usepackage{pdfpages}
\usepackage{ragged2e}

\usepackage{amsmath,amsfonts,bm}

\usepackage{amsthm}
\usepackage{amsmath}
\usepackage{amssymb,bbm}
\usepackage{textcomp}
\usepackage{siunitx}
\usepackage{wrapfig}
\usepackage{multirow}
\usepackage{multicol}
\usepackage{epsf}
\usepackage{epsfig}
\usepackage{fancyhdr}
\usepackage{booktabs}       
\usepackage{amsfonts}       
\usepackage{nicefrac}       
\usepackage{microtype}
\usepackage{soul}
\usepackage{mathtools}

\DeclarePairedDelimiterX{\infdivx}[2]{(}{)}{%
  #1\;\delimsize\|\;#2%
}

\def\eqref#1{equation~(\ref{#1})}

\def\1{\bm{1}}

\def\vs{{\bm{s}}}

\def\vx{{\bm{x}}}

\def\vz{{\bm{z}}}

\DeclareMathAlphabet{\mathsfit}{\encodingdefault}{\sfdefault}{m}{sl}
\SetMathAlphabet{\mathsfit}{bold}{\encodingdefault}{\sfdefault}{bx}{n}

\def\gC{{\mathcal{C}}}
\def\gD{{\mathcal{D}}}

\def\gL{{\mathcal{L}}}

\def\gR{{\mathcal{R}}}

\def\gX{{\mathcal{X}}}
\def\gY{{\mathcal{Y}}}

\newcommand{\bbE}{\mathbb{E}}

\DeclareMathOperator*{\argmax}{arg\,max}

\newcommand{\nuin}{\nu_i^n}

\newcommand{\dint}{\mathrm{d}}

\begin{document}

\begin{center}

{\bf{\LARGE{CompeteSMoE -- Statistically Guaranteed Mixture of Experts Training via Competition}}}
  
\vspace*{.2in}
{\large{
\begin{tabular}{cccccc}
Nam V. Nguyen$^{\dagger}$ & Huy Nguyen$^{\diamond}$ & Quang Pham$^{\ddagger}$ \\
Van Nguyen$^{\dagger}$ & Savitha Ramasamy$^{\clubsuit}$ & Nhat Ho$^{\diamond}$
\end{tabular}
}}

\vspace*{.2in}

\begin{tabular}{cc}
$^{\dagger}$ FPT Software AI Center \\
$^{\diamond} $ The University of Texas at Austin \\
$^{\ddagger}$ Independent Researcher \\
$^{\clubsuit}$ Institute for Infocomm Research, A${^*}$STAR \\
Correspondence to: \texttt{quangg2012@gmail.com}
\end{tabular}


\vspace*{.2in}

\today
\begin{figure*}[h]
    \centering
    \includegraphics[width=.9\linewidth]{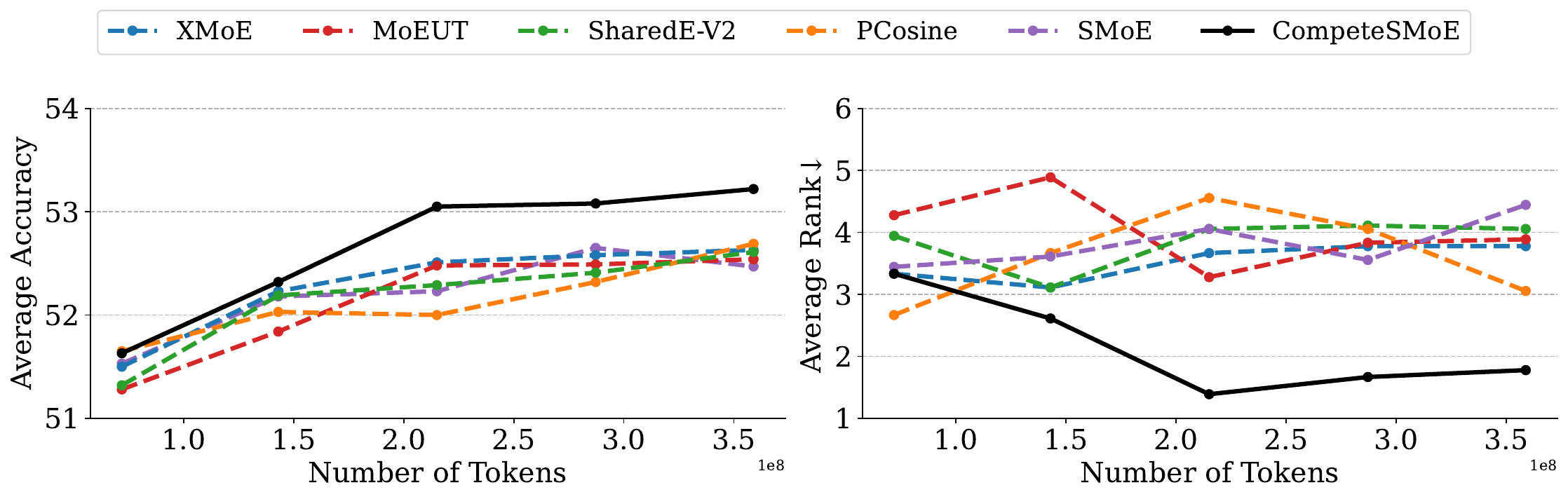}
    \caption{The evolution of zero-shot performance averaged over nine visual instruction tuning tasks throughout training of various SMoE algorithms using a 5.1B parameters backbone.}
    \label{fig:teaser}
\end{figure*}

\begin{abstract}
Sparse mixture of experts (SMoE) offers an appealing solution to scale up the model complexity beyond the mean of increasing the network's depth or width. 
However, we argue that effective SMoE training remains challenging because of the suboptimal routing process where experts that perform computation do not directly contribute to the routing process. 
In this work, we propose \emph{competition}, a novel mechanism to route tokens to experts with the highest neural response. Theoretically, we show that the competition mechanism enjoys a better sample efficiency than the traditional softmax routing. 
Furthermore, we develop CompeteSMoE, a simple yet effective algorithm to train large language models by deploying a router to learn the competition policy, thus enjoying strong performances at a low training overhead. 
Our extensive empirical evaluations on both the visual instruction tuning and language pre-training tasks demonstrate the efficacy, robustness, and scalability of CompeteSMoE compared to state-of-the-art SMoE strategies. We have made the implementation available at: \href{https://github.com/Fsoft-AIC/CompeteSMoE}{\faGithub \hspace{0.5mm}CompeteSMoE}. This work is an improved version of the previous study at \url{https://arxiv.org/abs/2402.02526}.
\end{abstract}
\end{center}

\addtocontents{toc}{\protect\setcounter{tocdepth}{-1}}
\section{Introduction} \label{sec:intro}
Large language models (LLMs) have emerged as a promising architecture for artificial general intelligence. In recent years, LLMs have shown remarkable success in solving many cognitive tasks, ranging from language, vision understanding~\citep{bao_vlmo_2022, gulati_conformer_2020,dosovitskiy_image_2021, ruiz_scaling_2021, bao_beit_2022,li2022blip,li2023blip}, to code generation~\citep{wang2021codet5}, reinforcement learning~\citep{chow_mixture_expert_2023} and life sciences~\citep{rives_biological_2021}. Since the release of the original Transformer model~\citep{vaswani_attention_2017}, extensive efforts have been devoted to scaling the model complexity to take advantage of massive datasets and advanced computing hardware~\citep{radford2019language,brown2020language,du_glam_2022}. To go beyond simply increasing the depth and width of the network, Sparse Mixture-of-experts (SMoE)~\citep{fedus_switch_2022} has emerged as an appealing solution for scaling LLMs. By modularizing the network and activating only subsets of experts per input, SMoE offers constant computational costs when increasing the model complexity, often resulting in improved performance.\\

\noindent
Despite the initial success, practical SMoE training has been known to be notoriously challenging in both engineering and algorithmic aspects. Thus, despite the rapid development of advanced SMoE research in theory and algorithm~\citep{lee_thorp_sparse_2022,riquelme2021scaling,chi_representation_2022}, limited progress has been made in leading industrial models such as DeepSeek~\citep{deepseekv2, deepseekv3} or Phi-MoE~\citep{Abdin2024Phi3TR} as they still implement variants of the vanilla routing mechanism since the original Switch Transformer~\citep{fedus_switch_2022}.
We argue that this discrepancy exists because many state-of-the-art strategies often rely on intuitive conceptualizations, which can only offer greedy solutions that work training in the limited training data and small model regimes. Furthermore, many of existing works~\citep{le2025mixtureexpertsmeetspromptbased, do2023hyperrouter, pertubed_cosine, dai2022stablemoe} still follow the in-domain evaluation and ignores the zero-shot generalization capabilities of pre-train language models, which are their main use cases.\\

\noindent
This work makes a step towards a statistically guaranteed SMoE training strategy that can yield improvements on a wide range of training settings in large-scale models. To this end, we investigate the core mechanism of routing tokens to experts in SMoE, arguing that it could be suboptimal because the experts performing the calculation do not directly contribute to the routing process. This limitation has motivated us to develop a radical routing strategy to distribute tokens to experts more effectively than using the traditional router. To this end, motivated by the Winner-take-all (WTA) principle~\cite{grossberg1982contour} originated in biology~\citep{riesenhuber1999hierarchical,andersen1969participation,eccles2013cerebellum}, we propose the \emph{competition} mechanism for SMoE training. The core mechanism of competition is activating all experts and defining a winning criterion so that tokens are only sent to the winning experts. Thus, competition addresses the fundamental limitation of traditional routing schemes by involving experts into the routing process, which we rigorously show to achieve a better sample efficiency or convergence rate than the traditional softmax routing. Furthermore, we go beyond statistical analysis by developing the CompeteSMoE algorithm that implements the competition mechanism into large-scale models at a modest overhead. Specifically, CompeteSMoE improved the zero-shot performance across 15 common benchmarks in both vision-language finetuning (Figure~\ref{fig:teaser}) and language pre-training settings.\\

\noindent
In summary, our work makes the following contributions. First, we propose a novel \emph{competition} mechanism for training SMoE, which enjoys a better convergence rate than softmax routing. Second, we develop \emph{CompeteSMoE}, a scalable and effective training strategy for SMoE training via competition. Lastly, we conduct extensive experiments to explore the behaviours of CompeteSMoE, including its performance, scalability, convergence property, and routing efficacy. 
\section{CompeteSMoE}
We first recap the foundation of MoE in transformers in Section~\ref{sec:background}. Then, we introduce the competition mechanism in Section~\ref{subsec:router_competion}, discuss the scheduled router training in Section~\ref{subsec:scheduler}, and detail the CompeteSMoE algorithm in Section~\ref{sec:competesmoe}.

\subsection{Background} \label{sec:background}

The traditional SMoE layer~\citep{shazeer_outrageously_2017} consists of a router $\gR(\cdot,W_r)$ parameterized by $W_r$ and $N$ experts $\{ g(\cdot,W_{e_i}) \}_{i=1}^N$ parameterized by $W_{e_i}, i \in [N]$, respectively. The router takes the input token $\vx$ as input and produces an affinity score vector on experts as $\vs_{\gR} = \sigma(\mathrm{TopK}_{-\infty}(\vx^{\top}W_r))$, where $\sigma$ is a scoring function, often implemented as a softmax or sigmoid function. The $\mathrm{TopK}_{-\infty}$ function keeps the largest $K$ elements in a vector and sets the other elements to negative infinity ($-\infty$). With this notation, the SMoE layer takes an input token $\vx$ and calculate the final output by aggregating the outputs of each expert weighted by their affinity scores as:
\begin{equation} \label{eqn:moe}
    \hat{y} = \sum_{i = 1}^N \vs_{\gR}^i  \cdot g(\vx; W_{e_i})
\end{equation}
In practice, it is common for $K$ to be smaller than $N$, i.e. $K < N$, to improve the model efficiency. For completeness, we provide a list of all notations and their meanings in Table~\ref{tab:table_notations}, Appendix~\ref{appendix:notations}.

\begin{figure*}[t]
    \raggedright 
    \begin{subfigure}{0.6\textwidth}
         \raggedright 
         \includegraphics[width=1.1\textwidth]{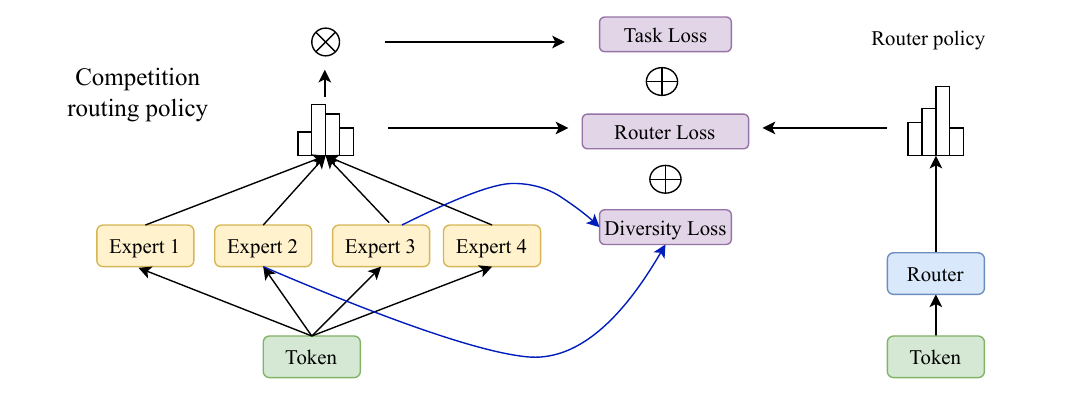}
         \caption{The router learns the competition policy.}
         \label{fig:head}
     \end{subfigure}
     \hfill
     \begin{subfigure}{0.39\textwidth}
         \centering
         \vspace{10pt}
         \includegraphics[width=1.1\textwidth]{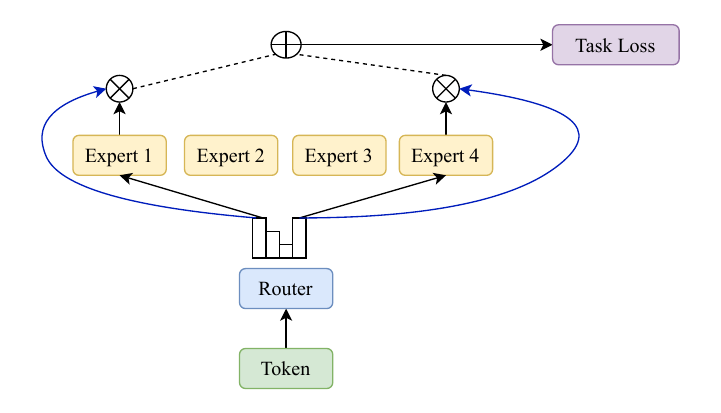}
         \caption{Normal routing using the router.}
         \label{fig:tail}
     \end{subfigure}
     \hfill

     \caption{An illustrative of the interleaved learning phases in CompeteSMoE: (a) activating all experts for the router to learn the competition policy; and (b) normal routing using the router. } \label{fig:algo}
     \vspace{-0.1in}
     
\end{figure*}
\subsection{Routing via Competition}
\label{subsec:router_competion}
We now introduce the \emph{competition} mechanism as an effective routing strategy to facilitate SMoE training. 
The key idea of competition is allowing all experts to calculate their outputs, and selection is performed via the winner-take-all mechanism. Thus, experts will compete with one another and the best ones are selected to calculate the final output. To implement the competition, we propose to use the expert's neural response as its affinity score, i.e. $s_i = \mathbb{E}[\kappa(g(\vx, W_{e_i}))]$, where $\kappa(\cdot)$ is an activation function over the expert's neural responses. In the experiments, we will implement $\kappa$ as the softplus function, unless otherwise stated. However, our competition mechanism and the theoretical analysis thereafter are general and do not make strong assumptions about $\kappa$. We will provide the results of other choices of $\kappa$ in Appendix~\ref{appendix:act_comp}.
With this notation, training of SMoE with competition is formulated via the following steps:
\newpage
\begin{enumerate} \label{eq:competesmoe_algo}
    \item Compute the output of all $N$ experts for a given input $\vx$ as $g(\vx, W_{e_i}), \quad \forall i \in [N].$
    \item Compute the affinity score of each expert:  $\vs_i = \mathbb{E}[\log(1 + e^{g(\vx, W_{e_i})})], \quad \forall i \in [N].$
    \item Select the Top-$K$ experts based on the highest neural response and compute the normalized affinity scores: 
    $\hat{\vs}_{\gC}^i = \mathrm{TopK}_0(s_i, K), \; \vs_{\gC}^i = \frac{\hat{\vs}_{\gC}^i}{\sum_{j = 1}^N \hat{\vs}_{\gC}^j }$. Here, $\mathrm{TopK}_0$ is similar to the traditional $\mathrm{TopK}_{-\infty}$ but sets the values outside the $K$ highest values to be $0$ instead of $-\infty$.
    \item Compute the final output as a weighted sum of the selected experts: \\
    $\hat{y} = \sum_{i=1}^N \vs_{\gC}^i \cdot g(\vz, W_{e_i}).$
\end{enumerate}
Competition starkly contrasts with the standard SMoE implementation discussed in Section~\ref{sec:background} where the affinity score is calculated as the dot product between the input $\vx$ and the experts' embeddings, i.e., columns of $W_r$, and only the few selected experts actually perform their calculation. Although using expert embedding is more efficient, it results in suboptimal routing policies because the embedding is detached from the expert's forward calculation. 
In contrast, competition proposes that experts who respond the strongest to an input are selected to process that input, while suppressing the other experts. We will rigorously show the theoretical guarantees of routing via competition in Section~\ref{sec:understanding}.   
\subsection{Scheduled Training of the Router}  
\label{subsec:scheduler}
One major drawback of competition-based expert selection is the high computational overhead of activating all experts, which limits its viability to large-scale models with billions of parameters. To make competition applicable to LLM training, we propose a scheduled training mechanism that trains a router to learn the competition policy. Thus, a well-trained router is expected to pick experts that would win competition without performing the full competition procedure. Furthermore, using routers is also efficient during inference since it enjoys the same complexity as the original SMoE.
To this end, we employ a learnable router $\gR(\cdot; W_r)$ trained to jointly minimize the task loss and approximate the competition policy. 
Although distilling the competition policy to a router network presents a promising solution for large models, this router should learn the competition policy at a minimal computational overhead. Thus, in the following, we present the router loss for effective training and discuss the router schedulers to ensure that training remains efficient.

\subsubsection{Router Loss}
The router is trained to learn the competition policy and use it to minimize the task loss. We propose to learn the competition policy by minimizing a distillation loss, $\mathcal{L}_{\gD}$, which characterizes the discrepancy between the competition and router policies. For ease of notation, we use $I_{\gC} \subset [N]$ to denote the indices of the experts who won the competition.
Then, the distillation loss $\mathcal{L}_{\gD}$ can be computed by minimizing the mean squared errors (MSE) between the competition and router policies, via their affinity scores as:
\begin{align} \label{eq:router_loss}
    \mathcal{L}_{\gD}({\vs}_{\gR}, {\vs}_{C}) 
    &= \mathrm{MSE}({\vs}_{\gR}, {\vs}_{C}) + \frac{\alpha}{K} \cdot \sum_{j\in I_{\gC}}(\vs_{\gC}^j - \vs_{\gR}^j)^2,
\end{align}
where $\alpha \in \mathbb{R}^{+}$ is a hyperparameter to encourage the router to pay more attention to winning experts from competition. 
\paragraph{Diversity Loss}
One of our main experimental settings is using sparse upcycling~\citep{sparse_upcyling} to bypass the expensive pre-training cost, which allows us to test SMoE algorithms on larger models with a low budget. However, sparse upcycling duplicates the experts and make them have similar outputs, which results in no competition in the early stages of training and limited training efficacy. To mitigate this issue, we introduce the Diversity Loss, $\mathcal{L}_{\text{div}}$, to promote diverse representations of the winning experts.
Formally, given the output matrix \( O \in \mathbb{R}^{K \times D} \) representing the outputs of \( K \) winning experts for an input \( \vx \), the diversity loss is computed as the mean of the off-diagonal elements in the correlation matrix constructed from  \( O\):
\begin{equation} \label{eq:diversity_loss}
    \mathcal{L}_{\text{div}}(O) = \frac{1}{K (K-1)} \sum_{i=1}^{K} \sum_{\substack{j=1 \\ j \neq i}}^{K} C_{i, j}, \; \mathrm{where} \, C = \frac{O \cdot O^\top}{\|O\|^2_2}.
\end{equation}
We apply the Diversity Loss only within the competition mechanism and emphasize the winning experts as defined in Eq.~\ref{eq:competesmoe_algo}, rather than those selected by the router $\gR(\cdot; W_r)$. By penalizing winning experts when they produce similar outputs, $\mathcal{L}_{\text{div}}$ promotes a more effective competition outcome when using the sparse upcycling strategy.

\subsubsection{Router Training Schedule}
Schedulers are essential to ensure that the routers can effectively learn a good routing policy while maintaining a limited computational overhead. In the worst case, when all layers of a deep network perform competition simultaneously, this SMoE becomes dense and could crash the training process. Thus, we need to carefully design a schedule to manage the competition frequency across layers. To this end, we employ two schedulers; one is applied per layer independently, while the other monitors the total competition frequency of all layers.\\

\noindent
For a layer $l$ in a deep network, we first employ a scheduler $\lambda_l(t)$ to determine whether competition should be activated at time step $t$ for this layer. We simply implement $\lambda_l(t)$ by sampling from a Bernoulli distribution with probability $\omega$, which is fixed for all layers. Furthermore, we also employ a global scheduler to monitor the competition frequency across layers. Specifically, we only allow the total number of layers performing competition at any time step to be $A_{\text{max}}$. Any layers exceeding this threshold are deferred to perform competition in the next step. Appendix~\ref{appendix:alcc} will provide a detailed formulation of the global scheduler.

\subsection{The CompeteSMoE Algorithm} \label{sec:competesmoe}
We are now ready to describe the CompeteSMoE algorithm to enhance SMoE training of large-scale models.
Before training, we use the schedulers to generate all time steps for which the competition mechanism is activated at each layer and store them in $\{\Lambda(l)\}_{l=1}^L$, where $\Lambda(l,t)=1$ indicating that the $l-$layer will perform competition at time $t$. Note that this step is performed offline, only one time before training starts. Then, according to the schedule $\Lambda(l,t)$, the training dynamic involves: (i) training the activated experts to minimize the task loss, $\gL_{\mathrm{NLL}}$, and (ii) training the activated router to minimize the task and router losses. We provide an illustration of CompeteSMoE training in Figure~\ref{fig:algo}. Formally, the training step at time $t$ is computed as:
\begin{align}
    W_e^l \gets& W_e^l - \xi_t \frac{\partial}{\partial W_e^l} \gL_{\mathrm{NLL}} (\hat{y}, y), \quad l \in [L] \\
    W_r^l \gets& W_r^l - \xi_t \frac{\partial}{\partial W_r^l} \big[ \gamma \times  \mathcal{L}_{\gD}(\vs_{\gR}, \vs_{C})  +  \beta \times  \mathcal{L}_{\text{div}}(C) + \gL_{\mathrm{NLL}} (\hat{y}, y)\big], \quad \mathrm{if} \, \Lambda(l,t) = 1 
\end{align}
where $\gL_{\mathrm{NLL}}$ is the negative log-likelihood (task loss) between the predicted output $\hat{y}$ and the ground-truth $y$, $\gL_{\gD}$ is the distillation loss defined in equation~(\ref{eq:router_loss}), $\mathcal{L}_{\text{div}}(C)$ is the diversity loss defined in equation~(\ref{eq:diversity_loss}) , $\xi_t$ is the step size. We also wish to emphasize that CompeteSMoE only uses the routers during inference, thus enjoying the same serving cost as the traditional SMoE.\\

\noindent
We now discuss a general guideline to set the hyper-parameters introduced by CompeteSMoE. We recommend the balancing hyper-parameters $\alpha, \beta, \gamma$ to be small values such as $0.01$ or $0.005$. The Bernoulli parameter $\omega$ should also be small (e.g. $0.07$) so that competition is not activated too often. The global scheduler thresholds should be set based on the specific backbone architecture and training infrastructure to ensure stability. We found $A_{\text{max}} = 9$ for vision-language models and $A_{\text{max}} = 3$ for language model pre-training to maximize the memory usage of our hardware. Lastly, we emphasize that the value ranges of these hyper-parameters can be derived by their definition, which greatly reduces the effort for hyper-parameter searching. As long as they follow this guideline, the final performance should be robust to the exact configuration as we will illustrate in Appendix~\ref{appendix:ablation}.
\section{Statistical Guarantee of the Competition Mechanism}
\label{sec:understanding}
In this section, we perform a convergence analysis of Gaussian MoE models equipped with the competition mechanism. Our primary objective is to theoretically justify the effectiveness of the competition mechanism by investigating its sample efficiency in terms of expert estimation.\\

\noindent
\textbf{Problem setting.}
Let $(X_1,Y_1),(X_2,Y_2),\ldots,(X_n,Y_n) \in \gX \times \gY$ be i.i.d samples drawn from bounded subsets $\gX\subset \mathbb{R}^{d_1}$ and $\gY \subset \mathbb{R}$ according to the following conditional density function:
\begin{align}
    \label{eq:density}
    p_{G_*}(Y|X)
    &:=\sum_{i=1}^{N^*}\frac{\exp(\log(1+\exp(g(X,W^*_{e_i}))))}{\sum_{j=1}^{N^*}\exp(\log(1+\exp(g(X,W^*_{e_j}))))}\cdot f(Y|g(X,W^*_{e_i}),\nu^*_i).
\end{align}
Here, $N^*$ is the number of ground-truth experts denoted by $g(X,W^*_{e_i})$, while $f(\cdot|\mu,\nu)$ stands for the Gaussian density with mean $\mu$ and variance $\nu$. In addition, we also define $G_*:=\sum_{i=1}^{N^*}\delta_{(W^*_{e_i},\nu^*_i)}$ as a mixing measure with ground-truth parameters $(W^*_{e_i},\nu^*_i)$, where $\delta$ denotes the Dirac measure. For the sake of theory, we assume that $(W^*_{e_1},\nu^*_1),(W^*_{e_2},\nu^*_2),\ldots,(W^*_{e_{N^*}},\nu^*_{N^*})$ are distinct parameters belonging to a compact space $\Theta\subset\mathbb{R}^{d_2}\times\mathbb{R}_+$ for some $d_2\in\mathbb{N}$.
Next, we assume that the expert function $g(X,W_e)$ is non-zero and differentiable with respect to its parameter $W_e$ for almost surely $X$. Furthermore, for any parameter $W_e\in\mathbb{R}^{d_2}$, if there exists $\alpha_{1}^{(u)},\alpha_{2}^{(uv)},\alpha_3^{(uv)}\in\mathbb{R}$ for $1\leq u,v\leq {d_2}$ such that $\sum_{u=1}^{d_2}\alpha_{1}^{(u)}\frac{\partial g}{\partial W_e^{(u)}}(X,W_e)+\sum_{u,v=1}^{d_2}\alpha_{2}^{(uv)}\frac{\partial^2 g}{\partial W^{(u)}_{e}\partial W^{(v)}_{e}}(X,W_{e})+\sum_{u,v=1}^{d_2}\alpha_{3}^{(uv)}\frac{\partial g}{\partial W^{(u)}_{e}}(X,W_{e})\frac{\partial g}{\partial W^{(v)}_{e}}(X,W_{e})=0$
for almost surely $X$, then we must have $\alpha_{1}^{(u)}=\alpha_2^{(uv)}=\alpha_3^{(uv)}=0$ for all $1\leq u,v\leq {d_2}$. For example, it can be verified that feed-forward networks (FFNs) of the form $g(X,(W_{e,2},W_{e,1},b))=W_{e,2}\mathrm{Softplus}(W_{e,1}^{\top}X+b)$ we used in Section~\ref{subsec:router_competion} satisfy this algebraic independence condition. On the other hand, since linear experts $g(X,(a,b))=a^{\top}X+b$ does not meet this condition, we will conduct a separate convergence analysis for them in Appendix~\ref{appendix:linear_experts}. \\

\noindent
\textbf{Maximum likelihood estimation.} Since the number of ground-truth experts $N^*$ is typically unknown in practice, we fit the model~\eqref{eq:density} with a mixture of $N>N^*$ experts. Then, we estimate the unknown parameters $(W^*_{e_i},\nu^*_i)$, for $1\leq i\leq N$, via estimating the ground-truth mixing measure $G_*$ using the maximum likelihood method as follows:  
\begin{align}
    \label{eq:MLE}
    \widehat{G}_n\in\argmax_{G\in\mathcal{G}_{N}(\Theta)}\frac{1}{n}\sum_{i=1}^{n}\log(p_{G}(Y_i|X_i)),
\end{align}
where we define $\mathcal{G}_{N}(\Theta):=\{G=\sum_{i=1}^{N'}\delta_{(W_{e_i},\nu_i)}: 1\leq N'\leq N, \ (W_{e_i},\nu_i)\in\Theta\}$. 
\begin{proposition}
    \label{prop:density_estimation}
     With the MLE defined in~\eqref{eq:MLE}, the convergence rate of the density estimation $p_{\widehat{G}_n}(Y|X)$ to the ground-truth density $p_{G_*}(Y|X)$ is given by:
    \begin{align*}
        \bbE_X[V(p_{\widehat{G}_n}(\cdot|X),p_{G_*}(\cdot|X))]=\mathcal{O}_P(\sqrt{\log(n)/n}),
    \end{align*}
    Above, we denote $V(p_1,p_2):=\frac{1}{2}\int|p_1-p_2|\dint m$ as the Total Variation distance between two probability density functions $p_1,p_2$ dominated by the Lebesgue measure $m$.
\end{proposition}
\noindent
The proof of Proposition~\ref{prop:density_estimation} can be found in Appendix~\ref{appendix:density_estimation}. The above result indicates that the density estimation $p_{\widehat{G}_n}$ converges to its true counterpart $p_{G_*}$ at a parametric rate of order $\widetilde{\mathcal{O}}_P(n^{-1/2})$. Thus, if we can construct some loss function between two mixing measures $\widehat{G}_n$ and $G_*$, denoted by $\mathcal{L}(\widehat{G}_n,G_*)$, such that $\bbE_X[V(p_{\widehat{G}_n}(\cdot|X),p_{G_*}(\cdot|X))]\gtrsim\mathcal{L}(\widehat{G}_n,G_*)$, then we will obtain parameter and expert estimation rates via the bound $\mathcal{L}(\widehat{G}_n,G_*)=\mathcal{O}_P(\sqrt{\log(n)/n})$. For that purpose, let us introduce the concept of Voronoi loss proposed in Manole et al. \cite{manole22refined}. \\

\noindent
\textbf{Voronoi loss.} For an arbitrary mixing measure $G$, we distribute its atoms to the following Voronoi cells generated by the support points of the ground-truth mixing measure $G_*$:
\begin{align}
    \mathcal{C}_{j}\equiv\mathcal{C}_{j}(G):=\{i\in[N]:\|\theta_i-\theta^*_j\|\leq\|\theta_i-\theta^*_{\ell}\|, \ \forall \ell\neq j\}, \label{eq:Voronoi_cells}
\end{align}
where we denote $\theta_i:=(W_{e_i},\nu_i)$ and $\theta^*_j:=(W^*_{e_j},\nu^*_j)$ for all $i\in[N]$ and $j\in[N^*]$. Here, the cardinality of each Voronoi cell $\mathcal{C}_j$ indicates the number of fitted atoms for the ground-truth atom $\theta^*_j$.
Then, we build a loss function based on these Voronoi cells as follows:
\begin{align}
    \label{eq:Voronoi_loss_over}
    \mathcal{L}_1(G,G_*):=\sum_{j=1}^{N^*}\Big|\sum_{i\in\mathcal{C}_{j}}\exp(c_i)&-\exp(c^*_j)\Big|+\sum_{j\in[N^*]:|\mathcal{C}_{j}|=1}\sum_{i\in\mathcal{C}_{j}}\exp(c_i)\Big[\|W_{e_{i}}-W^*_{e_{j}}\|+|\nu_{i}-\nu^*_{j}|\Big]\nonumber\\
    &+\sum_{j\in[N^*]:|\mathcal{C}_{j}|>1}\sum_{i\in\mathcal{C}_{j}}\exp(c_i)\Big[\|W_{e_{i}}-W^*_{e_{j}}\|^2+|\nu_{i}-\nu^*_{j}|^2\Big].
\end{align}
Given the above Voronoi loss, we are ready to capture the convergence rates of parameter estimation and expert estimation in Theorem~\ref{theorem:parameter_estimation_over} whose proof can be found in Appendix~\ref{appendix:parameter_estimation_over}. 
\begin{theorem}
    \label{theorem:parameter_estimation_over}
    The following lower bound holds for any mixing measure $G\in\mathcal{G}_{N}(\Theta)$:
    \begin{align}
        \label{eq:Hellinger_lower_bound}
        \mathbb{E}_X[V(p_{G}(\cdot|X),p_{G_*}(\cdot|X))]\gtrsim \mathcal{L}_1(G,G_*).
    \end{align}
    This lower bound and the result of Theorem~\ref{prop:density_estimation} imply that $\mathcal{L}_1(\widehat{G}_n,G_*)=\mathcal{O}_P(\sqrt{\log(n)/n})$.
\end{theorem}
A few remarks regarding Theorem~\ref{theorem:parameter_estimation_over} are in order.\\

\noindent
\emph{(i) Expert estimation rates.} From the above results and the formulation of the Voronoi loss $\mathcal{L}_1$, it follows that the rates for estimating exact-specified parameters $W^*_{e_j},\nu^*_j$, i.e., for $j\in[N^*]:|\mathcal{C}_j|=1$, are of parametric order $\widetilde{\mathcal{O}}_P(n^{-1/2})$. Meanwhile, those for over-specified parameters $W^*_{e_j},\nu^*_j$, i.e., for $j\in[N^*]:|\mathcal{C}_j|>1$, are slightly slower, of order $\widetilde{\mathcal{O}}_P(n^{-1/4})$. Since the expert function $g(X,W_e)$ is Lipschitz continuous w.r.t its parameter $W_e$, we have $|g(X,\widehat{W}^n_{e_i})-g(X,W^*_{e_j})|\lesssim\|\widehat{W}^n_{e_i}-W^*_{e_j}\|$ for almost surely $X$. As a result, the estimation rates for exact-specified and over-specified experts $g(X,W^*_{e_j})$ are also of orders $\widetilde{\mathcal{O}}_P(n^{-1/2})$ and $\widetilde{\mathcal{O}}_P(n^{-1/4})$, respectively. Furthermore, we show in Appendix~\ref{appendix:linear_experts} that experts of linear form $g(X,(a,b))=a^{\top}X+b$ also admit these estimation rates. \\

\noindent
\emph{(ii) Sample efficiency of the competition mechanism.} Therefore, we need at most $\mathcal{O}(\epsilon^{-4})$ data points to approximate these experts with a given error $\epsilon>0$. On the other hand, when not using the competition mechanism \cite{nguyen_demystifying_2023}, the convergence rates of expert estimation become significantly slow and decrease when the number of fitted experts increases. For instance, if an expert $g(X,W^*_{e_j})$ is fitted by three experts, i.e., $|\mathcal{C}_j|=3$, then its estimation rate is of order $\widetilde{\mathcal{O}}_P(n^{-1/12})$. Thus, we need much more data points, specifically $\mathcal{O}(\epsilon^{-12})$, to approximate this expert. Consequently, we conclude that the competition mechanism helps improve the sample efficiency in terms of expert estimation.

\section{Related Work} \label{sec:related}

\subsection{Sparse Mixture of Experts}

Mixture of Experts (MoE) is a fundamental model in machine learning~\citep{jacobs_adaptive_1991,jordan_hierarchical_1994} and an instance of the conditional computation framework where different experts are responsible for different regions of the input space~\citep{yuksel_twenty_2012,bengio_deep_2013,masoudnia_mixture_2014,nguyen_practical_2018,nguyen_model_2021}. Extensive efforts have been devoted to establishing a theoretical foundation for MoE, including the universal approximation properties~\citep{norets_approximation_2010,nguyen_universal_2016,nguyen_approximation_2019,nguyen_approximation_2020,nguyen_approximations_2021,nguyen_approximation_2023}, model selection criterion~\citep{khalili_new_2010,montuelle_mixture_2014,nguyen_l_1_oracle_2021,nguyen_non_asymptotic_2022,nguyen_non_asymptotic_2023}, convergence rate for density estimations~\citep{mendes_convergence_2012,norets_adaptive_2021,norets_adaptive_2022} and the problem of parameter estimation~\citep{ho_convergence_2022,nguyen_demystifying_2023,nguyen2024gaussian, Nguyen2024temperature}.
SMoE, the sparse variant of MoE, is more commonly applied to scale large language models~\citep{fedus_switch_2022}. It is often the architecture of choice in many leading industrial models such as Mixtral~\citep{jiang2024mixtralexperts} and DeepSeek~\citep{dai2024deepseekmoe,deepseekv2, deepseekv3}. Within the research community, developing novel routing strategies has been a major focus. Notable strategies include letting experts select tokens~\citep{zhou_mixture_experts_2022}, improving the expert selection process~\citep{lepikhin_gshard_2021,fedus_switch_2022,zuo_taming_2022,chi_representation_2022,dai_stablemoe_2022,chen2023sparse,do2023hyperrouter}, or a global expert assignment scheme\citep{lewis_base_2021,clark_unified_2022}. Despite the promising progress, many such strategies often do not scale well to LLMs with billions of parameters or the language pre-training setting. In contrast, our work goes beyond both the pure theoretical or analytical studies by developing a theoretically-grounded algorithm for effective training of large-scale LLM models. \\

\noindent
Orthogonal to the aforementioned papers, GShard~\citep{lepikhin_gshard_2021} developed an efficient framework to automatically sharding massive SMoE models across many devices. Lastly, sparse upcycling~\citep{sparse_upcyling} duplicated pre-trained models to build an MoE, which bypasses the expensive costs of training from scratch.

\subsection{Competitive Learning}

Competitive learning refers to a framework where computational units compete with one another for the right to response to an input~\citep{mcclelland1987parallel}. Its development is closely related to the biological brain where only certain cells respond strongly to a particular pattern and send suppressive signals to the remaining cells~\citep{andersen1969participation,stefanis1969interneuronal,eccles2013cerebellum}. 
Early investigations of competitive learning showed encouraging results in various learning strategies such as action selection~\citep{feldman1982connectionist}, self-organizing maps~\citep{von1973self,kohonen1982self}, feature discovery~\citep{rumelhart1985feature}, and spiking networks~\citep{oster2005spiking}. Recently, the competition mechanism also motivates the development of various advanced machine learning methods such as maxout networks~\citep{goodfellow2013maxout}, compete to compute~\citep{srivastava2013compete}, and independent mechanisms~\citep{alias2021neural,goyal_recurrent_2021}. Our study establishes a framework to apply competition to SMoE training and develops an algorithm to train large scale SMoE with improved performances at a low training overhead.


\section{Experiment} \label{sec:experiment}

\subsection{Experimental Settings}

\textbf{Training tasks.} We consider two tasks: (i) visual instruction tuning (VIT); and (ii) language pre-training. For VIT, we adopt the LibMoE~\citep{libmoe} framework, which uses sparse upcycling~\citep{sparse_upcyling} to transform an existing dense checkpoint into MoE. Training comprises three phases, where the first two focuses on initializing a dense vision-language connector, and the last phase employs sparse upcycling to compare different MoE algorithms on the LLaVA 1.5 Instruction Tuning dataset~\citep{liu2024improved}.
For the language pre-training task, we follow the MoEUT framework~\citep{moeut} and pretrain on a subset of the SlimPajama dataset~\citep{cerebras2023slimpajama}. 
\\

\noindent
\textbf{Hyper-Parameters.} For the VIT setting, we use Phi3.5 mini~\citep{Abdin2024Phi3TR} as the LLM and SigLiP~\citep{zhai2023sigmoid} as the vision encoder, totaling 5.1B parameters. Additionally, we sparse upcycled the dense models into four experts, and activated two per token. When training SMoE, all methods use the balancing loss~\citep{fedus_switch_2022} and z-loss~\citep{fedus_switch_2022}, and are trained on approximately $10^9$ tokens of the LLaVA 1.5 dataset.
For the language pretraining task, we construct a decoder-only transformer with 151M parameters (16 layers, four attention heads per layer, and a hidden dimension of 512), each SMoE layer consists of 64 experts and eight of which are activated per token ($K=8$). We train this model with a balancing loss~\citep{moeut} on $7\times10^9$ tokens from the SlimPajama dataset. All experiments are conducted on a cluster of 4xH100. Due to the expensive costs of the experiments, we only conducted one run using the same random seeds for all methods. We provide a full description of the training setting in Appendix~\ref{appendix:hyperparam}. 
\\

\noindent
\textbf{Evaluation Benchmarks.}\label{sec:detail_benchmarks} 
All models are evaluated under the zero-shot settings using the well-established benchmarks from the community. For the VIT task, we consider the following benchmarks: AI2D~\citep{Kembhavi2016ADI}, TextVQA~\citep{Singh2019TowardsVM}, GQA~\citep{Hudson2019GQAA}, HallusionBench~\citep{Guan2023HallusionBenchAA}, MathVista (test-mini split)~\citep{Lu2023MathVistaEM}, MMBench (English subset)~\citep{Liu2023MMBenchIY}, MME RealWorld Lite~\citep{mme_realworld}, MMMU Validation~\citep{Yue2023MMMUAM}, MMStar~\citep{Chen2024AreWO}, POPE~\citep{Li2023EvaluatingOH}, and OCRBench~\citep{ocr_bench}. For benchmarks requiring GPT-based evaluation, such as MathVista and HallusionBench, we use GPT-4o,version 2024-08-06. These benchmarks are selected to cover a wide range of capabilities of the model, from perception, reasoning, to assessing hallucination. For the language pre-training task, we consider the LAMBADA~\citep{lambada}, BLiMP~\citep{blimp}, Children’s Book Test (CBT)~\citep{cbt}, HellaSwag~\citep{hellaswag}, PIQA~\citep{pipa}, ARC-Easy~\citep{clark2018think}, and ARC-Challenge~\citep{clark2018think} benchmarks, which are common for the models at our scale.
\newline
\textbf{Baseline.} We compare CompeteSMoE against a suite of state-of-the-art SMoE algorithms. First, SMoE~\citep{fedus_switch_2022}, the original SMoE and still stands strong in today's leading models. Then, we consider activation-based SMoE such as XMoE~\citep{xmoe}, Perturbed Cosine Router (PCosine)~\citep{pertubed_cosine}, and MoEUT~\cite{moeut}, which incorporate cosine similarity or sigmoid activation to improve routing efficiency. Furthermore, inspired by the DeepSeek V2 architecture~\citep{deepseekv2}, we also considered the SharedExpert V2 (SharedE-V2) baseline, which enhances SMoE with one shared expert. Similarly, for the language pretraining task, we also implement the SharedE-V3 baseline, which follows the DeepSeek V3 architecture~\citep{deepseekv3}.
SharedE-V3 replaces the softmax routing in SharedE-V2 with the normalized sigmoid. We implement these two baselines according to the public DeepSeek repository\footnote{https://github.com/deepseek-ai/DeepSeek-V3}.
\subsection{Main Results}

\begin{table}[h]
\centering
\caption{Performance comparison of various SMoE strategies on the VIT setting with a 5B parameters model. \textbf{Bolded} numbers indicate the best result, \uline{underlined} numbers are second-best. $\downarrow$ indicates that lower values are better, and $\uparrow$ indicates that higher values are better.}
\resizebox{\textwidth}{!}{
\begin{tabular}{lccccccccccccc}
\toprule
\textbf{Method} & \textbf{AI2D} & \makecell{\textbf{Text}\\\textbf{VQA}} & \textbf{GQA} & \makecell{\textbf{MM}\\\textbf{Bench}} & \textbf{Hallusion} & \makecell{\textbf{Math}\\\textbf{Vista}} & \textbf{MMMU} & \textbf{MMStar} & \textbf{POPE} & \textbf{OCR} & \makecell{\textbf{MME}\\\textbf{RWL}} & \makecell{\textbf{Avg.}\\\textbf{Acc}}$\uparrow$ & \makecell{\textbf{Avg.}\\\textbf{Rank}}$\downarrow$ \\
\midrule
SMoE~\citep{fedus_switch_2022}          & \uline{65.90} & 41.23 & 60.96 & 70.88 & 39.64 & 31.40 & 42.22 & 40.52 & 86.56 & 32.10 & 31.89 & 49.39 & 4.55 \\
XMoE~\citep{chi_representation_2022}                & 65.19 & 41.14 & 60.63          & 71.31 & \textbf{41.22} & 31.50 & \textbf{42.89} & \textbf{42.60} & 86.12 & 31.30 & \uline{32.51} & \uline{49.67} & 3.50 \\
PCosine~\citep{pertubed_cosine}      & 65.45 & \uline{41.68} & \uline{61.38}          & \uline{71.56} & 40.27 & 30.80 & \uline{42.56} & 41.87 & \uline{86.90} & 30.80 & 32.05 & 49.57 & \uline{3.42} \\ 
MoEUT~\citep{moeut}               & 65.09 & 41.37 & \textbf{61.48}          & 71.39 & \uline{41.01} & \textbf{31.90} & 41.78 & 42.10 & 86.52 & 32.20 & 30.95 & 49.62 & 3.64 \\
SharedE-V2~\citep{deepseekv2}         & 64.93 & 41.53 & 61.15          & 71.05 & 41.20 & 31.20 & 42.56 & 41.44 & 86.08 & \uline{32.40} & 32.36 & 49.63 & 4.05 \\
\midrule
CompeteSMoE                & \textbf{66.22} & \textbf{41.92} & 61.25 & \textbf{72.59} & \textbf{41.22} & \uline{31.70} & 42.00 & \uline{42.25} & \textbf{86.91} & \textbf{33.20} & \textbf{32.52} & \textbf{50.16} & \textbf{1.77} \\
\bottomrule
\end{tabular}
}
\label{tab:moe_vision_language_results}
\end{table}

\begin{table}[h]
\centering
\caption{Performance comparison of various SMoE strategies on the language pre-training setting. \textbf{Bolded} numbers indicate the best results, \uline{underlined} numbers are second best. $\downarrow$ indicates that lower values are better, $\uparrow$ indicates that higher values are better.}
\resizebox{1.\linewidth}{!}{
\begin{tabular}{lcccccccccc}
\toprule
\textbf{MoE} & \textbf{PPL$\downarrow$} & \textbf{LAMBADA} & \textbf{BLiMP} & \textbf{CBT} & \textbf{HellaSwag} & \textbf{PIQA} & \textbf{ARC-E} & \textbf{ARC-C} & \makecell{ \textbf{Avg.}\\\textbf{Acc} }$\uparrow$ & \makecell{\textbf{Avg.}\\\textbf{Rank}}$\downarrow$ \\
\midrule
SMoE~\citep{fedus_switch_2022}       & 13.72 & 25.49 & 76.03 & 75.40          & 29.00          & \textbf{59.09} & 32.94 & 20.94 & 45.56 & 3.94 \\
XMoE~\citep{chi_representation_2022}          & 14.05 & 24.55 & 76.02 & 75.45          & 28.62          & 58.05          & 33.28 & 20.43 & 45.20 & 5.63 \\
PCosine~\citep{pertubed_cosine} & 14.39 & 25.43 & 76.10 & 74.21          & 28.66          & 57.07          & 31.97 & 20.17 & 44.80 & 6.25 \\
MoEUT~\citep{moeut}         & \uline{13.68} & \uline{25.78} & \uline{77.24} & 75.22          & 29.05          & \uline{59.03}  & \uline{33.45} & 20.94 & 45.82 & 2.94 \\
SharedE-V2~\citep{deepseekv2}   & 13.71 & 24.60 & 75.68 & 75.29          & \uline{29.18}  & 58.71          & 32.52 & 20.77 & 45.25 & 4.63 \\
SharedE-V3~\citep{deepseekv3}   & 13.72 & \uline{25.78} & 76.82 & \textbf{75.58} & \textbf{29.30} & 58.49          & 33.40 & \uline{21.97} & \uline{45.91} & \uline{2.88} \\
\midrule
CompeteSMoE          & \textbf{13.66} & \textbf{26.45} & \textbf{77.47} & \uline{75.51} & 29.10 & 58.54 & \textbf{33.74} & \textbf{22.40} & \textbf{46.17} & \textbf{1.75} \\
\bottomrule
\end{tabular}
}
\label{tab:moe_language_results}
\end{table}

\subsubsection{Performance Comparison}
We report the results of the considered algorithms under the VIT and language pre-training in Table~\ref{tab:moe_vision_language_results} and Table~\ref{tab:moe_language_results}, respectively. Overall, we observe that CompeteSMoE offers significant improvements over many benchmarks in both experiments. In addition, CompeteSMoE demonstrated the best performance in many of the challenging and important capabilities such as real-world visual perception and reasoning (MME RWL), reducing visual hallucination (Hallusion and POPE), OCR (OCRBench) and text-only reasoning (ARC-E and ARC-C). Furthermore, we report the evolution of the zero-shot performances on the VIT tasks during training in Figure~\ref{fig:teaser}. The results showed that CompeteSMoE consistently achieved better results than the baselines throughout training, corroborating our theoretical results that the competition mechanism enjoys a better sample efficiency.
\subsubsection{Expert Routing Behavior Analysis} \label{sec:behavior_analysis} 
\paragraph{(a) Evaluating the Effectiveness of Expert Routing.}
\begin{table}[!h]
\centering
\caption{Performance of SMoE and CompeteSMoE when changing top-1 expert to top-(K+1). Numbers in parentheses indicates the changes compared to the original routing results in Table~\ref{tab:moe_vision_language_results}.}
\resizebox{\linewidth}{!}{
\begin{tabular}{
>{\columncolor[HTML]{FFFFFF}}l 
>{\columncolor[HTML]{FFFFFF}}c 
>{\columncolor[HTML]{FFFFFF}}c 
>{\columncolor[HTML]{FFFFFF}}c 
>{\columncolor[HTML]{FFFFFF}}c 
>{\columncolor[HTML]{FFFFFF}}c 
>{\columncolor[HTML]{FFFFFF}}c
>{\columncolor[HTML]{FFFFFF}}c}
\toprule
\textbf{Method}  & \textbf{Text VQA} & \textbf{MMBench} & \textbf{MMMU} & \textbf{MMStar} & \textbf{POPE} & \textbf{OCR Bench} & \textbf{Avg. Change} \\
\midrule
SMoE     
& 41.09~{\scriptsize\textcolor{red}{(-0.14)}} 
& 71.39~{\scriptsize\textcolor{blue}{(+0.52)}} 
& 43.22~{\scriptsize\textcolor{blue}{(+1.00)}} 
& 42.94~{\scriptsize\textcolor{blue}{(+2.42)}} 
& 86.40~{\scriptsize\textcolor{red}{(-0.16)}} 
& 31.50~{\scriptsize\textcolor{red}{(-0.60)}} 
& 0.51 \\
\midrule
CompeteSMoE 
& 41.48~{\scriptsize\textcolor{red}{(-0.45)}} 
& 71.22~{\scriptsize\textcolor{red}{(-1.37)}} 
& 41.67~{\scriptsize\textcolor{red}{(-0.33)}} 
& 40.55~{\scriptsize\textcolor{red}{(-1.70)}} 
& 86.10~{\scriptsize\textcolor{red}{(-0.81)}} 
& 31.70~{\scriptsize\textcolor{red}{(-1.50)}}
& -1.03\\
\bottomrule
\end{tabular}
}
\label{tab:top1_dropout_comparison}
\end{table}

We investigate the experts selection's quality of different policies. To this end, during inference, we replace the expert with the highest affinity score with the expert with the $K+1$ highest score, which is equivalent to shifting the selected experts down by one rank. Table~\ref{tab:top1_dropout_comparison} reports the results of this experiment in the VIT setting. The results show that the SMoE routing policy is clearly suboptimal since selecting a worse expert led to improvements on several benchmarks. On the other hand, CompeteSMoE performances drop in all cases when we deliberately deviate from the router that learned the competition policy. This result shows that CompeteSMoE facilitated a more effective routing policy compared to the traditional SMoE.

\paragraph{(b) Stability of Expert Routing During Training.} \label{para:expert_change_rate}

\begin{figure}[h]
    \centering
    
    \includegraphics[width=1.\linewidth]{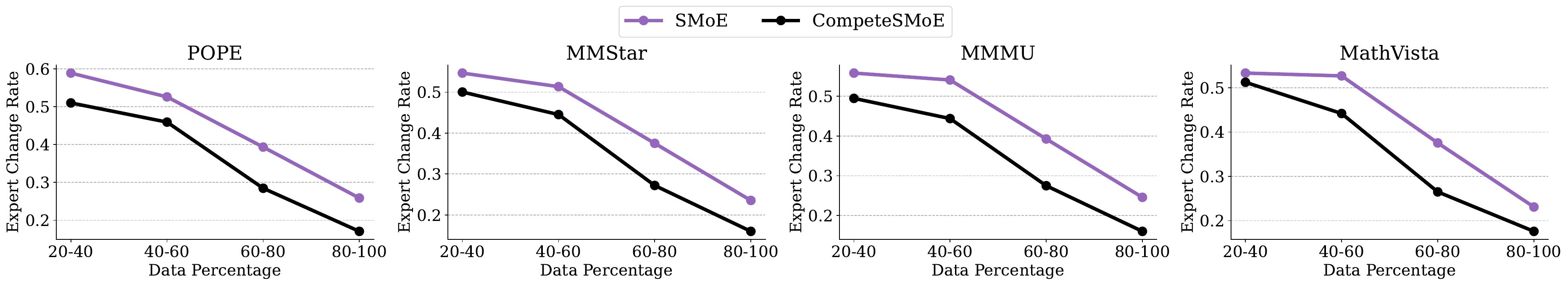}
    \caption{Comparison of expert change rates at different training stages. Lower values are better.}
    \label{fig:experts_change_rate}
\end{figure}
In Figure~\ref{fig:teaser}, we showed that CompeteSMoE achieved a better convergent rates on zero-shot benchmarks than many baselines. We now investigate the convergence rate of the router, showing that CompeteSMoE can quickly find a good routing policy on zero-shot evaluation benchmarks. To this end, we introduce \emph{Expert Change Rate (ECR)} to measure the convergence rate of routers. Specifically, given a dataset $\gD$, we record the expert assignments in all layers for each token in $\gD$ using two model checkpoints at time steps $T$ and $T'$. Then, the ECR of $\gD$ from $T$ to $T'$ is the number of mismatched assignments normalized by all assignments. If the router has converged, then we expect ECR to be low. Otherwise, high ECR values indicate that the router's policy is changing and unstable. Figure~\ref{fig:experts_change_rate} reports the ECR throughout training on four VIT zero-shot benchmarks. We can clearly see that CompeteSMoE has a lower ECR in all cases, suggesting that its routers have a faster convergence rate than SMoE. Together with the better performances as reported in Figure~\ref{fig:teaser} and Table~\ref{tab:moe_vision_language_results}, these experiments corroborate with our theoretical results in Section~\ref{sec:understanding} and showed that CompeteSMoE not only achieved a better sample efficiency, but also the final performance on zero-shot benchmarks.

\subsection{Complexity Analysis}

\begin{table}[h]
\centering
\caption{Computation complexities of various SMoE algorithms.}\label{tab:moe_speed_comparison}
\scriptsize
\setlength\tabcolsep{2.2pt}
\begin{tabular}{lccc}
\toprule
\multirow{2}{*}{Method} & \multirow{2}{*}{Training time} & \multicolumn{2}{c}{Throughput} \\ \cmidrule{3-4}
 &  & Training & Inference \\ \midrule
SMoE & 12h39m & 14.59 & 9.87 \\
XMoE & 13h37m & 13.57 & 8.97 \\
MoEUT & 12h59m & 14.23 & 9.61 \\
PCosine & 13h37m & 13.57 & 8.59 \\
SharedE-V2 & 12h21m & 14.95 & 9.66 \\
CompeteSMoE & 13h01m & 14.18 & 9.88 \\ \bottomrule
\end{tabular}
\end{table}
\noindent
We compare the computational complexities of various methods in Table~\ref{tab:moe_speed_comparison}. We report the wall-clock training time, training throughput, and inference throughput in the VIT setting of the 5.1B model. The results show that CompeteSMoE's training complexity is quite comparable to the standard SMoE, which is only about $3\%$ faster. During inference, CompeteSMoE only uses the simple router, which is exactly the same as SMoE, and is more efficient than cosine similarity-based strategies such as XMoE and PCosine because they introduce additional parameters to the router. In summary, this result shows that CompeteSMoE can effectively leverage competition to improve training with a modest training overhead.
\section{Conclusion} \label{sec:conclusion}

This work proposes competition, a novel strategy to route tokens to experts, and rigorously show that it enjoys a better sample efficiency than softmax routing. Building upon this foundation, we develop CompeteSMoE, an effective algorithm to train large-scale SMoE models with competition at a low computational overhead. Extensive experiments on the visual instruction tuning and language pre-training tasks demonstrate that CompeteSMoE enjoys both a faster convergence rate and final performance on many common zero-shot benchmarks at a minimal overhead.\\

\noindent
Despite achieving encouraging results, CompeteSMoE introduces several hyper-parameters, which may increase the cost for hyper-parameter search. In Section~\ref{sec:competesmoe}, we provided a guideline for hyper-parameter configuration to alleviate this issue.  Algorithmically, CompeteSMoE applies competition on each SMoE layer independently and does not take into account the interactions among experts at different layers. An ideal solution is to perform a graph traversal algorithm through the network depth to determine an optimal expert selection at all layers simultaneously. However, this idea goes beyond the scope of this work, and we will leave it for future studies.

\newpage

\appendix
\addtocontents{toc}{\protect\setcounter{tocdepth}{10}}
\begin{center}
{}\textbf{\Large{Supplement to
``CompeteSMoE -- Statistically Guaranteed Mixture of Experts Training via Competition''}}
\end{center}
This document provides the suppplementary materials for the paper CompeteSMoE -- Statistically Guaranteed Mixture of Experts Training via Competition, and is organized as follows. 
\begin{spacing}{1.3}
\tableofcontents
\end{spacing}
\newpage
\section{Summary of Main Notations} \label{appendix:notations}
\begin{table}[!th]
\centering
\caption{Summary of Main Notations.}
\resizebox{.8\textwidth}{!}{
\begin{tabular}{cl}
\toprule
Symbol & Description \\ 
\midrule
$\mathcal{R}$, $W_r$ & Router network (function) and its parameter\\
$g$, $W_{e}$ & Expert network (function), and its parameter \\
$\vx$ & Input\\
$\vs, \vs_{\gR}, \vs_{\gC}$ & Affinity scores, affinity scores from the router, affinity scores from competition\\
$\mathrm{TopK}_{-\infty}$ & Function retaining the $K$ largest vector elements and setting others to $-\infty$ \\
$\mathrm{TopK}_{0}$ & Function retaining the $K$ largest vector elements and setting others to $0$ \\
$K$ & Number of experts activated per input \\
$N$ & The total number of experts\\
$[M]$ & Set of $\{1, 2, . . . , M\}$ for any positive integer $M$\\
$\hat{y}, y$ & Predicted output, ground truth \\
$t$ & Current $t$-th iteration\\
$T$ & Total number of training steps\\
$l$ & The $l$-th SMoE layer \\
$L$ & Total number of SMoE layers in the model\\
$\kappa$ & Activation function\\
$\sigma$ & Scoring function\\
$\mathbb{E}[\cdot] $ & Mean of vector elements \\
$e$ & Base of the exponential function \\
$I_{\gC}$ & Indices of experts who won in the competition mechanism \\
$\alpha$ & Hyper-parameter prioritizing winning experts in distillation loss\\
$\gamma$ & Hyper-parameter for distillation loss\\
$\beta$ & Hyper-parameter for diversity loss\\
$\omega$ & Bernoulli probability for scheduling competition in each layer\\
$A_{\text{max}}$ & Maximum number of layers that can perform competition on a single time step\\
$\lambda(t)$ & A scheduler determining whether to perform competition at the $t$-th step\\
$\Lambda(l)$ & A vector storing the results of the scheduler $\lambda(t)$ at all time steps of the $l$-th layer \\
$\gL_{\mathrm{NLL}}$ & Negative log-likelihood function (task loss)\\
$\gL_{\gD}$ & Distillation loss\\
$\gL_{div}$ & Diversity loss\\
$\xi_t$ & Step size\\
$\gD$ & A benchmark dataset for evaluation \\ 
$Q_{\text{prev}}$ & Cumulative competition activations over layers $1$ to $l - 1$ \\
$a_n = \mathcal{O}(b_n)$ or $a_{n} \lesssim b_{n}$ & If $a_n \leq C b_n$ for all $ n\in\mathbb{N}$, where $C > 0$ is some universal constant\\
$a_n = \mathcal{O}_P(b_n)$& $\forall\epsilon>0,\exists M>0:$ $\mathbb{P}( A_{n}/b_{n} > M) < \epsilon $ for all sufficiently large $n$\\
$a_n = \widetilde{\mathcal{O}}_P(b_n)$& $a_n = \mathcal{O}_{P}(b_n \log^c(b_n))$, for some $c > 0$.\\
$w^{(u)}$, $w_u$ & The $u$-th entry of a vector $w\in\mathbb{R}^d$\\
$w^z$ & $w^z=w_{1}^{z_{1}}w_{2}^{z_{2}}\ldots w_{d}^{z_{d}}$, for any vector $w \in \mathbb{R}^{d}$ and $z\in\mathbb{N}^d$\\
$|w|$ & $|w|:=w_1+w_2+\ldots+w_d$,  for any vector $w \in \mathbb{R}^{d}$\\ 
$z!$ &$z!:=z_{1}!z_{2}!\ldots z_{d}!$, for any vector $z\in\mathbb{N}^d$\\
$N^*$ & The number of ground-truth experts\\
$f(\cdot|\mu,\nu)$ &Univariate Gaussian density with mean $\mu$ and variance
$\nu$\\
$G_*$ & Ground-truth mixing measure\\
$\delta$ & Dirac measure\\
$m$ & Lebesgue measure\\
$\Theta$ & Parameter space\\
$d_1$ & Dimension of input space\\
$d_2$ & Dimension of expert parameter space\\
$\widehat{G}_n$ & Maximum likelihood estimator for $G_*$\\
$\|\cdot\|,~\|\cdot\|_1$ & $\ell_2$-norm and $\ell_1$-norm value\\ 
$|A|$&
Cardinality of any set $A$\\
$h(p_1,p_2)$ & Hellinger distance $h(p_1,p_2):=\Big(\frac{1}{2}\int(\sqrt{p_1}-\sqrt{p_2})^2d m\Big)^{1/2}$ for any densities $p_1,p_2$\\
$V(p_1,p_2)$ & Total Variation distance $V(p_1,p_2):=\frac{1}{2}\int|p_1-p_2|dm$ for any densities $p_1,p_2$\\
\bottomrule 
\end{tabular}
}

\label{tab:table_notations}
\end{table}

We summarize the main notations used in the main paper in Table~\ref{tab:table_notations}, including those introduced later in the supplementary material.
\newpage
\section{Broader Impact} \label{appendix:broader_impact}
Although our work mostly contributes to the machine learning literature, it also drew inspiration from biology and neuroscience. Specifically, the competition mechanism is rooted in biology, has been studied in neuroscience, and has motivated a few machine learning algorithms. Our work contributed a theoretically grounded algorithm to train large-scale SMoE models, which could potentially push the frontier of the next LLM generation. Lastly, working with large models requires rather costly resources. We took serious precautions during the development of this work, including providing a guideline for hyper-parameter selection, and conducting a single experiment using the same random seed to ensure the results are reliable at a low cost.

\section{Adaptive Layer-wise Competition Control} \label{appendix:alcc}

While scheduled training reduces computational overhead, excessive simultaneous competition activations across multiple SMoE layers can destabilize the training process. To address this, we propose a dynamic mechanism that regulates the number of active competition layers at each training step, enhancing training efficiency. This is achieved by enforcing a global constraint on the maximum number of simultaneously active layers.
\\

\noindent
For a given layer $l$, we compute the cumulative competition activations from all preceding layers (i.e., layers 1 through $l-1$) as:
\begin{equation}
    Q_{\text{prev}} = \sum_{i=1}^{l-1} \Lambda(i),
\end{equation}
where $\Lambda(i) \in \mathbb{R}^T$ denotes the activation state vector of layer $i$ over $T$ training steps, and $Q_{\text{prev}} \in \mathbb{R}^T$ represents the cumulative competition activations up to layer $l-1$.
\\

\noindent
A predefined threshold $A_{\text{max}} \in \mathbb{R}$ governs the total number of active layers permitted per training step. If activating layer $l$ at step $t$ exceeds this threshold i.e., if $Q_{\text{prev}}(t) + \Lambda(l, t) > A_{\text{max}}$ with $\Lambda(l, t) = 1$ we redistribute the activation to an alternative step $t' \neq t$ satisfying:
\begin{equation} \label{eq:alcc_constraint}
    Q_{\text{prev}}(t') + 1 \leq A_{\text{max}}, \quad t' \in \{1, \dots, T\}, \quad \Lambda(l, t') = 0.
\end{equation}
Upon identifying $t'$, we update the activation schedule by setting $\Lambda(l, t') = 1$ and $\Lambda(l, t) = 0$. Empirical results indicate that only 0\% to 7\% of layers are active at any step, ensuring the availability of suitable $t'$ satisfying Eq.~\ref{eq:alcc_constraint}.
\\

\noindent
In summary, this approach dynamically balances competition activations across layers, substantially reducing computational overhead while maintaining training stability for CompeteSMoE.

\section{Effectiveness of Activation Functions in the Competition Mechanism}
\label{appendix:act_comp}

\begin{figure}[H]
    \centering
    
    \includegraphics[width=1.\linewidth]{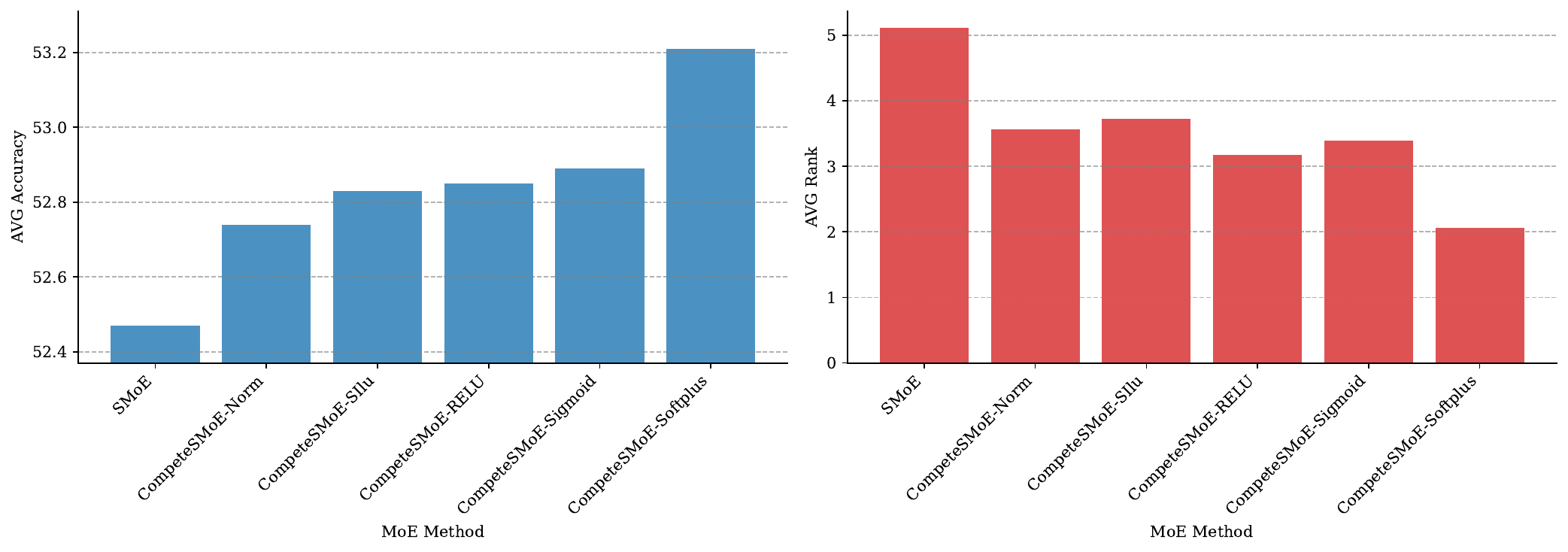}
    \caption{Performance comparison of different activation functions used within the Competition Mechanism over 9 benchmarks.}
    \label{fig:act_comparison}
\end{figure}
In this section, we investigate the impact of different activation functions on the effectiveness of the Competition Mechanism. Specifically, we analyze their role in computing affinity scores, as originally defined in Eq.~\ref{eq:competesmoe_algo}. To support a broader class of activation-based diversity functions, we generalize this formulation by redefining the affinity score as:

\begin{align} \label{eq:aff_score_act}
s_i = \mathbb{E}[\kappa(g(\vx, W_{e_i}))], \quad \forall i \in [N],
\end{align}
This formulation enables the competition mechanism to flexibly incorporate different activation profiles for expert selection.
\\

\noindent
As shown in Figure~\ref{fig:act_comparison}, we compare the performance of several widely used activation functions within the Competition Mechanism, including \texttt{Softplus}, \texttt{SiLU}, \texttt{Sigmoid}, \texttt{ReLU}, and \texttt{Softmax}. Among these, \texttt{Softplus} consistently achieves the highest overall accuracy and ranking. We attribute this to its smooth and well-behaved response across the input domain. Specifically, \texttt{Softplus} softly suppresses negative values while preserving the magnitude of positive values, enabling it to retain useful signal across the entire activation range. This property not only preserves important representational information but also ensures continuous gradient flow, contributing to more stable optimization. In contrast, \texttt{Sigmoid} also suppresses negative values but squashes the entire input range into \([0, 1]\), which can result in significant information loss and vanishing gradients for large magnitude inputs. \texttt{ReLU}, while preserving the magnitude of positive inputs, entirely discards negative values, potentially eliminating informative cues encoded in negative activations.
\\

\noindent
We additionally experimented with an alternative affinity scoring formulation using the exponential function, i.e., $\mathbb{E}[e^{g(\vx, W_{e_i})}]$. However, this led to uncontrolled growth in the output magnitudes, resulting in numerical instability and the emergence of NaN values during training. In contrast, \texttt{Softplus} provides a controlled approximation of the exponential while avoiding such instability, making it more suitable for robust training under the Competition Mechanism.
\\

\noindent
In summary, activation functions that gently suppress negative activations while maintaining linear or near linear behavior for positive inputs such as \texttt{Softplus} are better aligned with the requirements of the Competition Mechanism. Their balanced characteristics lead to more stable expert affinity computation and improved overall performance.

\section{Evaluation of Mean and Norm Strategies for Competition Mechanism}

We conduct an empirical investigation to compare the mean-based strategy, as defined in Eq.~\ref{eq:competesmoe_algo}, with a norm-based formulation. Specifically, we compute the affinity score of expert \( i \) using the L2 norm of its output vector:
\begin{align} \label{eq:aff_score_norm}
s_i = \| g(\vx, W_{e_i}) \|, \quad \forall i \in [N],
\end{align}
As shown in Figure~\ref{fig:act_comparison}, the CompeteSMoE-Norm variant using Equation~\ref{eq:aff_score_norm} yields higher performance compared to the SMoE standard. However, when we switch to the CompeteSMoE-Softplus configuration that employs a mean based strategy, a substantial improvement is observed in both average accuracy and ranking. In conclusion, the mean-based strategy proves to be the most effective setting for expert output aggregation within the Competition Mechanism.

\section{Evaluation of Distillation Loss Effectiveness} \label{appendix:eff_hrloss}

\begin{figure}[h]
    \centering
    \begin{minipage}[t]{0.49\linewidth}
        \centering
        \includegraphics[width=\linewidth]{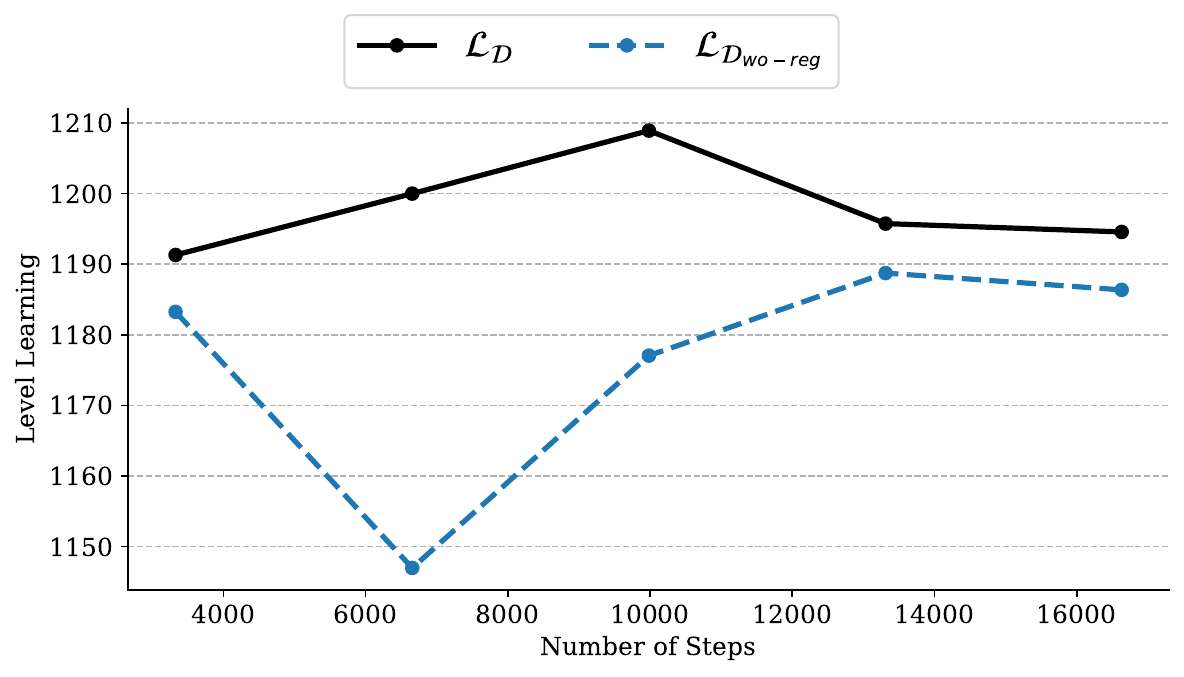}
        \caption{Learning performance of $\mathcal{L}_{\mathcal{D}}$ and $\mathcal{L}_{{\mathcal{D}}_{\text{wo-reg}}}$ measured by the Level Learning metric at every 20\% of training steps on the MMBench-EN benchmark.}
        \label{fig:eff_hrloss}
    \end{minipage}
    \hfill
    \begin{minipage}[t]{0.49\linewidth}
        \centering
        \vspace{-7.5em}
        \captionof{table}{Performance comparison between $\mathcal{L}_{\gD}$ and $\mathcal{L}_{{\gD}_{\text{wo-reg}}}$ across 9 benchmark datasets.}
        \renewcommand{\arraystretch}{1.2}
        \begin{tabular}{lcc}
            \toprule
            \textbf{Loss Function} & \textbf{Avg. Acc }& \textbf{Avg. Rank} \\
            \midrule
            $\mathcal{L}_{{\gD}_{\text{wo-reg}}}$ & 52.92 & 1.78 \\
            $\mathcal{L}_{\gD}$     & \textbf{53.21} & \textbf{1.22} \\
            \bottomrule
        \end{tabular}
        
        \label{tab:hr_vs_router}
    \end{minipage}
\end{figure}
\noindent
In Section~\ref{sec:understanding}, we established the theoretical foundation for the competition mechanism and demonstrated its empirical effectiveness in Table~\ref{tab:moe_vision_language_results}. A key challenge in optimizing the router network is accurately modeling the distribution of competitive routing decisions. We carefully investigated two objective functions: the distillation loss $\mathcal{L}_{\gD}$ (see details in Eq.~\ref{eq:router_loss}) and a variant distillation loss $\mathcal{L}_{{\gD}_{\text{wo-reg}}}$ without the regularization term, which emphasizes penalizing experts who won the competition. We define $\mathcal{L}_{{\gD}_{\text{wo-reg}}}$ as follows:
\begin{align} \label{eq:eff_router_loss}
    \mathcal{L}_{{\gD}_{\text{wo-reg}}}(\vs_{\gR}, \vs_{\gC}) 
    &= \mathrm{MSE}(\vs_{\gR}, \vs_{\gC}) 
\end{align}
Figure~\ref{fig:eff_hrloss} illustrates the progression of the Level Learning (LL) metric, which measures the number of Top-$K$ experts selected by the router network that align with the Top-$K$ experts from the competition mechanism. A high LL value indicates that the router network effectively learns from the competition mechanism, whereas a low value suggests poor learning performance. Notably, $\mathcal{L}_{\gD}$ consistently enables faster and more stable convergence compared to $\mathcal{L}_{{\gD}_{\text{wo-reg}}}$. In particular, during the initial 60\% of training (up to 9,600 steps), $\mathcal{L}_{\gD}$ maintains a clear advantage, effectively mitigating the early performance drop observed with $\mathcal{L}_{{\gD}_{\text{wo-reg}}}$. Moreover, $\mathcal{L}_{\gD}$ achieves a peak LL score of 1210 by 12,000 steps, surpassing the $\mathcal{L}_{{\gD}_{\text{wo-reg}}}$ peak of 1190, and exhibits more stable learning dynamics in later stages.
\\

\noindent
Additionally, quantitative results in Table~\ref{tab:hr_vs_router} further confirm this trend, with $\mathcal{L}_{\gD}$ yielding a higher average accuracy (53.21\% vs. 52.92\%) and a lower average rank (1.22 vs. 1.78) across nine benchmarks. These findings underscore the effectiveness of $\mathcal{L}_{\gD}$ in guiding the router network to better approximate the competition mechanism. Furthermore, they suggest its potential as a preferred optimization objective in competitive MoE architectures.

\section{Further Analysis of Router Behavior}
In this section, we further analyst about router behavior in SMoE and CompeteSMoE.

\FloatBarrier
\begin{figure}[h]
    \centering
    \includegraphics[width=1.\linewidth]{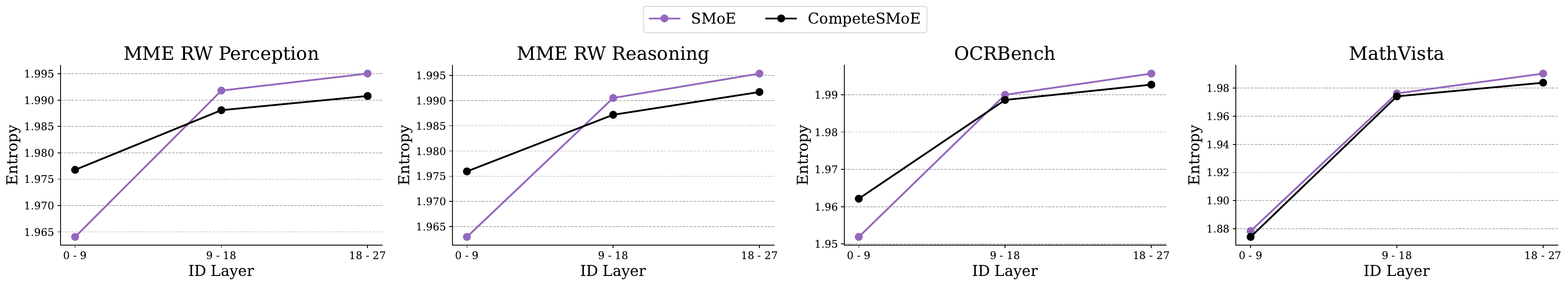}
    \caption{Entropy analysis of expert selection frequency across perception and reasoning tasks. Lower entropy indicates higher specialization in expert routing.}
    \label{fig:experts_dist}
\end{figure}
\noindent
\textbf{(a) Experts distribution on Reasoning and Perception.} As illustrated in Figure~\ref{fig:experts_dist}, we analyze the entropy of expert distribution across layers for SMoE and CompeteSMoE algorithms, evaluated on three benchmarks: MME Real-World Perception and OCR Bench for perception capacity, and MME Real-World Reasoning and MathVista for reasoning capacity. On perception tasks, CompeteSMoE exhibits higher entropy in the early layers, indicating exploratory behavior, but significantly reduces entropy in the middle and final layers. In contrast, on MathVista a benchmark requiring higher-level reasoning CompeteSMoE maintains low entropy in the early and intermediate layers, approaching entropy levels similar to SMoE in the final layers. Both models demonstrate increasing entropy toward the final layers, suggesting more balanced expert allocation as the network deepens, consistent with typical Transformer-based architectures where later layers aggregate information from multiple upstream experts. Regarding the \textbf{representation collapse issue}, both SMoE and CompeteSMoE achieve a high degree of balance in expert distribution, with entropy scores exceeding 1.99 (compared to the maximum entropy of 2 for four experts).
\FloatBarrier
\begin{figure}[h]
    \centering
    \includegraphics[width=1.\linewidth]{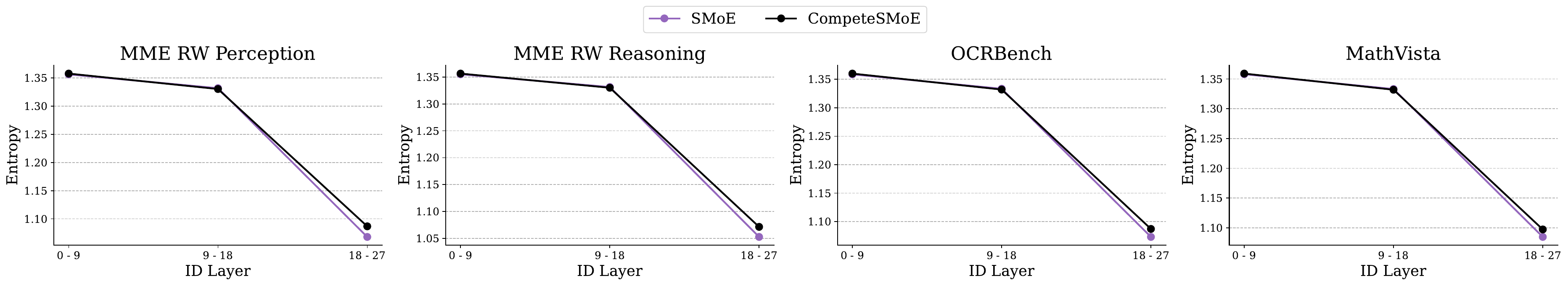}
    \caption{Layer-wise entropy of expert weight distributions for CompeteSMoE and SMoE across three tasks: Real-World Perception, Real-World Reasoning, and Mathematical Reasoning.}
    \label{fig:experts_conf}
\end{figure}
\noindent
\textbf{(b) Effective Expert Aggregation via Weight Distribution.} As shown in Figure~\ref{fig:experts_conf}, we analyze the entropy of expert weight distributions across layers and tasks, which reflects how expert contributions are aggregated. Lower entropy typically suggests more confident expert selection. Both SMoE and CompeteSMoE exhibit decreasing entropy across layers, implying increased decisiveness in expert routing at deeper layers. While SMoE generally maintains lower entropy, especially on MathVista, it tends to concentrate weights heavily on a small subset of experts. In contrast, CompeteSMoE distributes weights more evenly among the selected experts. This balanced aggregation allows CompeteSMoE to better leverage complementary knowledge from multiple experts. Finally, we observe a slight difference between the two models, with both showing a trend toward more confident weight distributions in the final layers.
\section{Ablation Study} \label{appendix:ablation}
We conducted an ablation study on a 5.1B parameter VLM, evaluating performance across various configurations. The best performance was observed with the large-scale model.

\newcommand{\xmark}{\ding{55}}  %
\begin{table}[h]
\centering

\begin{minipage}[t]{0.4\linewidth}
\centering
\caption{Ablation of Competition Mechanism (CM) and Diversity Loss (DL) on 9 benchmarks.}
\resizebox{\linewidth}{!}{
\begin{tabular}{cccc}
\toprule
CM & DL & Avg. Acc \(\uparrow\) & Avg. Rank \(\downarrow\) \\
\midrule
{\color{green!50!black}\checkmark} & {\color{green!50!black}\checkmark} & \textbf{53.21} & \textbf{1.45} \\
{\color{red!70!black}\xmark}        & {\color{green!50!black}\checkmark} & 52.71 & 2.91 \\
{\color{green!50!black}\checkmark} & {\color{red!70!black}\xmark}        & 52.90 & 2.27 \\
{\color{red!70!black}\xmark}        & {\color{red!70!black}\xmark}        & 52.47 & 3.36 \\
\bottomrule
\end{tabular}
}

\label{tab:ablation_competition_diversity}
\end{minipage}
\hfill

\end{table}

\begin{table}[h]
\centering
\begin{minipage}[t]{0.45\linewidth}
\centering
\caption{Ablation study on the activation frequency of the Competition Mechanism (CM) during training.}
\resizebox{0.8\linewidth}{!}{
\begin{tabular}{c c c}
\toprule
$\omega$ & Avg. Acc \(\uparrow\) & Avg. Rank \(\downarrow\) \\
\midrule
3\% & 52.81 & 2.72 \\
5\% & 52.92 & 2.61 \\
7\% & \textbf{53.21} & \textbf{1.83} \\
9\% & 52.82 & 2.83 \\
\midrule
\end{tabular}
}

\label{tab:competition_activation_frequency}
\end{minipage}
\hfill
\begin{minipage}[t]{0.5\linewidth}
\caption{Effect of coefficient \(\alpha\) in Distillation loss.}
\centering
\resizebox{0.8\linewidth}{!}{
\begin{tabular}{ccc}
\toprule
$\alpha$ & Avg. Acc \(\uparrow\) & Avg. Rank \(\downarrow\) \\
\midrule
0.0   & 52.92 & 2.83 \\
0.1 & \textbf{53.21} & \textbf{1.78} \\
0.2 & 52.98 & 2.56 \\
0.3 & 52.87 & 2.83 \\
\midrule
\end{tabular}
}

\label{tab:hybrid_router_theta}
\end{minipage}

\end{table}
\noindent
\textbf{Effect of Component-wise Design on Model Performance.}  
To better understand the individual contributions of the two core components, the Competition Mechanism (CM) and Diversity Loss (DL), we conduct a component-wise ablation study, as shown in Table~\ref{tab:ablation_competition_diversity}, across nine benchmark datasets. Overall, both components independently outperform the SMoE baseline. In detail, removing DL leads to a drop of 0.49\% in average accuracy and an increase of 1.45 in average rank. In contrast, removing CM results in a smaller accuracy drop of 0.30\% and an increase of 0.81 in average rank. These results indicate that the Competition Mechanism contributes more significantly to model performance than Diversity Loss when evaluated in isolation. Notably, combining both components yields the best overall performance, suggesting that CM and DL are complementary. Specifically, DL encourages output diversity among \textbf{won experts}, which in turn enables CM to compute more informative and discriminative affinity scores for expert routing.
\\

\noindent
\textbf{Effect of the Distillation Loss Coefficient \(\alpha\).}  
We investigate the impact of the Distillation Loss coefficient \(\alpha\), which balances the main objective with an auxiliary regularization term. As shown in Figure~\ref{tab:hybrid_router_theta}, setting \(\alpha = 0.1\) achieves the best performance, indicating that a moderate regularization strength provides a useful inductive bias. Increasing \(\alpha\) further leads to performance degradation, suggesting that excessive influence from the auxiliary loss may conflict with the primary learning signal. 
\\

\noindent
\textbf{Analysis of Competition Mechanism Activation Frequency.}  
We further investigate how often the Competition Mechanism (CM) should be activated during training. Table~\ref{tab:competition_activation_frequency} reports model performance when CM is applied at different $\omega$ of training steps. We observe that using a small $\omega$ (e.g., 3\%) leads to suboptimal results, likely due to insufficient competitive pressure. As $\omega$ increases, performance improves, with the best accuracy (53.21\%) and rank (1.83) achieved at 7\%. Notably, increasing $\omega$ beyond this point (e.g., 9\%) offers no further gain and may introduce instability, suggesting a saturation effect. These results highlight that a moderate activation schedule (e.g., $\omega$ in the range of 5–7\%) is sufficient to leverage the benefits of competition while maintaining training stability.

\newpage
\section{Hyperparameter Setting} \label{appendix:hyperparam}
\subsection{Vision Language Model}
\begin{table}[H]
\centering
\small 
\renewcommand{\arraystretch}{1.5} 
\caption{Hyperparameter configurations for three training stages of Phi-3.5 Mini: Pre-Training (PT),  Pre-FineTuning (PFT), and Visual Instruction Tuning (VIT).}
\begin{tabular}{lccc}
\hline
Hyperparameter           & \textbf{PT}     & \textbf{PFT}    & \textbf{VIT}    \\ \hline
Learning rate                    & 1e-3            & 2e-6            & 4e-6            \\
Learning rate schedule           & Cosine          & Cosine          & Cosine          \\
Batch size per GPU               & 64              & 6               & 5               \\
GPUs                             & 4$\times$H100   & 4$\times$H100   & 4$\times$H100   \\
ZeRO optimization                & ZeRO-2          & ZeRO-2          & ZeRO-3          \\
Optimizer                        & AdamW           & AdamW           & AdamW           \\
MLP parameters                   & Trained       & Trained       & Trained       \\
Vision encoders                  & Frozen          & Trained       & Trained       \\
Language model                   & Frozen          & Trained       & Trained       \\
MoE blocks                       & No              & No              & Yes             \\
Balance loss coefficient         & 0.0             & 0.0             & 0.01            \\
Z-loss coefficient               & 0.0             & 0.0             & 0.001           \\
Maximum tokens                   & 2048            & 2048            & 2048            \\ \hline
\end{tabular}

\label{tab:hyperparams_vlm}
\end{table}
\begin{table}[H]
\centering
\caption{Model configuration for the Visual Instruction Tuning (VIT) stage of Phi-3.5 Mini, incorporating a MoE architecture with 4 experts and top-2 expert selection.}
\small 
\renewcommand{\arraystretch}{1.5} 
\begin{tabular}{lcccc}
\hline
\#params & Language Model & Vision Encoder & $N_E$ & $K$ \\ \hline
5.1B               & Phi-3.5 Mini Instruct  & SigLIP-SO400M-Patch14-224 & 4             & 2             \\ \hline
\end{tabular}

\label{tab:model_vlm_config}
\end{table}
As shown in Table~\ref{tab:hyperparams_vlm}, we present the hyperparameter settings for three training stages, following prior work~\citep{libmoe, cumo}. Additionally, Table~\ref{tab:model_vlm_config} details the configurations of the pretrained language model and vision encoder. When training MoE blocks during the VIT stage, only the router network is initialized from scratch. So the weights of the router network are sampled from a normal distribution with a mean of 0 and a standard deviation of 0.02, referenced from the initialization method in the public GPT-2 repository\footnote{https://github.com/openai/gpt-2}, using a fixed random seed of 42 to ensure reproducibility and fairness in validation across algorithms. Overall, we train all MoE algorithms on a large-scale 5.1B-parameter model with identical settings to ensure a fair comparison.
\subsection{Language Model Pretrain}

\begin{table}[H]
\centering
\caption{Model architecture configuration for pretraining the MoE language model. Abbreviations: \texttt{\#params} (total trainable parameters), \texttt{${n}_{layers}$} (transformer layers), \texttt{$d_{expert}$} (expert dimension), \texttt{H} (hidden size), \texttt{$d_{head}$} (attention head dimension), \texttt{$N_E$} (number of experts), \texttt{K} (top-K experts per token), \texttt{$N_{warmup}$} (warmup steps for learning rate).}
\renewcommand{\arraystretch}{1.5}
\begin{tabular}{lcccccc}
\hline
 \#params & ${n}_{layers}$ & $d_{expert}$ & $H$ & $d_{head}$ & $N_E$ & $K$  \\ \hline
151M     & 16      & 128      & 512 & 82      & 64   & 8        \\ \hline

\end{tabular}
\label{tab:language_model_config}
\end{table}

\begin{table}[H]
\centering
\caption{Hyperparameter configuration used for language model pretraining.}
\label{tab:pretrain_lm_hyperparams}
\begin{tabular}{lllllll}
\toprule
\makecell{Learning \\ rate} & 
Schedule & 
\makecell{Batch size \\ / GPU} & 
GPUs & 
Optimizer & 
\makecell{Balance \\ coeff.} & 
\makecell{Z-loss \\ coeff.} \\
\midrule
0.00025 & Cosine & 64 & 4 × H100 & AdamW & 0.01 & 0.00 \\
\bottomrule
\end{tabular}
\end{table}
For pretraining the language model, we leverage the MoEUT~\cite{moeut} framework to train Mixture-of-Experts (MoE)-based architectures. Unlike approaches that employ parameter sharing across layers, we preserve the original Transformer architecture without such sharing. Furthermore, SMoE layers are integrated exclusively into the MLP blocks, leaving the attention modules unmodified, as detailed in Table~\ref{tab:language_model_config}. For hyper-parameters, we adopt the coefficient settings specified in the MoEUT framework. Model weights are initialized according to the MoEUT framework, with a fixed random seed of 42 to ensure reproducibility. Consistent with the MoEUT framework, we exclude the Z-loss term by setting its coefficient to 0, as reported in Table~\ref{tab:pretrain_lm_hyperparams}.

\subsection{Hyperparameter Settings for CompeteSMoE}
\begin{table}[H]
  \centering
  \small 
  \caption{Hyperparameter Configuration for CompeteSMoE on a Large-Scale Model.}
  \label{tab:hyper_competesmoe}
  \begin{tabular}{lcccccc}
    \toprule
    warm up & $\omega$ & $\gamma$ & $\alpha$ & $\beta$ & $A_{\text{max}}$ \\
    \midrule
    0.05 & 0.07 & 0.01 & 0.1 & 0.005 & 9 \\
    \bottomrule
  \end{tabular}
\end{table}

In Table~\ref{tab:hyper_competesmoe}, we present the training configuration for CompeteSMoE on a large-scale model. First, we conduct warm-up training on the SMoE for 5\% of the total training steps to stabilize the router network and experts before initiating the competition mechanism. Second, we set $A_{\text{max}} = 9$, which ensures stable training by preventing excessive simultaneous activation of competitive layers. This hyperparameter should be adjusted based on the number of training steps and the number of SMoE layers. Additionally, the hyper-parameters $\omega$ and $\alpha$ are analyzed in Appendix~\ref{appendix:ablation}. We fix $\beta$ at a small value of $0.005$, set $\gamma$ to a slightly larger value of $0.01$, and use the balanced loss across all training steps.

\section{Training Curves on Vision-Language Benchmarks} \label{appendix:curves}
\begin{figure}[H]
    \centering
    \includegraphics[width=1.\linewidth]{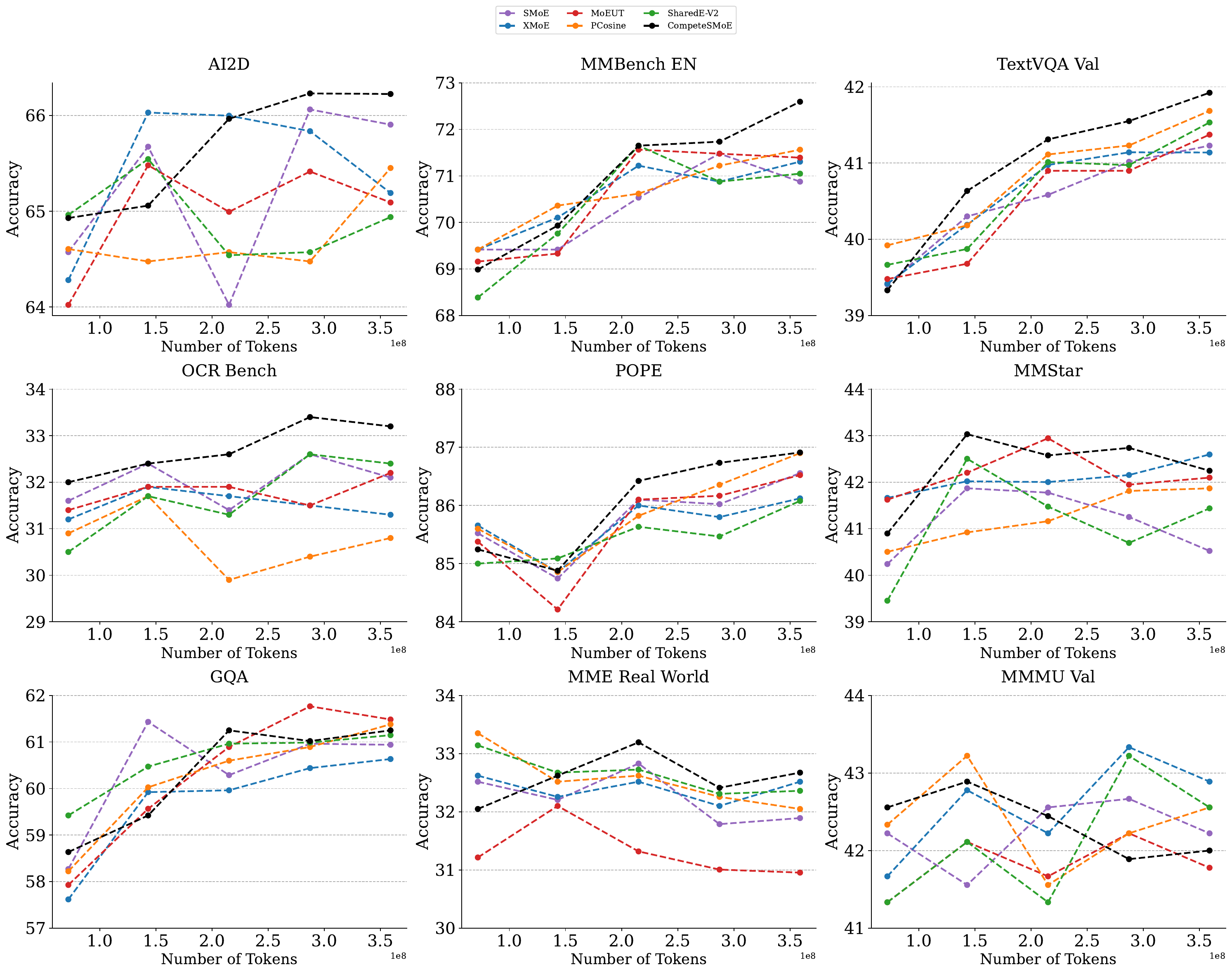}
    \caption{Training curves of CompeteSMoE compared to five advanced MoE algorithms on vision-language benchmarks.}
    \label{fig:curves_comparison}
\end{figure}

In Figure~\ref{fig:curves_comparison}, we include additional training performance curves for 9 benchmarks, supplementing the results presented in Figure~\ref{fig:teaser}.

\section{Additional Theoretical Results}
\label{appendix:linear_experts}
In this appendix, we analyze the convergence behavior of Gaussian mixture of linear experts equipped with the competition mechanism. In particular, we consider experts of the linear form $g(X,(a,b)):=a^{\top}X+b$, where $a\in\mathbb{R}^d$ and $b\in\mathbb{R}$. Then, the conditional density function $p_{G_*}(Y|X)$ in equation~\eqref{eq:density} becomes
\begin{align}
    \label{eq:density_linear}
     p_{G_*}(Y|X)
    &:=\sum_{i=1}^{N^*}\frac{\exp(\log(1+\exp((a^*_i)^{\top}X+b^*_i)))}{\sum_{j=1}^{N^*}\exp(\log(1+\exp((a^*_j)^{\top}X+b^*_j)))}\cdot f(Y|(a^*_i)^{\top}X+b^*_i,\nu^*_i).
\end{align}
Our ultimate goal is to compare the sample efficiency of this model to that without the competition mechanism \citep{nguyen_demystifying_2023} in terms of expert estimation. For that purpose, we use a Voronoi loss tailored to the setting of linear experts, which is given by
\begin{align}
    \label{eq:Voronoi_loss_linear}
    \mathcal{L}_{2}(G,G_*):&=\sum_{j=1}^{N^*}\Big|\sum_{i\in\mathcal{C}_{j}}\exp(c_i)-\exp(c^*_j)\Big|\nonumber\\
    &+\sum_{j\in[N^*]:|\mathcal{C}_{j}|=1}\sum_{i\in\mathcal{C}_{j}}\exp(c_i)\Big[\|a_{i}-a^*_{j}\|+|b_i-b^*_j|+|\nu_{i}-\nu^*_{j}|\Big]\nonumber\\
    &+\sum_{j\in[N^*]:|\mathcal{C}_{j}|=1}\sum_{i\in\mathcal{C}_{j}}\exp(c_i)\Big[\|a_{i}-a^*_{j}\|^2+|b_i-b^*_j|^2+|\nu_{i}-\nu^*_{j}|^2\Big].
\end{align}
Equipped with the above Voronoi loss, we establish the convergence rate of parameter and expert estimations in the Gaussian mixture of linear experts with the competition in Theorem~\ref{theorem:parameter_estimation_linear}.
\begin{theorem}
    \label{theorem:parameter_estimation_linear}
    The following lower bound holds for any mixing measure $G\in\mathcal{G}_{N}(\Theta)$:
    \begin{align}
        \label{eq:Hellinger_lower_bound_linear}
        \mathbb{E}_X[V(p_{G}(\cdot|X),p_{G_*}(\cdot|X))]\gtrsim \mathcal{L}_{2}(G,G_*).
    \end{align}
    This lower bound indicates that $\mathcal{L}_{2}(\widehat{G}_n,G_*)=\mathcal{O}_P(\sqrt{\log(n)/n})$.
\end{theorem}
The proof of Theorem~\ref{theorem:parameter_estimation_linear} can be found in Appendix~\ref{appendix:parameter_estimation_linear}. A few remarks regarding the results of this theorem are in order.

\emph{(i) Parameter estimation rates.} 
The bound of the Voronoi loss $\mathcal{L}_2(\widehat{G}_n,G_*)$ in Theorem~\ref{theorem:parameter_estimation_linear} reveals that the estimation rates for exact-specified parameters $a^*_{j},b^*_j,\nu^*_j$, i.e., for $j\in[N^*]:|\mathcal{C}_j|=1$, are of parametric order $\widetilde{\mathcal{O}}_P(n^{-1/2})$, whereas those for their over-specified counterparts, i.e., for $j\in[N^*]:|\mathcal{C}_j|>1$, are slightly slower, of order $\widetilde{\mathcal{O}}_P(n^{-1/4})$. 

\emph{(ii) Expert estimation rates.} Note that the input space is bounded, then we have
\begin{align*}
    \Big|(\widehat{a}^n_i)^{\top}X+\widehat{b}^n_i-(a^*_j)^{\top}X-b^*_j\Big|\lesssim\|\widehat{a}^n_{i}-a^*_{j}\|+|\widehat{b}^n_i-b^*_j|,
\end{align*}
for almost surely $X$. Consequently, the estimation rates for exact-specified and over-specified experts $(a^*_j)^{\top}X+b^*_j$ are also of orders $\widetilde{\mathcal{O}}_P(n^{-1/2})$ and $\widetilde{\mathcal{O}}_P(n^{-1/4})$, respectively. 

\emph{(iii) Sample efficiency of the competition mechanism.} Thus, we need polynomially many data points $\mathcal{O}(\epsilon^{-4})$ to estimate these linear experts with a given error $\epsilon>0$. By contrast, when not using the competition mechanism \cite{nguyen_demystifying_2023}, the linear expert estimation rates are substantially slowed down since they hinge on the solvability of some complex system of polynomial equations and are decelerated as the number of fitted experts grows. For example, if a linear expert $(a^*_j)^{\top}X+b^*_j$ is fitted by two experts (or three experts), that is, $|\mathcal{C}_j|=2$ (or $|\mathcal{C}_j|=3$), then the rate for estimating this linear expert is of order $\widetilde{\mathcal{O}}_P(n^{-1/8})$  (or $\widetilde{\mathcal{O}}_P(n^{-1/12})$). Therefore, we need $\mathcal{O}(\epsilon^{-8})$ (or $\mathcal{O}(\epsilon^{-12})$), to estimate this expert. For that reason, we claim that the Gaussian MoE becomes more sample-efficient when equipped with the competition mechanism.

\section{Proof of Theoretical Results}\label{appendix:theory}

\subsection{Proof of Theorem~\ref{theorem:parameter_estimation_over}}
\label{appendix:parameter_estimation_over}

In this proof, we aim to demonstrate that the following lower bound holds for any $G\in\mathcal{G}_{N}(\Theta)$:
\begin{align}
    \label{eq:target_bound_over}
    \bbE_X[V(p_{G}(\cdot|X),p_{G_*}(\cdot|X))]\gtrsim \mathcal{L}_1(G,G_*).
\end{align}
For that purpose, we first establish the local part of the above bound, that is,
\begin{align}
    \label{eq:local_part_over}
    \lim_{\varepsilon\to0}\inf_{G\in\mathcal{G}_{N}(\Theta):\mathcal{L}_1(G,G_*)\leq\varepsilon}\dfrac{\bbE_X[V(p_{G}(\cdot|X),p_{G_*}(\cdot|X))]}{\mathcal{L}_1(G,G_*)}>0.
\end{align}
This local part implies that there exists a positive constant $\varepsilon'$ that satisfies
\begin{align*}
    \inf_{G\in\mathcal{G}_{N}(\Theta):\mathcal{L}_1(G,G_*)\leq\varepsilon'}\dfrac{\bbE_X[V(p_{G}(\cdot|X),p_{G_*}(\cdot|X))]}{\mathcal{L}_1(G,G_*)}>0.
\end{align*}
Then, it is sufficient to derive the following global part of the bound in~\eqref{eq:target_bound_over}:
\begin{align}
    \label{eq:global_part_over}
    \inf_{G\in\mathcal{G}_{N}(\Theta):\mathcal{L}_1(G,G_*)>\varepsilon'}\dfrac{\bbE_X[V(p_{G}(\cdot|X),p_{G_*}(\cdot|X))]}{\mathcal{L}_1(G,G_*)}>0.
\end{align}
\textbf{Local part:} In this part, we will establish the local part in equation~\eqref{eq:local_part_over} using the proof by contradiction method.

Suppose that the local part is not true, then we can find a sequence of mixing measures $(G_n)$ given by $G_n:=\sum_{i=1}^{N}\exp(c^n_i)\delta_{(W^n_{e_i},\nuin)}\in\mathcal{G}_{N}(\Theta)$ such that $\mathcal{L}_1(G_n,G_*)\to0$ and
\begin{align*}
    \mathbb{E}_X[V(p_{G_n}(\cdot|X),p_{G_*}(\cdot|X))]/\mathcal{L}_1(G_n,G_*)\to0,
\end{align*}
as $n\to\infty$. As we use asymptotic arguments in this proof, we may assume without loss of generality (WLOG) that the Voronoi cells $\mathcal{C}^n_j:=\mathcal{C}_j(G_n)$ is independent of the sample size $n$. Then, the Voronoi loss of interest turns into
\begin{align}
    \label{eq:Voronoi_loss_assume}
    \mathcal{L}_1(G_n,G_*):=\sum_{j=1}^{N^*}\Big|\sum_{i\in\mathcal{C}_{j}}&\exp(c^n_i)-\exp(c^*_j)\Big|+\sum_{j\in[N^*]:|\mathcal{C}_{j}|=1}\sum_{i\in\mathcal{C}_{j}}\exp(c^n_i)\Big[\|W^n_{e_{i}}-W^*_{e_{j}}\|+|\nu^n_{i}-\nu^*_{j}|\Big]\nonumber\\
    &+\sum_{j\in[N^*]:|\mathcal{C}_{j}|=1}\sum_{i\in\mathcal{C}_{j}}\exp(c^n_i)\Big[\|W^n_{e_{i}}-W^*_{e_{j}}\|^2+|\nu^n_{i}-\nu^*_{j}|^2\Big].
\end{align}
Since $\mathcal{L}_1(G_n,G_*)\to0$ as $n\to\infty$, we have $(W^n_{e_i},\nu^n_i)\to (W^*_{e_j},\nu^*_j)$  for all $j\in[N^*]$ and $i\in\mathcal{C}_{j}$. 

Subsequently, we divide the rest of this proof into three main steps.

\textbf{Step 1: Taylor expansion.} In this step, we aim to decompose the term $T_n(Y|X):=\Big[\sum_{j=1}^{N^*}\exp(\log(1+\exp(g(x,W^*_{e_j}))))\Big]\cdot [p_{G_n}(Y|X)-p_{G_*}(Y|X)]$ can be decomposed as
\begin{align*}
    T_n(Y|X)&=\sum_{j=1}^{N^*}\sum_{i\in\mathcal{C}_{j}}\exp(c^n_i)\Big[\exp(\log(1+\exp(g(X,W^n_{e_i}))))f(Y|g(X,W^n_{e_i}),\nu^n_i)\\
    &\hspace{2cm} -\exp(\log(1+\exp(g(X,W^*_{e_j}))))f(Y|g(X,W^*_{e_j}),\nu^*_j)\Big]\\
    &-\sum_{j=1}^{N^*}\sum_{i\in\mathcal{C}_{j}}\exp(c^n_i)\Big[\exp(\log(1+\exp(g(X,W^n_{e_i}))))-\exp(\log(1+\exp(g(X,W^*_{e_j}))))]p_{G_n}(Y|X)\\
    &+\sum_{j=1}^{N^*}\Big[\sum_{i\in\mathcal{C}_{j}}\exp(c^n_i)-\exp(c^*_j)\Big]\cdot\exp(\log(1+\exp(g(X,W^*_{e_j}))))[f(Y|g(X,W^*_{e_j}),\nu^*_j)-p_{G_n}(Y|X)]\\
    &:=T_{n,1}(Y|X)-T_{n,2}(Y|X)+T_{n,3}(Y|X).
\end{align*}
Next, we continue to decompose the term $T_{n,1}(Y|X)$ as
\begin{align*}
    T_{n,1}(Y|X)&=\sum_{j\in[N^*]:|\mathcal{C}_{j}|=1}\sum_{i\in\mathcal{C}_{j}}\exp(c^n_i)\Big[\exp(\log(1+\exp(g(x,W^n_{e_i}))))f(Y|g(X,W^n_{e_i}),\nu^n_i)\\
    &\hspace{3cm} -\exp(\log(1+\exp(g(x,W^*_{e_j}))))f(Y|g(X,W^*_{e_j}),\nu^*_j)\Big]\\
    &+\sum_{j\in[N^*]:|\mathcal{C}_{j}|>1}\sum_{i\in\mathcal{C}_{j}}\exp(c^n_i)\Big[\exp(\log(1+\exp(g(x,W^n_{e_i}))))f(Y|g(X,W^n_{e_i}),\nu^n_i)\\
    &\hspace{3cm} -\exp(\log(1+\exp(g(x,W^*_{e_j}))))f(Y|g(X,W^*_{e_j}),\nu^*_j)\Big]\\
    &:=T_{n,1,1}(Y|X)+T_{n,1,2}(Y|X).
\end{align*}
Let us denote $F_{\rho}(Y|X;W_e,\nu):=\exp(\log(1+\exp(g(X,W_e))))\frac{\partial^{\rho}f}{\partial g^{\rho}}(Y|g(X,W_e),\nu)$. By applying the first-order Taylor expansion to the function $F_{0}(Y|X;W_e,\nu)$ around the point $(W^*_{e_j},\nu^*_j)$, we rewrite the term $T_{n,1,1}(Y|X)$ as
\begin{align*}
    &T_{n,1,1}(Y|X)=\sum_{j\in[N^*]:|\mathcal{C}_{j}|=1}\sum_{\rho=0}^{2}T_{n,1,1,\rho}^{(j)}(X)F_{\rho}(Y;X,W^*_{e_j},\nu^*_j)+R_{n,1,1}(Y|X),
\end{align*}
where $R_{n,1,1}(Y|X)$ is the Taylor remainder such that $R_{n,1,1}(Y|X)/\mathcal{L}_1(G_n,G_*)\to0$ as $n\to\infty$, and
\begin{align*}
    T_{n,1,1,0}^{(j)}(X)&:=\sum_{i\in\mathcal{C}_{j}}\exp(c^n_i)\sum_{u=1}^{d_2}(\Delta W^n_{e_{ij}})^{(u)}\frac{\partial g}{\partial W^{(u)}_{e}}(X,W^*_{e_j})\cdot\frac{1}{1+\exp(-g(X,W^*_{e_j}))},\\
    T_{n,1,1,1}^{(j)}(X)&:=\sum_{i\in\mathcal{C}_{j}}\exp(c^n_i)\sum_{u=1}^{d_2}(\Delta W^n_{e_{ij}})^{(u)}\frac{\partial g}{\partial W^{(u)}_{e}}(X,W^*_{e_j}),\\
    T_{n,1,1,2}^{(j)}(X)&:=\sum_{i\in\mathcal{C}_{j}}\frac{1}{2}\exp(c^n_i)(\Delta\nu^n_{ij}),
\end{align*}
in which $\Delta W^n_{e_{ij}}:=W^n_{e_i}-W^*_{e_j}$ and $\Delta\nu^n_{ij}:=\nu^n_i-\nu^*_j$.

Meanwhile, by means of the second-order Taylor expansion, the term $T_{n,1,2}(Y|X)$ can be represented as
\begin{align*}
    &T_{n,1,2}(Y|X)=\sum_{j\in[N^*]:|\mathcal{C}_{j}|>1}\sum_{\rho=0}^{4}T_{n,1,2,\rho}^{(j)}(X)F_{\rho}(Y;X,W^*_{e_j},\nu^*_j)+R_{n,1,2}(Y|X),
\end{align*}
where $R_{n,1,2}(Y|X)$ is the Taylor remainder such that $R_{n,1,2}(Y|X)/\mathcal{L}_1(G_n,G_*)\to0$ as $n\to\infty$, and
\begin{align*}
     T_{n,1,2,0}^{(j)}(X)&:=\sum_{i\in\mathcal{C}_{j}}\exp(c^n_i)\Bigg[\sum_{u=1}^{d_2}(\Delta W^n_{e_{ij}})^{(u)}\cdot\frac{\frac{\partial g}{\partial W^{(u)}_{e}}(X,W^*_{e_j})}{1+\exp(-g(X,W^*_{e_j}))}\\
     &+\sum_{u,v=1}^{d_2}\frac{(\Delta W^n_{e_{ij}})^{(u)}(\Delta W^n_{e_{ij}})^{(v)}}{1+1_{\{u=v\}}}\cdot\frac{\frac{\partial^2 g}{\partial W^{(u)}_{e}\partial W^{(v)}_{e}}(X,W^*_{e_j})+\frac{\partial g}{\partial W^{(u)}_{e}}(X,W^*_{e_j})\frac{\partial g}{\partial W^{(v)}_{e}}(X,W^*_{e_j})}{1+\exp(-g(X,W^*_{e_j}))}\Bigg],\\
    T_{n,1,2,1}^{(j)}(X)&:=\sum_{i\in\mathcal{C}_{j}}\exp(c^n_i)\Bigg[\sum_{u=1}^{d_2}(\Delta W^n_{e_{ij}})^{(u)}\frac{\partial g}{\partial W^{(u)}_{e}}(X,W^*_{e_j})\\
    &+\sum_{u,v=1}^{d_2}\frac{(\Delta W^n_{e_{ij}})^{(u)}(\Delta W^n_{e_{ij}})^{(v)}}{1+1_{\{u=v\}}}\Bigg(\frac{2\frac{\partial g}{\partial W^{(u)}_{e}}(X,W^*_{e_j})\frac{\partial g}{\partial W^{(v)}_{e}}(X,W^*_{e_j})}{1+\exp(-g(X,W^*_{e_j}))}+\frac{\partial^2 g}{\partial W^{(u)}_{e}\partial W^{(v)}_{e}}(X,W^*_{e_j})\Bigg)\Bigg],\\
    T_{n,1,2,2}^{(j)}(X)&:=\sum_{i\in\mathcal{C}_{j}}\exp(c^n_i)\Bigg[\frac{1}{2}(\Delta\nu^n_{ij})+\sum_{u,v=1}^{d_2}\frac{(\Delta W^n_{e_{ij}})^{(u)}(\Delta W^n_{e_{ij}})^{(v)}}{1+1_{\{u=v\}}}\cdot\frac{\partial g}{\partial W^{(u)}_{e}}(X,W^*_{e_j})\frac{\partial g}{\partial W^{(v)}_{e}}(X,W^*_{e_j})\\
    &+\sum_{u=1}^{d_2}(\Delta W^n_{e_{ij}})^{(u)}(\Delta \nu^n_{ij})\cdot\frac{1}{2}\frac{\frac{\partial g}{\partial W^{(u)}_{e}}(X,W^*_{e_j})}{1+\exp(-g(X,W^*_{e_j}))}\Bigg],\\
    T_{n,1,2,3}^{(j)}(X)&:=\sum_{i\in\mathcal{C}_{j}}\exp(c^n_i)\sum_{u=1}^{d_2}\frac{1}{2}(\Delta W^n_{e_{ij}})^{(u)}(\Delta\nu^n_{ij})\frac{\partial g}{\partial W^{(u)}_{e}}(X,W^*_{e_j}),\\
    T_{n,1,2,4}^{(j)}(X)&:=\sum_{i\in\mathcal{C}_{j}}\exp(c^n_i)\cdot\frac{1}{4}(\Delta\nu^n_{ij})^2.
\end{align*}
Next, we decompose the term $T_{n,2}(Y|X)$ as
\begin{align*}
    &T_{n,2}(Y|X)\\
    &:=\sum_{j\in[N^*]:|\mathcal{C}_{j}|=1}\sum_{i\in\mathcal{C}_{j}}\exp(c^n_i)\Big[\exp(\log(1+\exp(g(X,W^n_{e_i}))))-\exp(\log(1+\exp(g(X,W^*_{e_j}))))]p_{G_n}(Y|X)\\
    &+\sum_{j\in[N^*]:|\mathcal{C}_{j}|>1}\sum_{i\in\mathcal{C}_{j}}\exp(c^n_i)\Big[\exp(\log(1+\exp(g(X,W^n_{e_i}))))-\exp(\log(1+\exp(g(X,W^*_{e_j}))))]p_{G_n}(Y|X)\\
    &:=T_{n,2,1}(Y|X)+T_{n,2,2}(Y|X).
\end{align*}
Note that we can rewrite the term $T_{n,1,2}(Y|X)$ using the first-order Taylor expansion to the function $\exp(\log(1+\exp(g(W^n_{e_i}))))$ around the point $W^*_{e_j}$ as
\begin{align*}
    T_{n,2,1}(Y|X)=\sum_{j\in[N^*]:|\mathcal{C}_{j}|=1}\sum_{i\in\mathcal{C}_{j}}\exp(c^n_i)\sum_{u=1}^{d_2}(\Delta W^n_{e_{ij}})^{(u)}\cdot\frac{\frac{\partial g}{\partial W^{(u)}_{e}}(X,W^*_{e_j})}{1+\exp(-g(X,W^*_{e_j}))}H_n(Y|X;W^*_{e_j}) \\
    + R_{n,2,1}(Y|X),
\end{align*}
where we denote $H_n(Y|X;W_{e})=\exp(\log(1+\exp(g(X,W_{e}))))p_{G_n}(Y|X)$ and $R_{n,2,1}(Y|X)$ is the Taylor remainder such that $R_{n,2,1}(Y|X)/\mathcal{L}_1(G_n,G_*)\to0$ as $n\to\infty$.

On the other hand, by means of the second-order Taylor expansion, we have
\begin{align*}
    &T_{n,2,2}(Y|X)=\sum_{j\in[N^*]:|\mathcal{C}_{j}|>1}\sum_{i\in\mathcal{C}_{j}}\exp(c^n_i)\Bigg[\sum_{u=1}^{d_2}(\Delta W^n_{e_{ij}})^{(u)}\cdot\frac{\frac{\partial g}{\partial W^{(u)}_{e}}(X,W^*_{e_j})}{1+\exp(-g(X,W^*_{e_j}))}\\
    &+\sum_{u,v=1}^{d_2}\frac{(\Delta W^n_{e_{ij}})^{(u)}(\Delta W^n_{e_{ij}})^{(v)}}{1+1_{\{u=v\}}}\cdot\frac{\frac{\partial^2 g}{\partial W^{(u)}_{e}\partial W^{(v)}_{e}}(X,W^*_{e_j})+\frac{\partial g}{\partial W^{(u)}_{e}}(X,W^*_{e_j})\frac{\partial g}{\partial W^{(v)}_{e}}(X,W^*_{e_j})}{1+\exp(-g(X,W^*_{e_j}))}\Bigg]H_n(Y|X;W^*_{e_j}) \\
    &\hspace{12cm}+ R_{n,2,2}(Y|X),
\end{align*}
where $R_{n,2,1}(Y|X)$ is the Taylor remainder such that $R_{n,2,2}(Y|X)/\mathcal{L}_1(G_n,G_*)\to0$ as $n\to\infty$.

From the above equation, $[T_{n,1,1}(Y|X)-R_{n,1,1}(Y|X)]$, $[T_{n,1,2}(Y|X)-R_{n,1,2}(Y|X)]$, $[T_{n,2,1}(Y|X)-R_{n,2,1}(Y|X)]$, $[T_{n,2,2}(Y|X)-R_{n,2,2}(Y|X)]$ and $[T_{n,3}(Y|X)]$ can be seen as a combination of elements of the set $\mathcal{S}:=\bigcup_{j=1}^{N}\bigcup_{\rho=0}^{5}\mathcal{S}_{\rho,j}$, where we define
\begin{align*}
    \mathcal{S}_{0,j}&:=\Bigg\{\frac{\frac{\partial g}{\partial W^{(u)}_{e}}(X,W^*_{e_j})}{1+\exp(-g(X,W^*_{e_i}))}F_0(Y|X;W^*_{e_j},\nu^*_j), \ \frac{\frac{\partial^2 g}{\partial W^{(u)}_{e}\partial W^{(v)}_{e}}(X,W^*_{e_j})}{1+\exp(-g(X,W^*_{e_j}))}F_0(Y|X;W^*_{e_j},\nu^*_j),\\
    &\frac{\frac{\partial g}{\partial W^{(u)}_{e}}(X,W^*_{e_j})\frac{\partial g}{\partial W^{(v)}_{e}}(X,W^*_{e_j})}{1+\exp(-g(X,W^*_{e_j}))}F_0(Y|X;W^*_{e_j},\nu^*_j), \ F_0(Y|X;W^*_{e_j},\nu^*_j):1\leq u,v\leq d_2\Bigg\},\\
    \mathcal{S}_{1,j}&:=\Bigg\{\frac{\frac{\partial g}{\partial W^{(u)}_{e}}(X,W^*_{e_j})}{1+\exp(-g(X,W^*_{e_i}))}F_1(Y|X;W^*_{e_j},\nu^*_j), \ \frac{\frac{\partial g}{\partial W^{(u)}_{e}}(X,W^*_{e_j})\frac{\partial g}{\partial W^{(v)}_{e}}(X,W^*_{e_j})}{1+\exp(-g(X,W^*_{e_j}))}F_1(Y|X;W^*_{e_j},\nu^*_j)\\
    &\hspace{2cm}\frac{\partial^2 g}{\partial W^{(u)}_{e}\partial W^{(v)}_{e}}(X,W^*_{e_j})F_1(Y|X;W^*_{e_j},\nu^*_j):1\leq u,v\leq d_2\Bigg\},\\
    \mathcal{S}_{2,j}&:=\Bigg\{F_2(Y|X;W^*_{e_j},\nu^*_j), \ \frac{\frac{\partial g}{\partial W^{(u)}_{e}}(X,W^*_{e_j})}{1+\exp(-g(X,W^*_{e_i}))}F_2(Y|X;W^*_{e_j},\nu^*_j),\\
    &\hspace{2cm}\frac{\partial g}{\partial W^{(u)}_{e}}(X,W^*_{e_j})\frac{\partial g}{\partial W^{(v)}_{e}}(X,W^*_{e_j})F_2(Y|X;W^*_{e_j},\nu^*_j): 1\leq u,v\leq d_2\Bigg\},\\
    \mathcal{S}_{3,j}&:=\Bigg\{\frac{\partial g}{\partial W^{(u)}_{e}}(X,W^*_{e_i})F_3(Y|X;W^*_{e_j},\nu^*_j):1\leq u\leq d_2\Bigg\},\\
    \mathcal{S}_{4,j}&:=\Bigg\{F_4(Y|X;W^*_{e_j},\nu^*_j)\Bigg\},\\
    \mathcal{S}_{5,j}&:=\Bigg\{\frac{\frac{\partial g}{\partial W^{(u)}_{e}}(X,W^*_{e_j})}{1+\exp(-g(X,W^*_{e_i}))}H_n(Y|X;W^*_{e_j},\nu^*_j), \ \frac{\frac{\partial^2 g}{\partial W^{(u)}_{e}\partial W^{(v)}_{e}}(X,W^*_{e_j})}{1+\exp(-g(X,W^*_{e_j}))}H_n(Y|X;W^*_{e_j},\nu^*_j),\\
    &\frac{\frac{\partial g}{\partial W^{(u)}_{e}}(X,W^*_{e_j})\frac{\partial g}{\partial W^{(v)}_{e}}(X,W^*_{e_j})}{1+\exp(-g(X,W^*_{e_j}))}H_n(Y|X;W^*_{e_j},\nu^*_j), \ H_n(Y|X;W^*_{e_j},\nu^*_j):1\leq u,v\leq d_2\Bigg\}.
\end{align*}
\textbf{Step 2: Non-vanishing coefficients.} In this step, we will show that at least one among the coefficients in the representations of $[T_{n,1,1}(Y|X)-R_{n,1,1}(Y|X)]/\mathcal{L}_1(G_n,G_*)$, $[T_{n,1,2}(Y|X)-R_{n,1,2}(Y|X)]/\mathcal{L}_1(G_n,G_*)$, $[T_{n,2,1}(Y|X)-R_{n,2,1}(Y|X)]/\mathcal{L}_1(G_n,G_*)$, $[T_{n,2,2}(Y|X)-R_{n,2,2}(Y|X)]/\mathcal{L}_1(G_n,G_*)$ and $[T_{n,3}(Y|X)]/\mathcal{L}_1(G_n,G_*)$ does not approach zero when $n$ goes to infinity. Assume by contrary that all of them vanish as $n\to\infty$. Then, by considering the coefficients of the term
\begin{itemize}
    \item $F_0(Y|X;W^*_{e_j},\nu^*_j)$ for $j\in[N^*]$, we have 
    \begin{align*}
        \frac{1}{\mathcal{L}_1(G_n,G_*)}\cdot\sum_{j=1}^{N^*}\Big|\sum_{i\in\mathcal{C}_{j}}\exp(c^n_i)-\exp(c^*_j)\Big|\to0.
    \end{align*}
    \item $\frac{\frac{\partial g}{\partial W^{(u)}_{e}}(X,W^*_{e_j})}{1+\exp(-g(X,W^*_{e_i}))}F_0(Y|X;W^*_{e_j},\nu^*_j)$ for $j\in[N^*]:|\mathcal{C}_{j}|=1$, we have
    \begin{align*}
         \frac{1}{\mathcal{L}_1(G_n,G_*)}\cdot\sum_{j\in[N^*]:|\mathcal{C}_{j}|=1}\sum_{i\in\mathcal{C}_{j}}\exp(c^n_i)\|\Delta W^n_{e_{ij}}\|_1\to0.
    \end{align*}
    Due to the equivalence between the $\ell_1$-norm and the $\ell_2$-norm, we obtain
    \begin{align*}
        \frac{1}{\mathcal{L}_1(G_n,G_*)}\cdot\sum_{j\in[N^*]:|\mathcal{C}_{j}|=1}\sum_{i\in\mathcal{C}_{j}}\exp(c^n_i)\|\Delta W^n_{e_{ij}}\|\to0.
    \end{align*}
    \item $F_2(Y|X;W^*_{e_j},\nu^*_j)$ for $j\in[N^*]:|\mathcal{C}_{j}|=1$, we have
    \begin{align*}
        \frac{1}{\mathcal{L}_1(G_n,G_*)}\cdot\sum_{j\in[N^*]:|\mathcal{C}_{j}|=1}\sum_{i\in\mathcal{C}_{j}}\exp(c^n_i)|\Delta\nu^n_{ij}|\to0.
    \end{align*}
    \item $\frac{\frac{\partial g}{\partial W^{(u)}_{e}}(X,W^*_{e_j})\frac{\partial g}{\partial W^{(u)}_{e}}(X,W^*_{e_j})}{1+\exp(-g(X,W^*_{e_j}))}F_0(Y|X;W^*_{e_j},\nu^*_j)$ for $j\in[N^*]:|\mathcal{C}_{j}|>1$, we have
     \begin{align*}
        \frac{1}{\mathcal{L}_1(G_n,G_*)}\cdot\sum_{j\in[N^*]:|\mathcal{C}_{j}|>1}\sum_{i\in\mathcal{C}_{j}}\exp(c^n_i)\|\Delta W^n_{e_{ij}}\|^2\to0.
    \end{align*}
    \item $F_4(Y|X;W^*_{e_j},\nu^*_j)$ for $j\in[N^*]:|\mathcal{C}_{j}|>1$, we have
    \begin{align*}
        \frac{1}{\mathcal{L}_1(G_n,G_*)}\cdot\sum_{j\in[N^*]:|\mathcal{C}_{j}|=1}\sum_{i\in\mathcal{C}_{j}}\exp(c^n_i)|\Delta\nu^n_{ij}|^2\to0.
    \end{align*}
\end{itemize}
By taking the sum of the above limits, we obtain $1=\frac{\mathcal{L}_1(G_n,G_*)}{\mathcal{L}_1(G_n,G_*)}\to0$ as $n\to\infty$, which is a contradiction. Thus, not all the coefficients in the representations of $[T_{n,1,1}(Y|X)-R_{n,1,1}(Y|X)]/\mathcal{L}_1(G_n,G_*)$, $[T_{n,1,2}(Y|X)-R_{n,1,2}(Y|X)]/\mathcal{L}_1(G_n,G_*)$, $[T_{n,2,1}(Y|X)-R_{n,2,1}(Y|X)]/\mathcal{L}_1(G_n,G_*)$, $[T_{n,2,2}(Y|X)-R_{n,2,2}(Y|X)]/\mathcal{L}_1(G_n,G_*)$ and $[T_{n,3}(Y|X)]/\mathcal{L}_1(G_n,G_*)$ converge to zero as $n\to\infty$.

\textbf{Stage 3 - Fatou's argument:} In this stage, we use the Fatou's lemma to show a contradiction to the result of Step 2. For that purpose, let us denote $m_n$ as the maximum of the absolute values of the coefficients in the representations of $[T_{n,1,1}(Y|X)-R_{n,1,1}(Y|X)]/\mathcal{L}_1(G_n,G_*)$, $[T_{n,1,2}(Y|X)-R_{n,1,2}(Y|X)]/\mathcal{L}_1(G_n,G_*)$, $[T_{n,2,1}(Y|X)-R_{n,2,1}(Y|X)]/\mathcal{L}_1(G_n,G_*)$, $[T_{n,2,2}(Y|X)-R_{n,2,2}(Y|X)]/\mathcal{L}_1(G_n,G_*)$ and $[T_{n,3}(Y|X)]/\mathcal{L}_1(G_n,G_*)$. It follows from the result of Step 2 that $1/m_n\not\to\infty$ as $n\to\infty$. In addition, we also denote
\begin{align*}
   \frac{\sum_{i\in\mathcal{C}_{j}}\exp(c^n_i)(\Delta W^n_{e_{ij}})^{(u)}}{m_n\mathcal{L}_1(G_n,G_*)}\to\alpha_{1,j}^{(u)},& \qquad \frac{\sum_{i\in\mathcal{C}_{j}}\exp(c^n_i)(\Delta\nu^n_{ij})}{m_n\mathcal{L}_1(G_n,G_*)}\to\beta_{1,j},\\
    \frac{\sum_{i\in\mathcal{C}_{j}}\exp(c^n_i)(\Delta W^n_{e_{ij}})^{(u)}(\Delta W^n_{e_{ij}})^{(v)}}{m_n\mathcal{L}_1(G_n,G_*)}\to\alpha_{2,j}^{(uv)},& \qquad \frac{\sum_{i\in\mathcal{C}_{j}}\exp(c^n_i)(\Delta\nu^n_{ij})^2}{m_n\mathcal{L}_1(G_n,G_*)}\to\beta_{2,j},\\
    \frac{\sum_{i\in\mathcal{C}_{j}}\exp(c^n_i)(\Delta W^n_{e_{ij}})^{(u)}(\Delta\nu^n_{ij})}{m_n\mathcal{L}_1(G_n,G_*)}\to\gamma_{j}^{(u)},& \qquad \frac{\sum_{i\in\mathcal{C}_{j}}\exp(c^n_i)-\exp(c^*_j)}{m_n\mathcal{L}_1(G_n,G_*)}\to\xi_{j}, 
\end{align*}
as $n\to\infty$ for any $j\in[N^*]$ and $u,v\in[d_2]$ with a note that at least one among $\alpha_{1,j}^{(u)},\beta_{1,j},\alpha_{2,j}^{(uv)},\beta_{2,j}$, $\gamma_{j}^{(u)}$ and $\xi_j$ is non-zero.

By applying the Fatou's lemma, we have
\begin{align*}
    0=\lim_{n\to\infty}\dfrac{\mathbb{E}_X[V(p_{G}(\cdot|X),p_{G_*}(\cdot|X))]}{m_n\mathcal{L}_1(G_n,G_*)}=\frac{1}{2}\int\liminf_{n\to\infty}\dfrac{|p_{G_n}(Y|X)-p_{G_*}(Y|X)|}{m_n\mathcal{L}_1(G_n,G_*)}\dint (X,Y),
\end{align*}
which implies that $[p_{G_n}(Y|X)-p_{G_*}(Y|X)]/[m_n\mathcal{L}_1(G_n,G_*)]\to0$ as $n\to\infty$ for almost surely $(X,Y)$. Since the term $\sum_{j=1}^{N^*}\exp(\log(1+\exp(g(x,W^*_{e_i}))))$ is bounded, we also have $T_n(Y|X)/[m_n\mathcal{L}_1(G_n,G_*)]\to0$ as $n\to\infty$. Then, it follows that
\begin{align}
    \label{eq:limit_over}
    0&=\lim_{n\to\infty}\dfrac{T_{n,1,1}(Y|X)+T_{n,1,2}(Y|X)}{m_n\mathcal{L}_1(G_n,G_*)}-\lim_{n\to\infty}\dfrac{T_{n,2,1}(Y|X)+T_{n,2,2}(Y|X)}{m_n\mathcal{L}_1(G_n,G_*)}+\lim_{n\to\infty}\dfrac{T_{n,3}(Y|X)}{m_n\mathcal{L}_1(G_n,G_*)},
\end{align}
for almost surely $(X,Y)\in\mathcal{X}\times\mathcal{Y}$, where we have
\begin{align*}
    &\lim_{n\to\infty}\dfrac{T_{n,1,1}(Y|X)}{m_n\mathcal{L}_1(G_n,G_*)}:=\sum_{j\in[N^*]:|\mathcal{C}_{j}|=1}\Bigg[\sum_{u=1}^{d_2}\alpha_{1,j}^{(u)}\frac{\frac{\partial g}{\partial W^{(u)}_{e}}(X,W^*_{e_j})}{1+\exp(-g(X,W^*_{e_j}))}F_{0,j}(Y|X)\\
    &\hspace{2cm}+\sum_{u=1}^{d_2}\alpha_{1,j}^{(u)}\frac{\partial g}{\partial W^{(u)}_{e}}(X,W^*_{e_j})F_{1,j}(Y|X)+\frac{1}{2}\beta_{1,j}F_{2,j}(Y|X)\Bigg],\\
    &\lim_{n\to\infty}\dfrac{T_{n,1,2}(Y|X)}{m_n\mathcal{L}_1(G_n,G_*)}:=\sum_{j\in[N^*]:|\mathcal{C}_{j}|>1}\Bigg[\Bigg(\sum_{u=1}^{d_2}\alpha_{1,j}^{(u)}\frac{\frac{\partial g}{\partial W^{(u)}_{e}}(X,W^*_{e_j})}{1+\exp(-g(X,W^*_{e_j}))}\\
    &+\sum_{u,v=1}^{d_2}\frac{\alpha_{2,j}^{(uv)}}{1+1_{\{u=v\}}}\cdot\frac{\frac{\partial^2 g}{\partial W^{(u)}_{e}\partial W^{(v)}_{e}}(X,W^*_{e_j})+\frac{\partial g}{\partial W^{(u)}_{e}}(X,W^*_{e_j})\frac{\partial g}{\partial W^{(v)}_{e}}(X,W^*_{e_j})}{1+\exp(-g(X,W^*_{e_j}))}\Bigg)F_{0,j}(Y|X)\\
    &+\Bigg(\sum_{u=1}^{d_2}\alpha_{1,j}^{(u)}\frac{\partial g}{\partial W^{(u)}_{e}}(X,W^*_{e_j})+\sum_{u,v=1}^{d_2}\frac{\alpha_{2,j}^{(uv)}}{1+1_{\{u=v\}}}\Bigg(\frac{2\frac{\partial g}{\partial W^{(u)}_{e}}(X,W^*_{e_j})\frac{\partial g}{\partial W^{(v)}_{e}}(X,W^*_{e_j})}{1+\exp(-g(X,W^*_{e_j}))}\\
    &+\frac{\partial^2 g}{\partial W^{(u)}_{e}\partial W^{(v)}_{e}}(X,W^*_{e_j})\Bigg)\Bigg)F_{1,j}(Y|X)+\Bigg(\frac{1}{2}\beta_{1,j}+\sum_{u=1}^{d_2}\gamma_{j}^{(u)}\cdot\frac{1}{2}\frac{\frac{\partial g}{\partial W^{(u)}_{e}}(X,W^*_{e_j})}{1+\exp(-g(X,W^*_{e_j}))}\\
    &+\sum_{u,v=1}^{d_2}\frac{\alpha_{2,j}^{(uv)}}{1+1_{\{u=v\}}}\cdot\frac{\partial g}{\partial W^{(u)}_{e}}(X,W^*_{e_j})\frac{\partial g}{\partial W^{(v)}_{e}}(X,W^*_{e_j})\Bigg)F_{2,j}(Y|X)\\
    &+\sum_{u=1}^{d_2}\frac{1}{2}\gamma_{j}^{(u)}\frac{\partial g}{\partial W^{(u)}_{e}}(X,W^*_{e_j})F_{3,j}(Y|X)+\frac{1}{4}\beta_{2,j}F_{4,j}(Y|X)\Bigg],
\end{align*}
and
\begin{align*}
     &\lim_{n\to\infty}\dfrac{T_{n,2,1}(Y|X)}{m_n\mathcal{L}_1(G_n,G_*)}:=\sum_{j\in[N^*]:|\mathcal{C}_{j}|=1}\sum_{u=1}^{d_2}\alpha_{1,j}^{(u)}\cdot\frac{\frac{\partial g}{\partial W^{(u)}_{e}}(X,W^*_{e_j})}{1+\exp(-g(X,W^*_{e_j}))}H_j(Y|X),\\
     &\lim_{n\to\infty}\dfrac{T_{n,2,2}(Y|X)}{m_n\mathcal{L}_1(G_n,G_*)}:=\sum_{j\in[N^*]:|\mathcal{C}_{j}|>1}\Bigg[\sum_{u=1}^{d_2}\alpha_{1,j}^{(u)}\cdot\frac{\frac{\partial g}{\partial W^{(u)}_{e}}(X,W^*_{e_j})}{1+\exp(-g(X,W^*_{e_j}))}\\
     &+\sum_{u,v=1}^{d_2}\frac{\alpha_{2,j}^{(uv)}}{1+1_{\{u=v\}}}\cdot\frac{\frac{\partial^2 g}{\partial W^{(u)}_{e}\partial W^{(v)}_{e}}(X,W^*_{e_j})+\frac{\partial g}{\partial W^{(u)}_{e}}(X,W^*_{e_j})\frac{\partial g}{\partial W^{(v)}_{e}}(X,W^*_{e_j})}{1+\exp(-g(X,W^*_{e_j}))}\Bigg]H_j(Y|X),
\end{align*}
and
\begin{align*}
    &\lim_{n\to\infty}\dfrac{T_{n,3}(Y|X)}{m_n\mathcal{L}_1(G_n,G_*)}:=\sum_{j=1}^{N^*}\xi_j[F_{0,j}(Y|X)-H_j(Y|X)].
\end{align*}
It is worth noting that for almost every $X$, the set
\begin{align*}
    \Bigg\{F_{\rho,j}(Y|X), \ H_j(Y|X):0\leq\rho\leq 4, j\in[N^*]\Bigg\}
\end{align*}
is linearly independent w.r.t $Y$. Therefore, it follows that the coefficients of those terms in the limit in equation~\eqref{eq:limit_over} become zero.

For $j\in[N^*]$ such that $|\mathcal{C}_{j}|=1$, by considering the coefficients of 
\begin{itemize}
    \item $F_{0,j}(Y|X)$, we have $\xi_j+\sum_{u=1}^{d_2}\alpha_{1,j}^{(u)}\cdot\frac{\frac{\partial g}{\partial W^{(u)}_{e}}(X,W^*_{e_j})}{1+\exp(-g(X,W^*_{e_j}))}=0$ for almost surely $X$. Since the expert function $g$ is strongly identifiable, we deduce $\xi_j=\alpha_{1,j}^{(u)}=0$ for all $u\in[d_2]$;
    \item $F_{2,j}(Y|X)$, we have $\beta_{1,j}=0$.
\end{itemize}
For $j\in[N^*]$ such that $|\mathcal{C}_{j}|>1$, by considering the coefficients of 
\begin{itemize}
    \item $F_{0,j}(Y|X)$, we have 
    \begin{align*}
        &\xi_j+\sum_{u=1}^{d_2}\alpha_{1,j}^{(u)}\frac{\frac{\partial g}{\partial W^{(u)}_{e}}(X,W^*_{e_j})}{1+\exp(-g(X,W^*_{e_j}))}\\
    &+\sum_{u,v=1}^{d_2}\frac{\alpha_{2,j}^{(uv)}}{1+1_{\{u=v\}}}\cdot\frac{\frac{\partial^2 g}{\partial W^{(u)}_{e}\partial W^{(v)}_{e}}(X,W^*_{e_j})+\frac{\partial g}{\partial W^{(u)}_{e}}(X,W^*_{e_j})\frac{\partial g}{\partial W^{(v)}_{e}}(X,W^*_{e_j})}{1+\exp(-g(X,W^*_{e_j}))}=0
    \end{align*}
    for almost surely $X$. Since the expert function $g$ is strongly identifiable, we deduce $\xi_j=\alpha_{1,j}^{(u)}=\alpha_{2,j}^{(uv)}=0$ for all $u,v\in[d_2]$;
    \item $F_{3,j}(Y|X)$, we have $\sum_{u=1}^{d_2}\frac{1}{2}\gamma_{j}^{(u)}\frac{\partial g}{\partial W^{(u)}_{e}}(X,W^*_{e_j})=0$ for almost surely $X$. Since the expert function $g$ is strongly identifiable, we deduce $\gamma_{j}^{(u)}=0$ for all $u\in[d_2]$;
    \item $F_{4,j}(Y|X)$, we have $\beta_{2,j}=0$.
\end{itemize}
Putting the above results together, we have (ii) $\xi_j=\alpha_{1,j}^{(u)}=\beta_{1,j}=\alpha_{2,j}^{(uv)}=\beta_{2,j}=\gamma_{j}^{(u)}=0$ for all $j\in[N^*]$ and $u,v\in[d_2]$. This contradicts to the fact that at least one among them is non-zero. Consequently, we achieve the local part in~\eqref{eq:local_part_over}.

\textbf{Global part:} Now, it suffices to demonstrate that 
\begin{align*}
    \inf_{G\in\mathcal{G}_{N}(\Theta):\mathcal{L}_1(G,G_*)>\varepsilon'}\dfrac{\bbE_X[V(p_{G}(\cdot|X),p_{G_*}(\cdot|X))]}{\mathcal{L}_1(G,G_*)}>0,
\end{align*}
for some positive constant $\varepsilon'$. Given the above result, it is sufficient to derive the global part in~\eqref{eq:global_part_over}, that is,
\begin{align*}
    \inf_{G\in\mathcal{G}_{N}(\Theta):\mathcal{L}_1(G,G_*)>\varepsilon'}\mathbb{E}_X[V(p_{G}(\cdot|X),p_{G_*}(\cdot|X))]/\mathcal{L}_1(G,G_*)>0.
\end{align*}
Assume by contrary that the global part does not hold true, then we can find a sequence $\widetilde{G}_n\in\mathcal{G}_{N}(\Theta)$ such that $\mathcal{L}_1(\widetilde{G}_n,G_*)>\varepsilon'$ and $\mathbb{E}_X[V(p_{\widetilde{G}_n}(\cdot|X),p_{G_*}(\cdot|X))]\to0$ as $n\to\infty$. Since $\Theta$ is a compact set, we are able to replace $\widetilde{G}_n$ with its subsequence which converges to some mixing measure $\widetilde{G}\in\mathcal{G}_{N}(\Theta)$. Recall that $\mathcal{L}_1(\widetilde{G}_n,G_*)>\varepsilon'$, then we also get that $\mathcal{L}_1(\widetilde{G},G_*)>\varepsilon'$.

On the other hand, by means of the Fatou's lemma, we have
\begin{align*}
    0=\lim_{n\to\infty}\mathbb{E}_X[2V(p_{\widetilde{G}_n}(\cdot|X),p_{G_*}(\cdot|X))]\geq\int\liminf_{n\to\infty}|p_{\widetilde{G}_n}(Y|X)-p_{G_*}(Y|X)|\dint(X,Y),
\end{align*}
which follows that $p_{\widetilde{G}}(Y|X)-p_{G_*}(Y|X)=0$ for almost surely $(X,Y)$. Thus, we achieve that $\widetilde{G}\equiv G_*$, or equivalently $\mathcal{L}_1(\widetilde{G},G_*)=0$. This contradicts to the fact that $\mathcal{L}_1(\widetilde{G},G_*)>\varepsilon'>0$. 

Hence, we reach the conclusion in \eqref{eq:global_part_over}, and the proof is completed.

\subsection{Proof of Theorem~\ref{theorem:parameter_estimation_linear}}
\label{appendix:parameter_estimation_linear}
As in Appendix~\ref{appendix:parameter_estimation_over}, we also start with establishing the local part 
\begin{align}
    \label{eq:local_part_linear}
    \lim_{\varepsilon\to0}\inf_{G\in\mathcal{G}_{N}(\Theta):\mathcal{L}_2(G,G_*)\leq\varepsilon}\dfrac{\bbE_X[V(p_{G}(\cdot|X),p_{G_*}(\cdot|X))]}{\mathcal{L}_2(G,G_*)}>0.
\end{align}
Assume by contrary that the local part is not true, then we can find a sequence of mixing measures $(G_n)$ given by $G_n:=\sum_{i=1}^{N}\exp(c^n_i)\delta_{(a^n_{i},b^n_i,\nuin)}\in\mathcal{G}_{N}(\Theta)$ such that $\mathcal{L}_2(G_n,G_*)\to0$ and
\begin{align*}
    \mathbb{E}_X[V(p_{G_n}(\cdot|X),p_{G_*}(\cdot|X))]/\mathcal{L}_2(G_n,G_*)\to0,
\end{align*}
as $n\to\infty$. Recall that the Voronoi loss $\mathcal{L}_2(G_n,G_*)$ is given by
\begin{align}
    \label{eq:Voronoi_loss_linear_n}
    \mathcal{L}_2(G_n,G_*):=\sum_{j=1}^{N^*}\Big|\sum_{i\in\mathcal{C}_{j}}&\exp(c^n_i)-\exp(c^*_j)\Big|+\sum_{j\in[N^*]:|\mathcal{C}_{j}|=1}\sum_{i\in\mathcal{C}_{j}}\exp(c^n_i)\Big[\|W^n_{e_{i}}-W^*_{e_{j}}\|+|\nu^n_{i}-\nu^*_{j}|\Big]\nonumber\\
    &+\sum_{j\in[N^*]:|\mathcal{C}_{j}|=1}\sum_{i\in\mathcal{C}_{j}}\exp(c^n_i)\Big[\|W^n_{e_{i}}-W^*_{e_{j}}\|^2+|\nu^n_{i}-\nu^*_{j}|^2\Big].
\end{align}
Since $\mathcal{L}_2(G_n,G_*)\to0$ as $n\to\infty$, we obtain $(a^n_{i},b^n_i,\nu^n_i)\to (a^*_{j},b^*_j,\nu^*_j)$  for all $j\in[N^*]$ and $i\in\mathcal{C}_{j}$. 

Next, we divide the rest of this proof into three main steps.

\textbf{Step 1: Taylor expansion.} In this step, we aim to decompose the term $T_n(Y|X):=\Big[\sum_{j=1}^{N^*}\exp(\log(1+\exp((a^*_j)^{\top}X+b^*_j)))\Big]\cdot [p_{G_n}(Y|X)-p_{G_*}(Y|X)]$ can be decomposed as
\begin{align*}
    &T_n(Y|X)=\sum_{j=1}^{N^*}\sum_{i\in\mathcal{C}_{j}}\exp(c^n_i)\Big[\exp(\log(1+\exp((a^n_i)^{\top}X+b^n_i)))f(Y|(a^n_i)^{\top}X+b^n_i,\nu^n_i)\\
    &\hspace{2cm} -\exp(\log(1+\exp((a^*_j)^{\top}X+b^*_j)))f(Y|(a^*_j)^{\top}X+b^*_j,\nu^*_j)\Big]\\
    &-\sum_{j=1}^{N^*}\sum_{i\in\mathcal{C}_{j}}\exp(c^n_i)\Big[\exp(\log(1+\exp((a^n_i)^{\top}X+b^n_i)))-\exp(\log(1+\exp((a^*_j)^{\top}X+b^*_j)))]p_{G_n}(Y|X)\\
    &+\sum_{j=1}^{N^*}\Big[\sum_{i\in\mathcal{C}_{j}}\exp(c^n_i)-\exp(c^*_j)\Big]\cdot\exp(\log(1+\exp((a^*_j)^{\top}X+b^*_j)))[f(Y|(a^*_j)^{\top}X+b^*_j,\nu^*_j)-p_{G_n}(Y|X)]\\
    &:=T_{n,1}(Y|X)-T_{n,2}(Y|X)+T_{n,3}(Y|X).
\end{align*}
Next, we continue to decompose the term $T_{n,1}(Y|X)$ as
\begin{align*}
    T_{n,1}(Y|X)&=\sum_{j\in[N^*]:|\mathcal{C}_{j}|=1}\sum_{i\in\mathcal{C}_{j}}\exp(c^n_i)\Big[\exp(\log(1+\exp((a^n_i)^{\top}X+b^n_i)))f(Y|(a^n_i)^{\top}X+b^n_i,\nu^n_i)\\
    &\hspace{3cm} -\exp(\log(1+\exp((a^*_j)^{\top}X+b^*_j)))f(Y|(a^*_j)^{\top}X+b^*_j,\nu^*_j)\Big]\\
    &+\sum_{j\in[N^*]:|\mathcal{C}_{j}|>1}\sum_{i\in\mathcal{C}_{j}}\exp(c^n_i)\Big[\exp(\log(1+\exp((a^n_i)^{\top}X+b^n_i)))f(Y|(a^n_i)^{\top}X+b^n_i,\nu^n_i)\\
    &\hspace{3cm} -\exp(\log(1+\exp((a^*_j)^{\top}X+b^*_j)))f(Y|(a^*_j)^{\top}X+b^*_j,\nu^*_j)\Big]\\
    &:=T_{n,1,1}(Y|X)+T_{n,1,2}(Y|X).
\end{align*}
Let us denote $F_{\rho}(Y|X;a,b,\nu):=\exp(\log(1+\exp(a^{\top}X+b)))\frac{\partial^{\rho}f}{\partial g^{\rho}}(Y|a^{\top}X+b,\nu)$. By applying the first-order Taylor expansion to the function $F_{0}(Y|X;a,b,\nu)$ around the point $(a^*_j,b^*_j,\nu^*_j)$, we rewrite the term $T_{n,1,1}(Y|X)$ as
\begin{align*}
    &T_{n,1,1}(Y|X)=\sum_{j\in[N^*]:|\mathcal{C}_{j}|=1}\sum_{\rho=0}^{2}T_{n,1,1,\rho}^{(j)}(X)F_{\rho}(Y;X,a^*_j,b^*_j,\nu^*_j)+R_{n,1,1}(Y|X),
\end{align*}
where $R_{n,1,1}(Y|X)$ is the Taylor remainder such that $R_{n,1,1}(Y|X)/\mathcal{L}_2(G_n,G_*)\to0$ as $n\to\infty$, and
\begin{align*}
    T_{n,1,1,0}^{(j)}(X)&:=\sum_{i\in\mathcal{C}_{j}}\exp(c^n_i)\cdot\frac{\sum_{u=1}^{d}(\Delta a^n_{ij})^{(u)}X^{(u)}+(\Delta b^n_{ij})}{1+\exp(-(a^*_j)^{\top}X-b^*_j)},\\
    T_{n,1,1,1}^{(j)}(X)&:=\sum_{i\in\mathcal{C}_{j}}\exp(c^n_i)\Bigg[\sum_{u=1}^{d}(\Delta a^n_{ij})^{(u)}X^{(u)}+(\Delta b^n_{ij})\Bigg],\\
    T_{n,1,1,2}^{(j)}(X)&:=\sum_{i\in\mathcal{C}_{j}}\frac{1}{2}\exp(c^n_i)(\Delta\nu^n_{ij}),
\end{align*}
in which $\Delta a^n_{ij}:=a^n_{i}-a^*_{j}$, $\Delta b^n_{ij}:=b^n_{i}-b^*_{j}$ and $\Delta\nu^n_{ij}:=\nu^n_i-\nu^*_j$.

Meanwhile, by means of the second-order Taylor expansion, the term $T_{n,1,2}(Y|X)$ can be represented as
\begin{align*}
    &T_{n,1,2}(Y|X)=\sum_{j\in[N^*]:|\mathcal{C}_{j}|>1}\sum_{\rho=0}^{4}T_{n,1,2,\rho}^{(j)}(X)F_{\rho}(Y;X,a^*_{j},b^*_j,\nu^*_j)+R_{n,1,2}(Y|X),
\end{align*}
where $R_{n,1,2}(Y|X)$ is the Taylor remainder such that $R_{n,1,2}(Y|X)/\mathcal{L}_2(G_n,G_*)\to0$ as $n\to\infty$, and
\begin{align*}
     T_{n,1,2,0}^{(j)}(X)&:=\sum_{i\in\mathcal{C}_{j}}\exp(c^n_i)\Bigg[\frac{\sum_{u=1}^{d}(\Delta a^n_{ij})^{(u)}X^{(u)}+(\Delta b^n_{ij})}{1+\exp(-(a^*_j)^{\top}X-b^*_j)}+\frac{\sum_{u,v=1}^{d}\frac{(\Delta a^n_{ij})^{(u)}(\Delta a^n_{ij})^{(v)}}{1+1_{\{u=v\}}}X^{(u)}X^{(v)}}{1+\exp(-(a^*_j)^{\top}X-b^*_j)}\\
     &\hspace{4cm}+\frac{\sum_{u=1}^{d}(\Delta a^n_{ij})^{(u)}(\Delta b^n_{ij})X^{(u)}+\frac{1}{2}(\Delta b^n_{ij})^2}{1+\exp(-(a^*_j)^{\top}X-b^*_j)}\Bigg],\\
    T_{n,1,2,1}^{(j)}(X)&:=\sum_{i\in\mathcal{C}_{j}}\exp(c^n_i)\Bigg[\sum_{u=1}^{d}(\Delta a^n_{ij})^{(u)}X^{(u)}+(\Delta b^n_{ij})+\frac{2\sum_{u,v=1}^{d}\frac{(\Delta a^n_{ij})^{(u)}(\Delta a^n_{ij})^{(v)}}{1+1_{\{u=v\}}}X^{(u)}X^{(v)}}{1+\exp(-(a^*_j)^{\top}X-b^*_j)}\\
    &\hspace{4cm}+\frac{(\Delta b^n_{ij})^2+2\sum_{u=1}^{d}(\Delta a^n_{ij})^{(u)}(\Delta b^n_{ij})X^{(u)}}{1+\exp(-(a^*_j)^{\top}X-b^*_j)}\Bigg],\\
    T_{n,1,2,2}^{(j)}(X)&:=\sum_{i\in\mathcal{C}_{j}}\exp(c^n_i)\Bigg[\frac{1}{2}(\Delta\nu^n_{ij})+\sum_{u,v=1}^{d}\frac{(\Delta a^n_{ij})^{(u)}(\Delta a^n_{ij})^{(v)}X^{(u)}X^{(v)}}{1+1_{\{u=v\}}}+\frac{1}{2}(\Delta b^n_{ij})^2\\
    &+\sum_{u=1}^{d}(\Delta a^n_{ij})^{(u)}(\Delta b^n_{ij})X^{(u)}+\frac{1}{2}\cdot\frac{\sum_{u=1}^{d}(\Delta a^n_{ij})^{(u)}(\Delta\nu^n_{ij})X^{(u)}+(\Delta b^n_{ij})(\Delta\nu^n_{ij})}{1+\exp(-(a^*_j)^{\top}X-b^*_j)}\Bigg],\\
    T_{n,1,2,3}^{(j)}(X)&:=\sum_{i\in\mathcal{C}_{j}}\exp(c^n_i)\Bigg[\sum_{u=1}^{d}\frac{1}{2}(\Delta a^n_{ij})^{(u)}(\Delta\nu^n_{ij})X^{(u)}+\frac{1}{2}(\Delta b^n_{ij})(\Delta\nu^n_{ij})\Bigg],\\
    T_{n,1,2,4}^{(j)}(X)&:=\sum_{i\in\mathcal{C}_{j}}\exp(c^n_i)\cdot\frac{1}{4}(\Delta\nu^n_{ij})^2.
\end{align*}
Next, we decompose the term $T_{n,2}(Y|X)$ as
\begin{align*}
    &T_{n,2}(Y|X)\\
    &:=\sum_{j\in[N^*]:|\mathcal{C}_{j}|=1}\sum_{i\in\mathcal{C}_{j}}\exp(c^n_i)\Big[\exp(\log(1+\exp((a^n_i)^{\top}X+b^n_i)))-\exp(\log(1+\exp((a^*_j)^{\top}X+b^*_j)))]p_{G_n}(Y|X)\\
    &+\sum_{j\in[N^*]:|\mathcal{C}_{j}|>1}\sum_{i\in\mathcal{C}_{j}}\exp(c^n_i)\Big[\exp(\log(1+\exp((a^n_i)^{\top}X+b^n_i)))-\exp(\log(1+\exp((a^*_j)^{\top}X+b^*_j)))]p_{G_n}(Y|X)\\
    &:=T_{n,2,1}(Y|X)+T_{n,2,2}(Y|X).
\end{align*}
Note that we can rewrite the term $T_{n,1,2}(Y|X)$ using the first-order Taylor expansion to the function $\exp(\log(1+\exp((a^n_i)^{\top}X+b^n_i)))$ around the point $(a^*_{j},b^*_j)$ as
\begin{align*}
    T_{n,2,1}(Y|X)=\sum_{j\in[N^*]:|\mathcal{C}_{j}|=1}\sum_{i\in\mathcal{C}_{j}}\exp(c^n_i)\cdot\frac{\sum_{u=1}^{d}(\Delta a^n_{ij})^{(u)}X^{(u)}+(\Delta b^n_{ij})}{1+\exp(-(a^*_j)^{\top}X-b^*_j)}H_n(Y|X;a^*_{j},b^*_{j}) \\
    + R_{n,2,1}(Y|X),
\end{align*}
where we denote $H_n(Y|X;a,b)=\exp(\log(1+\exp(a^{\top}X+b)))p_{G_n}(Y|X)$ and $R_{n,2,1}(Y|X)$ is the Taylor remainder such that $R_{n,2,1}(Y|X)/\mathcal{L}_2(G_n,G_*)\to0$ as $n\to\infty$.

On the other hand, by means of the second-order Taylor expansion, we have
\begin{align*}
    &T_{n,2,2}(Y|X)=\sum_{j\in[N^*]:|\mathcal{C}_{j}|>1}\sum_{i\in\mathcal{C}_{j}}\exp(c^n_i)\Bigg[\frac{\sum_{u=1}^{d}(\Delta a^n_{ij})^{(u)}X^{(u)}+(\Delta b^n_{ij})}{1+\exp(-(a^*_j)^{\top}X-b^*_j)}\\
     &+\frac{\sum_{u,v=1}^{d}\frac{(\Delta a^n_{ij})^{(u)}(\Delta a^n_{ij})^{(v)}}{1+1_{\{u=v\}}}X^{(u)}X^{(v)}}{1+\exp(-(a^*_j)^{\top}X-b^*_j)}+\frac{\sum_{u=1}^{d}(\Delta a^n_{ij})^{(u)}(\Delta b^n_{ij})X^{(u)}+\frac{1}{2}(\Delta b^n_{ij})^2}{1+\exp(-(a^*_j)^{\top}X-b^*_j)}\Bigg]H_n(Y|X;W^*_{e_j}) \\
    &\hspace{12cm}+ R_{n,2,2}(Y|X),
\end{align*}
where $R_{n,2,1}(Y|X)$ is the Taylor remainder such that $R_{n,2,2}(Y|X)/\mathcal{L}_2(G_n,G_*)\to0$ as $n\to\infty$.

From the above equation, $[T_{n,1,1}(Y|X)-R_{n,1,1}(Y|X)]$, $[T_{n,1,2}(Y|X)-R_{n,1,2}(Y|X)]$, $[T_{n,2,1}(Y|X)-R_{n,2,1}(Y|X)]$, $[T_{n,2,2}(Y|X)-R_{n,2,2}(Y|X)]$ and $[T_{n,3}(Y|X)]$ can be seen as a combination of elements of the set $\mathcal{S}:=\bigcup_{j=1}^{N}\bigcup_{\rho=0}^{5}\mathcal{S}_{\rho,j}$, where we define
\begin{align*}
    \mathcal{S}_{0,j}&:=\Bigg\{\frac{X^{(u)}}{1+\exp(-(a^*_j)^{\top}X-b^*_j)}F_{0,j}(Y|X), \ \frac{X^{(u)}X^{(v)}}{1+\exp(-(a^*_j)^{\top}X-b^*_j)}F_{0,j}(Y|X),\\
    &\hspace{3cm}\frac{1}{1+\exp(-(a^*_j)^{\top}X-b^*_j)}F_{0,j}(Y|X), \ F_{0,j}(Y|X):1\leq u,v\leq d\Bigg\},\\
    \mathcal{S}_{1,j}&:=\Bigg\{F_{1,j}(Y|X), \ X^{(u)}F_{1,j}(Y|X), \ \frac{X^{(u)}}{1+\exp(-(a^*_j)^{\top}X-b^*_j)}F_{1,j}(Y|X),\\
    &\frac{X^{(u)}X^{(v)}}{1+\exp(-(a^*_j)^{\top}X-b^*_j)}F_{1,j}(Y|X), \ \frac{1}{1+\exp(-(a^*_j)^{\top}X-b^*_j)}F_{1,j}(Y|X):1\leq u,v\leq d\Bigg\},\\
    \mathcal{S}_{2,j}&:=\Bigg\{F_{2,j}(Y|X), \ X^{(u)}F_{2,j}(Y|X), \ X^{(u)}X^{(v)}F_{2,j}(Y|X),\\
    & \frac{X^{(u)}}{1+\exp(-(a^*_j)^{\top}X-b^*_j)}F_{2,j}(Y|X), \ \frac{1}{1+\exp(-(a^*_j)^{\top}X-b^*_j)}F_{2,j}(Y|X): 1\leq u,v\leq d_2\Bigg\},\\
    \mathcal{S}_{3,j}&:=\Bigg\{F_{3,j}(Y|X), \ X^{(u)}F_{3,j}(Y|X):1\leq u\leq d\Bigg\},\\
    \mathcal{S}_{4,j}&:=\Bigg\{F_{4,j}(Y|X)\Bigg\},\\
    \mathcal{S}_{5,j}&:=\Bigg\{\frac{X^{(u)}}{1+\exp(-(a^*_j)^{\top}X-b^*_j)}H_{n,j}(Y|X), \ \frac{X^{(u)}X^{(v)}}{1+\exp(-(a^*_j)^{\top}X-b^*_j)}H_{n,j}(Y|X),\\
    &\hspace{3cm}\frac{1}{1+\exp(-(a^*_j)^{\top}X-b^*_j)}H_{n,j}(Y|X), \ H_{n,j}(Y|X):1\leq u,v\leq d\Bigg\}.
\end{align*}
\textbf{Step 2: Non-vanishing coefficients.} In this step, we will show that at least one among the coefficients in the representations of $[T_{n,1,1}(Y|X)-R_{n,1,1}(Y|X)]/\mathcal{L}_2(G_n,G_*)$, $[T_{n,1,2}(Y|X)-R_{n,1,2}(Y|X)]/\mathcal{L}_2(G_n,G_*)$, $[T_{n,2,1}(Y|X)-R_{n,2,1}(Y|X)]/\mathcal{L}_2(G_n,G_*)$, $[T_{n,2,2}(Y|X)-R_{n,2,2}(Y|X)]/\mathcal{L}_2(G_n,G_*)$ and $[T_{n,3}(Y|X)]/\mathcal{L}_2(G_n,G_*)$ does not approach zero when $n$ goes to infinity. Assume by contrary that all of them vanish as $n\to\infty$. Then, by considering the coefficients of the term
\begin{itemize}
    \item $F_{0,j}(Y|X)$ for $j\in[N^*]$, we have 
    \begin{align*}
        \frac{1}{\mathcal{L}_2(G_n,G_*)}\cdot\sum_{j=1}^{N^*}\Big|\sum_{i\in\mathcal{C}_{j}}\exp(c^n_i)-\exp(c^*_j)\Big|\to0.
    \end{align*}
    \item $\frac{X^{(u)}}{1+\exp(-(a^*_j)^{\top}X-b^*_j)}F_{0,j}(Y|X)$ for $j\in[N^*]:|\mathcal{C}_{j}|=1$, we have
    \begin{align*}
        \frac{1}{\mathcal{L}_2(G_n,G_*)}\cdot\sum_{j\in[N^*]:|\mathcal{C}_{j}|=1}\sum_{i\in\mathcal{C}_{j}}\exp(c^n_i)\|\Delta a^n_{ij}\|\to0.
    \end{align*}
    \item $\frac{1}{1+\exp(-(a^*_j)^{\top}X-b^*_j)}F_{0,j}(Y|X)$ for $j\in[N^*]:|\mathcal{C}_{j}|=1$, we have
    \begin{align*}
        \frac{1}{\mathcal{L}_2(G_n,G_*)}\cdot\sum_{j\in[N^*]:|\mathcal{C}_{j}|=1}\sum_{i\in\mathcal{C}_{j}}\exp(c^n_i)\|\Delta b^n_{ij}\|\to0.
    \end{align*}
    \item $F_{2,j}(Y|X)$ for $j\in[N^*]:|\mathcal{C}_{j}|=1$, we have
    \begin{align*}
        \frac{1}{\mathcal{L}_2(G_n,G_*)}\cdot\sum_{j\in[N^*]:|\mathcal{C}_{j}|=1}\sum_{i\in\mathcal{C}_{j}}\exp(c^n_i)|\Delta\nu^n_{ij}|\to0.
    \end{align*}
    \item $\frac{X^{(u)}X^{(v)}}{1+\exp(-(a^*_j)^{\top}X-b^*_j))}F_{0,j}(Y|X)$ for $j\in[N^*]:|\mathcal{C}_{j}|>1$, we have
     \begin{align*}
        \frac{1}{\mathcal{L}_2(G_n,G_*)}\cdot\sum_{j\in[N^*]:|\mathcal{C}_{j}|>1}\sum_{i\in\mathcal{C}_{j}}\exp(c^n_i)\|\Delta a^n_{j}\|^2\to0.
    \end{align*}
    \item $\frac{1}{1+\exp(-(a^*_j)^{\top}X-b^*_j))}F_{1,j}(Y|X)$ for $j\in[N^*]:|\mathcal{C}_{j}|>1$, we have
     \begin{align*}
        \frac{1}{\mathcal{L}_2(G_n,G_*)}\cdot\sum_{j\in[N^*]:|\mathcal{C}_{j}|>1}\sum_{i\in\mathcal{C}_{j}}\exp(c^n_i)|\Delta b^n_{j}|^2\to0.
    \end{align*}
    \item $F_{4,j}(Y|X)$ for $j\in[N^*]:|\mathcal{C}_{j}|>1$, we have
    \begin{align*}
        \frac{1}{\mathcal{L}_2(G_n,G_*)}\cdot\sum_{j\in[N^*]:|\mathcal{C}_{j}|=1}\sum_{i\in\mathcal{C}_{j}}\exp(c^n_i)|\Delta\nu^n_{ij}|^2\to0.
    \end{align*}
\end{itemize}
By taking the sum of the above limits, we obtain $1=\frac{\mathcal{L}_2(G_n,G_*)}{\mathcal{L}_2(G_n,G_*)}\to0$ as $n\to\infty$, which is a contradiction. Thus, not all the coefficients in the representations of $[T_{n,1,1}(Y|X)-R_{n,1,1}(Y|X)]/\mathcal{L}_2(G_n,G_*)$, $[T_{n,1,2}(Y|X)-R_{n,1,2}(Y|X)]/\mathcal{L}_2(G_n,G_*)$, $[T_{n,2,1}(Y|X)-R_{n,2,1}(Y|X)]/\mathcal{L}_2(G_n,G_*)$, $[T_{n,2,2}(Y|X)-R_{n,2,2}(Y|X)]/\mathcal{L}_2(G_n,G_*)$ and $[T_{n,3}(Y|X)]/\mathcal{L}_2(G_n,G_*)$ converge to zero as $n\to\infty$.

\textbf{Stage 3 - Fatou's argument:} In this stage, we use the Fatou's lemma to show a contradiction to the result of Step 2. For that purpose, let us denote $m_n$ as the maximum of the absolute values of the coefficients in the representations of $[T_{n,1,1}(Y|X)-R_{n,1,1}(Y|X)]/\mathcal{L}_2(G_n,G_*)$, $[T_{n,1,2}(Y|X)-R_{n,1,2}(Y|X)]/\mathcal{L}_2(G_n,G_*)$, $[T_{n,2,1}(Y|X)-R_{n,2,1}(Y|X)]/\mathcal{L}_2(G_n,G_*)$, $[T_{n,2,2}(Y|X)-R_{n,2,2}(Y|X)]/\mathcal{L}_2(G_n,G_*)$ and $[T_{n,3}(Y|X)]/\mathcal{L}_2(G_n,G_*)$. It follows from the result of Step 2 that $1/m_n\not\to\infty$ as $n\to\infty$. In addition, we also denote
\begin{align*}
   \frac{\sum_{i\in\mathcal{C}_{j}}\exp(c^n_i)(\Delta a^n_{ij})^{(u)}}{m_n\mathcal{L}_2(G_n,G_*)}\to\alpha_{1,j}^{(u)},& \qquad \frac{\sum_{i\in\mathcal{C}_{j}}\exp(c^n_i)(\Delta\nu^n_{ij})}{m_n\mathcal{L}_2(G_n,G_*)}\to\beta_{1,j},\\
    \frac{\sum_{i\in\mathcal{C}_{j}}\exp(c^n_i)(\Delta a^n_{ij})^{(u)}(\Delta a^n_{ij})^{(v)}}{m_n\mathcal{L}_2(G_n,G_*)}\to\alpha_{2,j}^{(uv)},& \qquad \frac{\sum_{i\in\mathcal{C}_{j}}\exp(c^n_i)(\Delta\nu^n_{ij})^2}{m_n\mathcal{L}_2(G_n,G_*)}\to\beta_{2,j},\\
    \frac{\sum_{i\in\mathcal{C}_{j}}\exp(c^n_i)(\Delta b^n_{ij})}{m_n\mathcal{L}_2(G_n,G_*)}\to\phi_{1,j}^{(u)},& \qquad \frac{\sum_{i\in\mathcal{C}_{j}}\exp(c^n_i)(\Delta b^n_{ij})^2}{m_n\mathcal{L}_2(G_n,G_*)}\to\phi_{2,j},\\
    \frac{\sum_{i\in\mathcal{C}_{j}}\exp(c^n_i)(\Delta a^n_{ij})^{(u)}(\Delta\nu^n_{ij})}{m_n\mathcal{L}_2(G_n,G_*)}\to\gamma_{1,j}^{(u)},& \qquad \frac{\sum_{i\in\mathcal{C}_{j}}\exp(c^n_i)(\Delta a^n_{ij})^{(u)}(\Delta b^n_{ij})}{m_n\mathcal{L}_2(G_n,G_*)}\to\gamma_{2,j}^{(u)},\\
    \frac{\sum_{i\in\mathcal{C}_{j}}\exp(c^n_i)(\Delta b^n_{ij})(\Delta\nu^n_{ij})}{m_n\mathcal{L}_2(G_n,G_*)}\to\gamma_{3,j},& \qquad\frac{\sum_{i\in\mathcal{C}_{j}}\exp(c^n_i)-\exp(c^*_j)}{m_n\mathcal{L}_2(G_n,G_*)}\to\xi_{j}, 
\end{align*}
as $n\to\infty$ for any $j\in[N^*]$ and $u,v\in[d_2]$ with a note that at least one among $\alpha_{1,j}^{(u)},\beta_{1,j},\alpha_{2,j}^{(uv)},\beta_{2,j}$, $\phi_{1,j}$, $\phi_{2,j}$, $\gamma_{1,j}^{(u)}$, $\gamma_{2,j}^{(u)}$, $\gamma_{3,j}$ and $\xi_j$ is non-zero.

By applying the Fatou's lemma, we have
\begin{align*}
    0=\lim_{n\to\infty}\dfrac{\mathbb{E}_X[V(p_{G}(\cdot|X),p_{G_*}(\cdot|X))]}{m_n\mathcal{L}_2(G_n,G_*)}=\frac{1}{2}\int\liminf_{n\to\infty}\dfrac{|p_{G_n}(Y|X)-p_{G_*}(Y|X)|}{m_n\mathcal{L}_2(G_n,G_*)}\dint (X,Y),
\end{align*}
which implies that $[p_{G_n}(Y|X)-p_{G_*}(Y|X)]/[m_n\mathcal{L}_2(G_n,G_*)]\to0$ as $n\to\infty$ for almost surely $(X,Y)$. Since the term $\sum_{j=1}^{N^*}\exp(\log(1+\exp((a^*_j)^{\top}X+b^*_j)))$ is bounded, we also have $T_n(Y|X)/[m_n\mathcal{L}_2(G_n,G_*)]\to0$ as $n\to\infty$. Then, it follows that
\begin{align}
    \label{eq:limit_linear}
    0&=\lim_{n\to\infty}\dfrac{T_{n,1,1}(Y|X)+T_{n,1,2}(Y|X)}{m_n\mathcal{L}_2(G_n,G_*)}-\lim_{n\to\infty}\dfrac{T_{n,2,1}(Y|X)+T_{n,2,2}(Y|X)}{m_n\mathcal{L}_2(G_n,G_*)}+\lim_{n\to\infty}\dfrac{T_{n,3}(Y|X)}{m_n\mathcal{L}_2(G_n,G_*)},
\end{align}
for almost surely $(X,Y)\in\mathcal{X}\times\mathcal{Y}$, where we have
\begin{align*}
    &\lim_{n\to\infty}\dfrac{T_{n,1,1}(Y|X)}{m_n\mathcal{L}_2(G_n,G_*)}:=\sum_{j\in[N^*]:|\mathcal{C}_{j}|=1}\Bigg[\frac{\sum_{u=1}^{d}\alpha_{1,j}^{(u)}X^{(u)}+\phi_{1,j}}{1+\exp(-(a^*_j)^{\top}X-b^*_j)}F_{0,j}(Y|X)\\
    &\hspace{2cm}+\Big(\sum_{u=1}^{d}\alpha_{1,j}^{(u)}X^{(u)}+\phi_{1,j}\Big)F_{1,j}(Y|X)+\frac{1}{2}\beta_{1,j}F_{2,j}(Y|X)\Bigg],\\
    &\lim_{n\to\infty}\dfrac{T_{n,1,2}(Y|X)}{m_n\mathcal{L}_2(G_n,G_*)}:=\sum_{j\in[N^*]:|\mathcal{C}_{j}|>1}\Bigg[\Bigg(\frac{\sum_{u=1}^{d}\alpha_{1,j}^{(u)}X^{(u)}+\phi_{1,j}}{1+\exp(-(a^*_j)^{\top}X-b^*_j)}\\
    &+\frac{\sum_{u,v=1}^{d}\frac{\alpha_{2,j}^{(uv)}}{1+1_{\{u=v\}}}X^{(u)}X^{(v)}}{1+\exp(-(a^*_j)^{\top}X-b^*_j)}+\frac{\sum_{u=1}^{d}\gamma_{2,j}^{(u)}X^{(u)}+\frac{1}{2}\phi_{2,j}}{1+\exp(-(a^*_j)^{\top}X-b^*_j)}\Bigg)F_{0,j}(Y|X)\\
    &+\Bigg(\sum_{u=1}^{d}\alpha_{1,j}^{(u)}X^{(u)}+\phi_{1,j}+\frac{2\sum_{u,v=1}^{d}\frac{\alpha_{2,j}^{(uv)}}{1+1_{\{u=v\}}}X^{(u)}X^{(v)}}{1+\exp(-(a^*_j)^{\top}X-b^*_j)}\\
    &+\frac{\phi_{2,j}+2\sum_{u=1}^{d}\gamma_{2,j}^{(u)}X^{(u)}}{1+\exp(-(a^*_j)^{\top}X-b^*_j)}\Bigg)F_{1,j}(Y|X)+\Bigg(\frac{1}{2}\beta_{1,j}+\sum_{u,v=1}^{d}\frac{\alpha_{2,j}^{(uv)}X^{(u)}X^{(v)}}{1+1_{\{u=v\}}}+\frac{1}{2}\phi_{2,j}\\
    &+\sum_{u=1}^{d}\gamma_{2,j}^{(u)}X^{(u)}+\frac{1}{2}\cdot\frac{\sum_{u=1}^{d}\gamma_{1,j}^{(u)}X^{(u)}+\gamma_{3,j}}{1+\exp(-(a^*_j)^{\top}X-b^*_j)}\Bigg)F_{2,j}(Y|X)\\
    &+\Big(\sum_{u=1}^{d}\frac{1}{2}\gamma_{1,j}^{(u)}X^{(u)}+\frac{1}{2}\gamma_{3,j}\Big)F_{3,j}(Y|X)+\frac{1}{4}\beta_{2,j}F_{4,j}(Y|X)\Bigg],
\end{align*}
and
\begin{align*}
     &\lim_{n\to\infty}\dfrac{T_{n,2,1}(Y|X)}{m_n\mathcal{L}_2(G_n,G_*)}:=\sum_{j\in[N^*]:|\mathcal{C}_{j}|=1}\frac{\sum_{u=1}^{d}\alpha_{1,j}^{(u)}X^{(u)}+\phi_{1,j}}{1+\exp(-(a^*_j)^{\top}X-b^*_j)}H_j(Y|X),\\
     &\lim_{n\to\infty}\dfrac{T_{n,2,2}(Y|X)}{m_n\mathcal{L}_2(G_n,G_*)}:=\sum_{j\in[N^*]:|\mathcal{C}_{j}|>1}\Bigg[\frac{\sum_{u=1}^{d}\alpha_{1,j}^{(u)}X^{(u)}+\phi_{1,j}}{1+\exp(-(a^*_j)^{\top}X-b^*_j)}\\
     &+\frac{\sum_{u,v=1}^{d}\frac{\alpha_{2,j}^{(uv)}}{1+1_{\{u=v\}}}X^{(u)}X^{(v)}}{1+\exp(-(a^*_j)^{\top}X-b^*_j)}+\frac{\sum_{u=1}^{d}\gamma_{2,j}^{(u)}X^{(u)}+\frac{1}{2}\phi_{2,j}}{1+\exp(-(a^*_j)^{\top}X-b^*_j)}\Bigg]H_j(Y|X),
\end{align*}
and
\begin{align*}
    &\lim_{n\to\infty}\dfrac{T_{n,3}(Y|X)}{m_n\mathcal{L}_2(G_n,G_*)}:=\sum_{j=1}^{N^*}\xi_j[F_{0,j}(Y|X)-H_j(Y|X)].
\end{align*}
It is worth noting that for almost every $X$, the set
\begin{align*}
    \Bigg\{F_{\rho,j}(Y|X), \ H_j(Y|X):0\leq\rho\leq 4, j\in[N^*]\Bigg\}
\end{align*}
is linearly independent w.r.t $Y$. Therefore, it follows that the coefficients of those terms in the limit in equation~\eqref{eq:limit_linear} become zero.

For $j\in[N^*]$ such that $|\mathcal{C}_{j}|=1$, by considering the coefficients of 
\begin{itemize}
    \item $F_{1,j}(Y|X)$, we have $\sum_{u=1}^{d}\alpha_{1,j}^{(u)}X^{(u)}+\phi_{1,j}=0$ for almost surely $X$, indicating that $\alpha_{1,j}^{(u)}=\phi_{1,j}=0$ for all $u\in[d]$;
    \item $F_{0,j}(Y|X)$, we have $\xi_j+\sum_{u=1}^{d}\alpha_{1,j}^{(u)}\cdot\frac{X^{(u)}}{1+\exp(-(a^*_j)^{\top}X-b^*_j)}+\frac{\phi_{1,j}}{1+\exp(-(a^*_j)^{\top}X-b^*_j)}=0$ for almost surely $X$. Since $\alpha_{1,j}^{(u)}=\phi_{1,j}=0$ for all $u\in[d]$, we also get $\xi_j=0$.
    \item $F_{2,j}(Y|X)$, we have $\beta_{1,j}=0$.
\end{itemize}
For $j\in[N^*]$ such that $|\mathcal{C}_{j}|>1$, by considering the coefficients of 
\begin{itemize}
    \item $F_{1,j}(Y|X)$, we have
    \begin{align*}
        &\sum_{u=1}^{d}\alpha_{1,j}^{(u)}X^{(u)}+\phi_{1,j}+\frac{2\sum_{u,v=1}^{d}\frac{\alpha_{2,j}^{(uv)}}{1+1_{\{u=v\}}}X^{(u)}X^{(v)}}{1+\exp(-(a^*_j)^{\top}X-b^*_j)}+\frac{\phi_{2,j}+2\sum_{u=1}^{d}\gamma_{2,j}^{(u)}X^{(u)}}{1+\exp(-(a^*_j)^{\top}X-b^*_j)}=0,
    \end{align*}
    for almost surely $X$. Since the set 
    \begin{align*}
        \Bigg\{1, \ X^{(u)}, \ \frac{1}{1+\exp(-(a^*_j)^{\top}X-b^*_j)}, \ \frac{X^{(u)}}{1+\exp(-(a^*_j)^{\top}X-b^*_j)},\\
        \frac{X^{(u)}X^{(v)}}{1+\exp(-(a^*_j)^{\top}X-b^*_j)}:u,v\in[d]\Bigg\}
    \end{align*}
    is linearly independent w.r.t $X$, we deduce $\alpha_{1,j}^{(u)}=\phi_{1,j}=\alpha_{2,j}^{(uv)}=\phi_{2,j}=\gamma_{2,j}^{(u)}=0$ for all $u,v\in[d]$.
    \item $F_{0,j}(Y|X)$, we have 
    \begin{align*}
        &\xi_j+\frac{\sum_{u=1}^{d}\alpha_{1,j}^{(u)}X^{(u)}+\phi_{1,j}}{1+\exp(-(a^*_j)^{\top}X-b^*_j)}\\
    &\hspace{3cm}+\frac{\sum_{u,v=1}^{d}\frac{\alpha_{2,j}^{(uv)}}{1+1_{\{u=v\}}}X^{(u)}X^{(v)}}{1+\exp(-(a^*_j)^{\top}X-b^*_j)}+\frac{\sum_{u=1}^{d}\gamma_{2,j}^{(u)}X^{(u)}+\frac{1}{2}\phi_{2,j}}{1+\exp(-(a^*_j)^{\top}X-b^*_j)}=0,
    \end{align*}
    for almost surely $X$. Since $\alpha_{1,j}^{(u)}=\phi_{1,j}=\alpha_{2,j}^{(uv)}=\phi_{2,j}=\gamma_{2,j}^{(u)}=0$ for all $u,v\in[d]$, we get $\xi_j=0$.
    \item $F_{3,j}(Y|X)$, we have $\sum_{u=1}^{d}\frac{1}{2}\gamma_{1,j}^{(u)}X^{(u)}+\frac{1}{2}\gamma_{3,j}=0$ for almost surely $X$, indicating that $\gamma_{1,j}^{(u)}=\gamma_{3,j}=0$ for all $u\in[d]$;
    \item $F_{2,j}(Y|X)$, we have 
    \begin{align*}
        \frac{1}{2}\beta_{1,j}+\sum_{u,v=1}^{d}\frac{\alpha_{2,j}^{(uv)}X^{(u)}X^{(v)}}{1+1_{\{u=v\}}}+\frac{1}{2}\phi_{2,j}
    +\sum_{u=1}^{d}\gamma_{2,j}^{(u)}X^{(u)}+\frac{1}{2}\frac{\sum_{u=1}^{d}\gamma_{1,j}^{(u)}X^{(u)}+\gamma_{3,j}}{1+\exp(-(a^*_j)^{\top}X-b^*_j)}=0,
    \end{align*}
    for almost surely $X$. Since $\alpha_{2,j}^{(uv)}=\phi_{2,j}=\gamma_{2,j}^{(u)}=\gamma_{1,j}^{(u)}=\gamma_{3,j}=0$ for all $u,v\in[d]$, we also get $\beta_{1,j}=0$.
    \item $F_{4,j}(Y|X)$, we have $\beta_{2,j}=0$.
\end{itemize}
Putting the above results together, we have $\xi_j=\alpha_{1,j}^{(u)}=\phi_{1,j}=\beta_{1,j}=\alpha_{2,j}^{(uv)}=\phi_{2,j}=\beta_{2,j}=\gamma_{1,j}^{(u)}=\gamma_{2,j}^{(u)}=\gamma_{3,j}=0$ for all $j\in[N^*]$ and $u,v\in[d]$. This contradicts the fact that at least one among them is different from zero. Consequently, we achieve the local part in~\eqref{eq:local_part_linear}.

\subsection{Proof of Proposition~\ref{prop:density_estimation}}
\label{appendix:density_estimation}
In this proof, we first present some fundamental results on the density estimation problem for M-estimators in \cite{Vandegeer-2000} in Appendix~\ref{appendix:fundamental}, and then provide the main proof in Appendix~\ref{appendix:main_proof}. 

\subsubsection{Preliminaries}
\label{appendix:fundamental}
To streamline our discussion, let us introduce some necessary concepts from the empirical process theory. In particular, let $\mathcal{P}_k(\Theta)$ be the set of all conditional densities with respect to mixing measures in $\mathcal{G}_{N}(\Theta)$, i.e.
\begin{align*}
    \mathcal{P}_{N}(\Theta):=\{p_{G}(Y|X):G\in\mathcal{G}_{N}(\Theta)\}.
\end{align*}
Additionally, we also consider two following variants of the set $\mathcal{P}_{N}(\Theta)$:
\begin{align*}
    \overline{\mathcal{P}}_k(\Theta)&:=\{p_{(G+G_*)/2}(Y|X):G\in\mathcal{G}_{N}(\Theta)\},\\
    \overline{\mathcal{P}}^{1/2}_{N}(\Theta)&:=\{p^{1/2}_{(G+G_*)/2}(Y|X):G\in\mathcal{G}_{N}(\Theta)\}.
\end{align*}
Next, we define for each $\delta>0$ a Hellinger ball centered around the true conditional density $p_{G_*}(Y|X)$ and intersect with the set $\overline{\mathcal{P}}^{1/2}_{N}(\Theta)$ as below
\begin{align*}
    \overline{\mathcal{P}}^{1/2}_{N}(\Theta,\delta):=\{p^{1/2}(Y|X)\in\overline{\mathcal{P}}^{1/2}_{N}(\Theta):h(p_{G},p_{G_*})\leq\delta\}.
\end{align*}
Moreover, the size of this Hellinger ball is quantified by the following term:
\begin{align}
    \label{eq:integral}
    \mathcal{J}_B(\delta,\overline{\mathcal{P}}^{1/2}_{N}(\Theta,\delta)):=\int_{\delta^2/2^{13}}^{\delta}H_B^{1/2}(t,\overline{\mathcal{P}}^{1/2}_{N}(\Theta,t),\|\cdot\|_2)\dint t\vee\delta,
\end{align}
where $H_B(t,\overline{\mathcal{P}}^{1/2}_{N}(\Theta,t),\|\cdot\|_2)$ stands for the bracketing entropy of $\overline{\mathcal{P}}^{1/2}_{N}(\Theta,t)$ under the $L^2$-norm, and $t\vee\delta:=\max\{t,\delta\}$. Now, we are ready to recall the results in \cite{Vandegeer-2000}. 
\begin{lemma}[Theorem 7.4,\cite{Vandegeer-2000}]
    \label{lemma:vandegeer}
    Take $\Psi(\delta)\geq\mathcal{J}_B(\delta,\overline{\mathcal{P}}^{1/2}_{N}(\Theta,\delta))$ such that $\Psi(\delta)/\delta^2$ is a non-increasing function of $\delta$. Then, for a universal constant $c$ and $\sqrt{n}\delta^2_n\geq c\Psi(\delta_n)$, we achieve that
    \begin{align*}
        \mathbb{P}\Big(\bbE_X[h(p_{\widehat{G}_n}(\cdot|X),p_{G_*}(\cdot|X))]>\delta\Big)\leq c\exp(-n\delta^2/c^2),
    \end{align*}
    for any $\delta\geq\delta_n$.
\end{lemma}
Proof of Lemma~\ref{lemma:vandegeer} is available in \cite{Vandegeer-2000}. Apart from this result, we also need to introduce the upper bounds of the covering number $N(\varepsilon,\mathcal{P}_{N}(\Theta),\|\cdot\|_{\infty})$ and the bracketing entropy $H_B(\varepsilon,\mathcal{P}_{N}(\Theta),\|\cdot\|_2)$ as follows:
\begin{lemma}
    \label{lemma:upper_bounds}
    Suppose that $\Theta$ is a bounded set, then we have for any $\varepsilon\in(0,1/2)$ that
    \begin{itemize}
        \item[(a)] $\log N(\varepsilon,\mathcal{P}_{N}(\Theta),\|\cdot\|_{\infty})\lesssim\log(1/\varepsilon)$;
        \item[(b)] $H_B(\varepsilon,\mathcal{P}_{N}(\Theta),\|\cdot\|_2)\lesssim\log(1/\varepsilon)$. 
    \end{itemize}
\end{lemma}
\begin{proof}[Proof of Lemma~\ref{lemma:upper_bounds}]
    \textbf{Part (a).} Recall that $\Theta$ is a compact set, then there exists an $\varepsilon$-cover, which we denote as $\overline{\Theta}_{\varepsilon}$. Moreover, it can be verified that $|\overline{\Theta}_{\varepsilon}|\leq\mathcal{O}(\varepsilon^{-(d_2+1)N})$. Next, for each mixing measure $G=\sum_{i=1}^{N}\delta_{(W_{e_i},\nu_i)}\in\mathcal{G}_{N}(\Theta)$, we consider another one $\overline{G}=\sum_{i=1}^{N}\delta_{(\overline{W}_{e_i},\overline{\nu}_i)}$, where $(\overline{W}_{e_i},\overline{\nu}_i)\in\overline{\Theta}_{\varepsilon}$ is the closest point to $(W_{e_i},\nu_i)$ in this set for any $i\in[N]$. 
Subsequently, we demonstrate that the set
\begin{align*}
    \mathcal{Q}:=\Big\{p_{\overline{G}}(Y|X):(\overline{W}_{e_i},\overline{\nu}_i)\in\overline{\Theta}_{\varepsilon},\forall i\in[N]\Big\}
\end{align*}
is an $\varepsilon$-cover of the metric space $(\mathcal{P}_{N}(\Theta),\|\cdot\|_{\infty})$. In other words, we need to show that for any $p_{G}(Y|X)\in\mathcal{P}_{N}(\Theta)$, there exists some density $p_{\overline{G}}(Y|X)\in\mathcal{Q}$ such that $\|p_{G}-p_{\overline{G}}\|_{\infty}\lesssim\varepsilon$.
Next, we decompose the term $T_n(Y|X):=\Big[\sum_{j=1}^{N}\exp(\log(1+\exp(g(X,\overline{W}_{e_j}))))\Big]\cdot[p_{G}(Y|X)-p_{\overline{G}}(Y|X)]$ as
\begin{align*}
    &T_n(Y|X)=\sum_{i=1}^{N}\exp(\log(1+\exp(g(X,W_{e_i}))))\Big[f(Y|g(X,W_{e_i}),\nu_i)-f(Y|g(X,\overline{W}_{e_i}),\overline{\nu}_i)\Big]\\
    &+\sum_{i=1}^{N}\Big[\exp(\log(1+\exp(g(X,W_{e_i}))))-\exp(\log(1+\exp(g(X,\overline{W}_{e_j}))))\Big]\cdot\Big[f(Y|g(X,\overline{W}_{e_i}),\overline{\nu}_i)-p_{G}(Y|X)\Big].
\end{align*}
As $\Theta$ and $\mathcal{X}$ are bounded, we may assume that $\exp(\log(1+\exp(g(X,W_{e_i}))))\leq B_1$ and $|f(Y|g(X,\overline{W}_{e_i}),\overline{\nu}_i)-p_{G}(Y|X)|\leq B_2$ for some positive constants $B_1,B_2$. Thus, we obtain that
\begin{align*}
    |T_n(Y|X)|\lesssim \sum_{i=1}^{N}B_1\cdot\Big[\|W_{e_i}-\overline{W}_{e_i}\|+|\nu_i-\overline{\nu}_i|\Big]+\sum_{i=1}^{N}B_2\cdot\|W_{e_i}-\overline{W}_{e_i}\|\lesssim \varepsilon.
\end{align*}
Additionally, since the term  $\sum_{j=1}^{K}\exp(|g(X,\overline{W}_{e_j})|)$ is bounded, we obtain $|p_{G}(Y|X)-p_{\overline{G}}(Y|X)|\lesssim\varepsilon$ for almost surely $(X,Y)$, or equivalently,
\begin{align*}
    \|p_{G}-p_{\overline{G}}\|_{\infty}=\sup_{(X,Y)\in\mathcal{X}\times\mathcal{Y}}|p_{G}(Y|X)-p_{\overline{G}}(Y|X)|\lesssim\varepsilon.
\end{align*}
This result indicates that $\mathcal{Q}$ is an $\varepsilon$-cover of the metric space $(\mathcal{P}_{N}(\Theta),\|\cdot\|_{\infty})$. Therefore, we get
\begin{align*}
    N(\varepsilon,\mathcal{P}_{N}(\Theta),\|\cdot\|_{\infty})\leq |\overline{\Theta}_{\varepsilon}|\leq \mathcal{O}(\varepsilon^{-(d_2+1)N}),
\end{align*}
or equivalently,
\begin{align*}
    \log N(\varepsilon,\mathcal{P}_{N}(\Theta),\|\cdot\|_{\infty})\leq |\overline{\Theta}_{\varepsilon}|\lesssim \log(1/\varepsilon).
\end{align*}
\textbf{Part (b).} Firstly, we will derive an upper bound for the Gaussian experts $f(Y|g(X,W_e),\nu)$. Since $\Theta$ is a compact set, we have $|g(X,W_e)|\leq M_1$ and $M_2\leq \nu\leq M_3$ for any $X\in\mathcal{X}$ and $(W_e,\nu)\in\Theta$. Then, it follows that $f(Y|g(X,W_e),\nu)\leq B(Y|X)$, where
\begin{align*}
    B(Y|X):=\begin{cases}
        \dfrac{1}{\sqrt{2\pi M_2}}\exp(-Y^2/(8M_3^2)), \hspace{1cm} \text{for } |Y|\geq 2M_1\\
        \dfrac{1}{\sqrt{2\pi M_2}}, \hspace{3.7cm} \text{for } |Y|<2M_1,
    \end{cases}
\end{align*}
for any $X\in\mathcal{X}$. Next, let $\eta\leq\varepsilon$ be some positive constant that we choose later, then we denote $\{\pi_1,\pi_2,\ldots,\pi_N\}$ as an $\eta$-cover over $\mathcal{P}_{N}(\Theta)$. Based on this cover, we build the following brackets $L_i(Y|X):=\max\{\pi_i(Y|X)-\eta,0\}$ and $U_i(Y|X):=\max\{\pi_i(Y|X)+\eta,B(Y|X)\}$, for any $i\in[N]$. We can validate that $\mathcal{P}_{N}(Y|X)\subseteq\bigcup_{i=1}^{N}[L_i(Y|X),U_i(Y|X)]$ and $U_i(X,Y)-L_i(X,Y)\leq\min\{2\eta,B(Y|X)\}$. As a result, we have
\begin{align*}
    \|U_i-L_i\|_2=\Big(\int[U_i(Y|X)-L_i(Y|X)]^2\dint (X,Y)\Big)^{1/2}\leq 2\eta.
\end{align*} 
The above result implies that
\begin{align*}
    H_B(2\eta,\mathcal{P}_{N}(\Theta),\|\cdot\|_2)\leq \log N(\eta,\mathcal{P}_{N}(\Theta),\|\cdot\|_{\infty})\lesssim\log(1/\eta).
\end{align*}
Then, by setting $\eta=\varepsilon/2$, we arrive at
\begin{align*}
    H_B(\varepsilon,\mathcal{P}_{N}(\Theta),\|\cdot\|_1)\lesssim\log(1/\varepsilon).
\end{align*}
Hence, the proof is completed.
\end{proof}

\subsubsection{Main Proof}
\label{appendix:main_proof}
Since $\overline{\mathcal{P}}^{1/2}_{N}(\Theta,t)\subset\overline{\mathcal{P}}^{1/2}_{N}(\Theta)$ for any $t>0$, we have
    \begin{align}
        \label{eq:bracketing_inequality}
        H_B(t,\overline{\mathcal{P}}^{1/2}_{N}(\Theta,t),\|\cdot\|_{2})&\leq H_B(t,\overline{\mathcal{P}}^{1/2}_{N}(\Theta),\|\cdot\|_{2})=H_B(t/\sqrt{2},\overline{\mathcal{P}}_{N}(\Theta),h),
    \end{align}
    where the last equality is due to the relationship between the Hellinger distance $h$ and the $L^2$-norm. Note that for any two mixing measure $G$ and $G'$, Lemma 4.2 in \cite{Vandegeer-2000} indicates that
    \begin{align*}
        h^2\Big(\frac{1}{2}p_{G}+\frac{1}{2}p_{G_*},\frac{1}{2}p_{G'}+\frac{1}{2}p_{G_*}\Big)\leq\frac{1}{2}h^2(p_{G},p_{G'}),
    \end{align*}
    which yields $H_B(t/\sqrt{2},\overline{\mathcal{P}}_{N}(\Theta),h)\leq H_B(t,\mathcal{F}_{k_1,k_2}(\Theta),h)$. This result together with equation~\eqref{eq:bracketing_inequality} implies that 
    \begin{align*}
        H_B(t,\overline{\mathcal{P}}^{1/2}_{N}(\Theta,t),\|\cdot\|_{2})\leq H_B(t,\mathcal{P}_{N}(\Theta),h).
    \end{align*}
From \eqref{eq:integral} and part (b) of Lemma~\ref{lemma:upper_bounds}, we have that
\begin{align*}
    \mathcal{J}_B(\delta,\overline{\mathcal{P}}^{1/2}_{N}(\Theta,\delta))&=\int_{\delta^2/2^{13}}^{\delta}H_B^{1/2}(t,\overline{\mathcal{P}}^{1/2}_{N}(\Theta,t),\|\cdot\|_2)\dint t\vee\delta\\
    &\leq\int_{\delta^2/2^{13}}^{\delta}H_B^{1/2}(t,\overline{\mathcal{P}}^{1/2}_{N}(\Theta,t),h)\dint t\vee\delta\\
    &\lesssim\int_{\delta^2/2^{13}}^{\delta}\log(1/t)\dint t\vee\delta.
\end{align*}
Next, let $\Psi(\delta)=\delta\sqrt{\log(1/\delta)}$, then 
it can be verified that $\Psi(\delta)/\delta^2$ is a non-increasing function of $\delta$. Furthermore, the above result indicates that $\Psi(\delta)\geq \mathcal{J}_B(\delta,\widetilde{\mathcal{F}}^{1/2}_{k_1,k_2}(\Theta,\delta),\|\cdot\|_{2})$. By considering the sequence $(\delta_n)$ defined as $\delta_n:=\sqrt{\log(n)/n}$, we have $\sqrt{n}\delta_n^2\geq c\Psi(\delta_n)$ for some universal constant $c>0$.
It follows from Lemma~\ref{lemma:vandegeer} that 
\begin{align*}
    \mathbb{P}\Big(\bbE_X[h(p_{\widehat{G}_n}(\cdot|X),p_{G_*}(\cdot|X))]>C\sqrt{\log(n)/n}\Big)\lesssim\exp(-c\log(n)),
\end{align*}
for some universal constant $C>0$ depending only on $\Theta$. Since the Total Variation distance is upper bounded by the Hellinger distance, we deduce
\begin{align*}
    \mathbb{P}\Big(\bbE_X[V(p_{\widehat{G}_n}(\cdot|X),p_{G_*}(\cdot|X))]>C\sqrt{\log(n)/n}\Big)\lesssim\exp(-c\log(n)),
\end{align*}
or equivalently,
\begin{align*}
    \bbE_X[V(p_{\widehat{G}_n}(\cdot|X),p_{G_*}(\cdot|X))]=\mathcal{O}_P(\sqrt{\log(n)/n}).
\end{align*}
Hence, the proof is completed.

\bibliography{main}
\bibliographystyle{abbrv}
\end{document}